\documentclass[12pt]{article}
\usepackage{amsmath, amsfonts, mathtools}
\usepackage{graphicx}
\usepackage{url} 

\usepackage[
  backend=biber,
  style=apa,
  natbib=true,
  uniquename=init,    
]{biblatex}
\usepackage{enumitem}
\usepackage[table,dvipsnames]{xcolor}
\usepackage{booktabs}
\usepackage{makecell}
\usepackage{rotating}
\usepackage{caption}
\usepackage{hyperref}
\hypersetup{
    colorlinks=true,
    linkcolor=NavyBlue,
    filecolor=magenta,      
    urlcolor=cyan,
    citecolor=MidnightBlue
}

\newcommand{\blind}{0}

\renewcommand{\Vec}[1]{\boldsymbol{#1}}
\newcommand{\Mat}[1]{\mathbf{#1}}

\newcommand{\fh}{\hat{f}}

\newcommand{\bbeta}{\boldsymbol{\beta}}

\newcommand{\bA}{\mathbf{A}}

\newcommand{\bD}{\mathbf{D}}

\newcommand{\bL}{\mathbf{L}}

\newcommand{\sL}{\mathcal{L}}

\newcommand{\sT}{\mathcal{T}}

\newcommand{\R}{\mathbb{R}}

\newcommand{\E}{\mathbb{E}}

\newcommand{\diag}{\textnormal{diag}}

\DeclarePairedDelimiter{\braces}{\lbrace}{\rbrace}

\DeclarePairedDelimiter{\norm}{\|}{\|}

\DeclareMathOperator*{\argmin}{arg\,min}

\newcommand{\node}{s}
\newcommand{\tree}{t}

\definecolor{cyan}{cmyk}{1, 0.2, 0, 0}

\definecolor{green}{cmyk}{1,0,1,0}

\definecolor{amethyst}{rgb}{0.6, 0.4, 0.8}

\definecolor{darkred}{rgb}{0.545, 0, 0}

\definecolor{orange}{HTML}{CC8400}

\newcommand{\plus}{\texttt{+}}

\newcommand{\method}{NeRF\plus}
\newcommand{\rfplus}{RF\plus}
\newcommand{\mdiplus}{MDI\plus}

\newcommand{\permimp}[2]{\text{PI}_{#1, #2}}
\newcommand{\mdi}[2]{\text{MDI}^{+}_{#1, #2}}
\newcommand{\lfi}[2]{\text{LFI}_{#1, #2}}

\begin{document}

\def\spacingset#1{\renewcommand{\baselinestretch}%
{#1}\small\normalsize} \spacingset{1}

\newcommand{\B}[1]{{\color{blue}{#1}}}


\if0\blind
{
  \title{\bf Interpretable Network-assisted Random Forest$\plus$}
  \author{Tiffany M. Tang\\
    Department of Applied and Computational Mathematics and Statistics, \\University of Notre Dame\\
    and \\
    Elizaveta Levina \\
    Department of Statistics, University of Michigan\\
    and \\
    Ji Zhu \\
    Department of Statistics, University of Michigan}
  \date{}
  \maketitle
} \fi

\if1\blind
{
  \bigskip
  \bigskip
  \bigskip
  \begin{center}
    {\LARGE\bf Interpretable Network-assisted Random Forest$\plus$}
\end{center}
  \medskip
} \fi

\bigskip
\begin{abstract}
Machine learning algorithms often assume that training samples are independent. When data points are connected by a network, the induced dependency between samples is both a challenge, reducing effective sample size, and an opportunity to improve prediction by leveraging information from network neighbors. Multiple methods taking advantage of this opportunity are available, but many, including graph neural networks, are not easily interpretable, limiting their usefulness for understanding how models make predictions. Others, such as network-assisted linear regression, are interpretable but often yield worse prediction performance. We bridge this gap by proposing a family of flexible network-assisted models built upon a generalization of random forests (\rfplus), which achieves highly-competitive prediction accuracy and can be understood through intrinsic interpretability measures, derived directly from the model parameters and structure. In particular, we develop a suite of interpretation tools that enable researchers to both identify important features that drive model predictions and quantify the importance of the network contribution to prediction.  Importantly, we provide global and local feature importances as well as sample influence measures to assess the impact of individual observations.  This suite of tools broadens the scope and applicability of network-assisted machine learning for high-impact problems where interpretability and transparency are essential.
\end{abstract}

\noindent%
\vfill


\section{Introduction}\label{sec:intro}
Current machine learning algorithms achieve state-of-the-art prediction accuracy for many tasks, but the vast majority of these methods assume that the samples in the training data are independent of one another. In many modern applications, data points are connected by a network, which creates dependencies between samples. For example, when investigating student beliefs at a given school, there is inevitably an underlying social network (where nodes represent students and edges represent social or friendship ties between two students) which elicits peer-to-peer effects since student beliefs are highly likely to be influenced by their friends \citep{paluck2016changing}.    These network-induced dependencies present a challenge as they reduce the effective sample size and introduce unwanted biases, leading to poor generalization performance.  At the same time, they offer an opportunity to improve prediction by exploiting potential similarities between network neighbors, which we informally refer to as network cohesion: for example, knowing one student's beliefs can help predict the beliefs of that student's close friends, and generally, network neighbors behaving more similarly than non-neighbors (cohesion) is widely observed in practice.   Though some methods for improving prediction by leveraging the network information have been developed (to be discussed below), it remains challenging to balance model flexibility with interpretability, and especially to disentangle the importance of network connections for prediction from that of other node features.  

In general, finding a balance between flexibility and interpretability has long been a major question in statistical learning; see for example \cite{rudin2019stop}. In practice, random forests (RFs) \citep{breiman2001random} have often emerged as a popular solution, striking a reasonable balance between strong prediction performance and interpretability. Notably, RFs generally do not require massive amounts of training data and have been shown to perform competitively compared to deep learning across a wide array of medium-sized datasets \citep{grinsztajn2022tree}. RFs have also been widely adopted in applications including clinical and health diagnostics \citep{loef2022using}, scientific experimental recommendations \citep{wang2023epistasis}, and ecological modeling \citep{simon2023interpreting}, primarily due to their ability to interpret how each feature contributes to the model's predictions. Recently, \citet{agarwal2023mdi+} built upon a connection between decision trees and linear regression \citep{klusowski2024large, golea1997generalization} to develop Random Forest Plus (\rfplus), an extension of RFs combining both nonlinear tree-based features and linear features. \citet{agarwal2023mdi+} showed \rfplus~outperforms ordinary RF on a variety of prediction tasks while maintaining interpretability. We take this method as a basis for our proposed network-assisted approach, \textbf{Ne}twork-assisted \textbf{R}andom \textbf{F}orest Plus (\textbf{\method}), because of its flexibility:   \method~inherits the ability of \rfplus~to include both linear and nonlinear features and to provide local and global importance measures. Moreover, it offers a framework to incorporate the network in multiple ways, as we will discuss below. We will demonstrate that, like \rfplus, \method~methods achieve highly competitive prediction accuracy while maintaining interpretability.

Several approaches to incorporate network information into prediction are available.  Starting from the flexible end of the spectrum, the class of deep learning methods called graph neural networks (GNNs) \citep{kipf2016semi,  hamilton2017inductive, wu2020comprehensive, zhou2020graph}, work by incorporating the network connections into the deep learning architecture, summing up inputs from neighbors when making predictions for a node.  Like all deep learning models, these are generally black-box methods. Though post-hoc explainability tools such as GNNExplainer \citep{ying2019gnnexplainer}, GraphSVX \citep{duval2021graphsvx}, and others \citep{GNNBook2022} can aid in explaining how GNNs make their predictions, these methods rely on unstable approximations of the trained GNN, which are easily susceptible to adversarial attacks and are known to often result in misleading explanations \citep{li2024explainable}. Moreover, GNNs typically require large amounts of data to achieve their state-of-the-art prediction accuracy, limiting their utility in many high-impact biomedical and social science problems, where data availability is often limited by cost or by the nature of the problem, and where the need for transparency and ability to understand how the model is making its predictions is at least as important as the prediction itself. 

In traditional statistical learning, two main approaches have been used for incorporating network information into prediction modeling -- engineering network-based features and regularization.
In the former approach, researchers augment their covariate matrix with additional node features \citep{chen2018tutorial, lunde2023conformal}. These node features can be constructed from network summaries (e.g., node degrees, or neighborhood averages of other features), which are a popular approach in econometrics \citep{manski1993identification}, or from latent variables obtained from fitting a latent variable network model \citep[e.g.,][]{hoff2002latent, belkin2003laplacian, sussman2012consistent}. This approach is attractively modular: any network latent variable model or embedding method can be used to produce features that can then be fed into any learning algorithm. However, these network embeddings are typically learned in an unsupervised manner and thus not directly optimized for the prediction task at hand.

Regularization-based approaches, on the other hand, provide a supervised alternative to incorporating network information into prediction models \citep{chen2010graph, wang2016trend,li2019prediction}. An example of such regularization is regression with network cohesion (RNC), introduced by \cite{li2019prediction}, with inference tools developed under additional assumptions by \cite{le2022linear}. Briefly, RNC adds a node-specific network effect to a standard linear regression model and uses a ``cohesion penalty" to shrink network effects between neighboring nodes towards each other, hence encouraging its predictions for network neighbors to be more similar. This approach is readily interpretable but may in general lack flexibility to provide competitive prediction performance due to its underlying linear model assumption. 

Finally, network-assisted variants of Bayesian Additive Regression Trees have been proposed by \citet{luo2022bamdt} (BAMDT) and \citet{deshpande2022flexbart} (Network BART). These methods augment the splitting options during tree construction by allowing the tree to either split on a variable as usual, or by partitioning the network into two connected components, implementing an efficient search strategy over network partitions through pre-computed minimum spanning trees. While the network-assisted BART models have been shown to improve prediction, the interpretation of such models has received significantly less attention. There are currently no available interpretability tools for Network BART. BAMDT, on the other hand, measures feature importances by computing the proportion of splits made using each feature; however, this approach often favors features with more potential split points \citep{strobl2007bias}. Due to the large number of potential network splits, we will see that this leads to a bias in favor of network features, making it difficult to interpret and disentangle the importance of the network from that of other predictor features.

In this paper, we address the need for an interpretable {\em and} flexible prediction method on networks by developing Network-assisted Random Forest Plus (\method). We will show \method~achieves a good balance between flexibility and interpretability for network-assisted prediction, maintaining competitive prediction accuracy while providing a suite of tools to interpret findings. 
Importantly, unlike GNNs which rely on post-hoc explainability tools, \method~comes with intrinsic, model-based interpretations, derived directly from the fitted model parameters and structure.
These interpretability tools enable researchers to quantify how a feature or the network contributes to the model's predictions, either globally or for a particular sample, and to identify important or unusual training points through influence scores. 
It is our hope that this suite of interpretability tools will facilitate broader adoption of network-assisted prediction in a variety of high-stakes applications where interpretation is crucial.

The remainder of the paper is organized as follows. 
We introduce the \method~prediction models in Section~\ref{sec:nerf} and the suite of interpretability tools for \method~in Section~\ref{sec:interpret}. In Section~\ref{sec:sims}, we compare \method~to competitors on simulated data and examine its behavior on several illustrative cases.   Section \ref{sec:case_study} presents an in-depth analysis of two case studies,  one on school conflict and another on crime in Philadelphia. We conclude with a discussion in Section~\ref{sec:discussion}.

\section{Network-assisted Random Forest Plus (\method)}\label{sec:nerf}
In this section, we introduce \method, a family of network-assisted RF-based prediction models which leverage network information to improve the overall prediction performance.  
To begin, we set up notation and review relevant background in Section~\ref{subsec:nerf_background} before introducing \method~in Section~\ref{subsec:nerf}. 

\subsection{Notation and Background}\label{subsec:nerf_background}

Suppose we have training data for $n$ samples with $p$ features (covariates) and a quantitative response of interest. Let $\Mat{X} \in \R^{n \times p}$ be the observed feature matrix and $\Vec{y} \in \R^n$ the vector of responses for the training data. Without loss of generality, assume that the responses $\Vec{y}$ and each column of $\Mat{X}$ have been centered to have mean 0. Suppose that we also observed an undirected network $G$ connecting the $n$ samples, represented by an $n \times n$ adjacency matrix $\bA$. We allow $G$ to have potentially multiple connected components. For simplicity, assume $\bA$ is binary, with $A_{ij} = 1$ if samples $i$ and $j$ are connected in $G$ and $0$ otherwise. We note, however, that all proposed methods easily extend to weighted networks, where $A_{ij}$ is a non-negative weight capturing the strength of the connection between samples $i$ and $j$. Define the unnormalized network Laplacian by $\bL = \bD - \bA$, where $\bD = \diag(d_1, \dots, d_n)$ is the diagonal matrix of node degrees $d_i = \sum_{j=1}^n A_{ij}$.  

Our proposed method, \method, builds on ideas from three key tools --- network embeddings, regression with network cohesion, and RF+ --- reviewed next.

\subsubsection{Network embeddings}\label{subsec:network_embeddings}

As mentioned in Section \ref{sec:intro}, a simple and popular network-assisted approach is to augment the original covariate matrix $\Mat{X}$ with additional network embedding features $\Mat{Z} \in \R^{n \times r}$, where $r$ is the number of latent dimensions, and then apply any standard prediction model to the augmented feature matrix $[\Mat{X}, \Mat{Z}]$ to predict the response $\Vec{y}$.

Such $\Mat{Z}$ can be derived from the network using various approaches including spectral embeddings \citep{belkin2003laplacian, sussman2012consistent}, GNN latent representations \citep{kipf2016semi, hamilton2017inductive}, and/or other node embeddings \citep{grover2016node2vec}.
When the goal is purely prediction, these network embeddings can generally be learned using the network $\Mat{A}$ and training data $(\Mat{X}, \Vec{y})$ together. 
However, given our aim for interpretability, we focus on the case where the network embedding features $\Mat{Z}$ are learned using only the network $\Mat{A}$.
In particular, for simplicity, we use the Laplacian spectral embedding \citep{belkin2003laplacian} to construct $\Mat{Z}$ from $\Mat{A}$ for all empirical studies in this work.

\subsubsection{Regression with Network Cohesion}\label{subsec:rnc}

An alternative approach to network-assisted prediction is via regularization with network-induced penalties.   One example of this approach is regression with network cohesion (RNC) \citep{li2019prediction}. RNC adds a node-specific network effect to ordinary linear regression models and models the response $\Vec{y}$ as
\begin{align*}
    \Vec{y} = \Vec{\alpha} + \Mat{X} \Vec{\beta} + \Vec{\epsilon},
\end{align*}
where $\Vec{\beta} \in \R^p$ is a vector of regression coefficients, $\Vec{\epsilon} \in \R^n$ is a noise vector, and $\Vec{\alpha} \in \R^n$ is a vector of individual node effects (replacing the traditional scalar intercept). This model (which requires regularization to estimate its $n+p$ parameters) is fitted by minimizing the usual least squares loss plus a network cohesion penalty, that is, 
\begin{align}\label{eq:rnc}
   (\hat{\Vec{\alpha}}, \hat{\Vec{\beta}}) = 
   \argmin_{\Vec{\alpha} \in \R^n, \; \Vec{\beta} \in \R^p} \;\; 
   \norm{\Vec{y} - \Vec{\alpha} - \Mat{X} \Vec{\beta}}_2^2 
   + \lambda_{\alpha} \Vec{\alpha}^{\top} \Mat{L} \Vec{\alpha},
\end{align}
where $\lambda_{\alpha} > 0$ is a tuning parameter, $\Mat{L}$ is the Laplacian of the network $\Mat{A}$, and $\Vec{\alpha}^{\top} \Mat{L} \Vec{\alpha} = \sum_{i, j} A_{ij} (\alpha_i - \alpha_j)^2$ is the network cohesion penalty, shrinking node effects towards each other when nodes are connected. If desired, penalties on $\bbeta$ can be added (e.g., if $p > n$), and \eqref{eq:rnc} can be extended to general loss functions.
Additionally, to guarantee existence of a solution for \eqref{eq:rnc}, \citet{li2019prediction} replace the Laplacian $\Mat{L}$ with a regularized Laplacian $\Mat{L} + \lambda_L \Mat{I}$ for some small $\lambda_L > 0$. For all empirical studies in this paper, we use the regularized Laplacian with $\lambda_L = 0.05$, but still denote it by $\Mat{L}$ for notational simplicity.  

\subsubsection{Random Forest Plus}\label{subsec:rfplus}

Random Forest Plus (\rfplus) was proposed by \citet{agarwal2023mdi+} as a generalization of RF which combines the powerful predictive capabilities of RFs with the ease of interpretability of generalized linear models.  The key tool of \rfplus~is a particular representation of decision tree predictions through a linear model, reported in previous work \citep{klusowski2024large,agarwal2022hierarchical}. 
Specifically, suppose we are given a fitted (regression or classification) decision tree $\tree$, trained on data $(\Mat{X}, \Vec{y})$ and composed of $m_{\tree}$ total decision splits such that (i) each split in the decision tree is binary (i.e., it has two child nodes), and (ii) the tree predictions are made by averaging the training responses $\Vec{y}$ in each leaf node. These assumptions are satisfied by commonly used decision tree algorithms such as CART \citep{breiman1984classification}. Then, one can extract a map from the original features $\Mat{X}$ to new features $\Psi_{\tree}(\Mat{X})$ such that the tree predictions $\hat{\Vec{y}}^{(\tree)}$ can be \textit{exactly} recovered by ordinary linear regression of the response $\Vec{y}$ on the new features $\Psi_{\tree}(\Mat{X})$, i.e.,
\begin{align}\label{eq:connection_linear_trees}
    \hat{\Vec{y}}^{(\tree)} = \Psi_{\tree}(\Mat{X}) \hat{\Vec{\gamma}}^{(\tree)}, \quad \text{where} \quad \hat{\Vec{\gamma}}^{(\tree)} = \argmin_{\Vec{\gamma} \in \R^{m_{\tree}}} \;\; \norm{\Vec{y} -  \Psi_{\tree}(\Mat{X}) \Vec{\gamma}}_2^2.
\end{align}

These new features $\Psi_{\tree}(\Mat{X})$ are constructed as follows. Given the decision tree $\tree$, each internal node $\node$ in the tree $\tree$ gets split into two children nodes defined by $\node_L = \braces{\Vec{x} \in \node \colon x_{k} \leq \tau}$ and $\node_R = \braces{\Vec{x} \in \node \colon x_{k} > \tau}$ for some feature $k$ and threshold $\tau$. Let $N(\node)$ be the number of training points from the sample $(\Mat{X}, \Vec{y})$ that reach node $\node$. Define the decision stump function for each internal node (i.e., split) $\node$ as
\begin{align}
    \psi_s(\Vec{x}) = \frac{N\left(\node_{R}\right)\mathbf{1}\braces*{\Vec{x} \in \node_{L}} - N\left(\node_{L}\right)\mathbf{1}\braces*{\Vec{x} \in \node_{R}}}{\sqrt{N\left(\node_{L}\right)N\left(\node_{R}\right)}}.
\end{align}
Each split depends on the training data $(\Mat{X}, \Vec{y})$;  we suppress this dependence for simplicity of notation.  The function $\psi$ takes three values:  positive if sample $\Vec{x}$ was sent to child node $\node_L$, negative if it was sent to $\node_R$, and 0 if it did not reach node $\node$ at all.   Assuming the tree $\tree$ consists of $m_{\tree}$ splits $\braces{s_1, \ldots, s_{m_{\tree}}}$, concatenating all their decision functions yields the feature map $\Psi_{\tree}(\Vec{x})   
= (\psi_{s_1}(\Vec{x}), \ldots, \psi_{s_{m_{\tree}}}(\Vec{x}))$ and the corresponding transformed training data $\Psi_{\tree}(\Mat{X})\in \R^{n \times m_{\tree}}$ to be plugged into \eqref{eq:connection_linear_trees}.

\citet{agarwal2023mdi+} expanded upon this linear regression interpretation of decision trees, proposing \rfplus~and demonstrating its  improved prediction performance over RFs in a variety of real data applications.
In short, \rfplus~generalizes \eqref{eq:connection_linear_trees} to allow for 
different loss functions (e.g., logistic loss for classification) while also adding non-stump features (e.g., original features used in at least one tree) 
and  regularization penalties on the coefficients, tuned via a computationally efficient leave-one-out cross validation. Formally, given a fitted decision tree $\tree$, the predictions from each ``tree'' in \rfplus~are obtained by combining linear models and nonlinear decision trees via
\begin{align}
    \hat{\Vec{y}}^{(\tree)} = \Mat{X} \hat{\Vec{\beta}}^{(\tree)} + \Psi_{\tree}(\Mat{X}) \hat{\Vec{\gamma}}^{(\tree)},
\end{align}
where
\begin{align}\label{eq:rfplus}
    \hat{\Vec{\beta}}^{(\tree)}, \hat{\Vec{\gamma}}^{(\tree)} = 
    \argmin_{\Vec{\beta} \in \R^p, \; \Vec{\gamma} \in \R^{m_{\tree}}} \;\; \ell(\Vec{y}, \Mat{X} \Vec{\beta} + \Psi_{\tree}(\Mat{X}) \Vec{\gamma}) + P_{\Vec{\beta}}(\Vec{\beta}) + P_{\Vec{\gamma}}(\Vec{\gamma}).
\end{align}
Here, $\ell$ is a user-specified loss function, 
and $P_{\Vec{\beta}}$ and $P_{\Vec{\gamma}}$ are optional penalties on the regression coefficients $\Vec{\beta}$ and $\Vec{\gamma}$, respectively (e.g., the lasso $\ell_1$ or the ridge $\ell_2$ penalties).

This extension of a decision tree is related to but differs from another extension known as ``model trees'' \citep{quinlan1992learning,chaudhuri1994piecewise,kunzel2019linear}. 
Model trees fit flexible local models (e.g., linear or piecewise-polynomial regressions) in each leaf node of the tree, instead of averaging the training responses per leaf node as in standard decision trees.
Similar to model trees, \rfplus~allows for a more flexible model to be learned in conjunction with the tree.
However, while model trees fit each of their local models separately \textit{per} leaf node, \rfplus~fits a single model across \textit{all} leaf nodes, enabling joint learning of the global linear coefficients while still allowing the decision stump features $\Psi_{\tree}(\Mat{X})$ to capture local non-linear effects.

An \rfplus~is then simply an ensemble of $T$ of these trees, which predicts by averaging over the trees: 
\begin{align}
    \hat{\Vec{y}} = \frac{1}{T} \sum_{t=1}^T \hat{\Vec{y}}^{(\tree)} 
    = \frac{1}{T} \sum_{t=1}^T \left( \Mat{X} \hat{\Vec{\beta}}^{(\tree)} + \Psi_{\tree}(\Mat{X}) \hat{\Vec{\gamma}}^{(\tree)} \right).
\end{align}

\subsection{Network-assisted RF$\plus$ (\method) Prediction Models}\label{subsec:nerf}

Beyond its flexible form and improved prediction over RF, an attractive property of \rfplus~when it comes to incorporating the network information is that it is amenable to both network-assisted approaches discussed above: it can accommodate an expanded feature matrix $\tilde{\Mat{X}} = [\Mat{X}, \Mat{Z}]$, where $\Mat{Z}$ is learned from the network $\Mat{A}$, and it can be regularized by adding a network cohesion-type penalty. One may then ask which of these two approaches is ``better'':  we argue that, as usual, the answer depends on the data. Some problems may exhibit a smoothness in responses between connected nodes that the network cohesion penalty is designed to pick up; in others, responses may exhibit more similarity for nodes close together in the learned latent embedding space $\Mat{Z}$, which may or may not be reflected in the neighborhood smoothness encouraged by the network cohesion penalty. These two approaches are not mutually exclusive, and there is no need to choose: the most flexible approach is to allow the data to speak for itself.

With that in mind, we introduce \textbf{\method}, a family of \textbf{Ne}twork-assisted \textbf{\rfplus}~prediction models that incorporates network information through both network embedding features $\Mat{Z}$ and a network cohesion penalty. Given an observed feature data matrix $\Mat{X} \in \R^{n \times p}$, graph adjacency matrix $\bA \in \R^{n \times n}$, response $\Vec{y} \in \R^n$, and  network embedding features $\Mat{Z} \in \R^{n \times r}$ which are learned from $\Mat{A}$, we fit \method~according to the following three-step procedure:
\begin{enumerate}
    \item Train an RF with $T$ trees to predict $\Vec{y}$ from $\tilde{\Mat{X}} = [\Mat{X}, \Mat{Z}] \in \R^{n \times (p + r)}$. 
    \item For each tree $t = 1, \ldots, T$, extract the decision stump feature matrix $\Psi_t(\tilde{\Mat{X}})$ from the fitted RF, and minimize
    \begin{align}\label{eq:nerf}
        \hat{\Vec{\alpha}}^{(\tree)}, \hat{\Vec{\beta}}^{(\tree)}, \hat{\Vec{\gamma}}^{(\tree)} =
        \argmin_{
            \substack{
                \Vec{\alpha} \in \R^n\\
                \Vec{\beta} \in \R^{p + r}\\
                \Vec{\gamma} \in \R^{m_t}
            }
        }\; \ell(\Vec{y}, \Vec{\alpha} + \tilde{\Mat{X}} \Vec{\beta} + \Psi_t(\tilde{\Mat{X}}) \Vec{\gamma}) + \lambda_{\alpha} \Vec{\alpha}^{\top} \bL \Vec{\alpha} + P_{\Vec{\beta}}(\Vec{\beta}) + P_{\Vec{\gamma}}(\Vec{\gamma}),
    \end{align}
    where again $\ell$ is a user-specified loss function, $\lambda_{\alpha} > 0$ is the cohesion penalty tuning parameter, $\bL$ is the Laplacian of the network $\bA$, and $P_{\Vec{\beta}}$ and $P_{\Vec{\gamma}}$ are optional penalties on the regression coefficients $\Vec{\beta}$ and $\Vec{\gamma}$, respectively.
    \item Compute predictions as 
    \begin{align}\label{eq:nerf_prediction}
        \hat{\Vec{y}} 
        = \frac{1}{T} \sum_{t=1}^T \left( \hat{\Vec{\alpha}}^{(\tree)} + \tilde{\Mat{X}} \hat{\Vec{\beta}}^{(\tree)} + \Psi_{\tree}(\tilde{\Mat{X}}) \hat{\Vec{\gamma}}^{(\tree)} \right).
    \end{align}
\end{enumerate}
As before, we can interpret \method~as an ensemble of $T$ trees, where each tree $t$ is a sum of individual node effects $\hat{\Vec{\alpha}}^{(\tree)}$, reflecting smoothness in the responses between connected nodes in the network, and  linear $\tilde{\Mat{X}} \hat{\Vec{\beta}}^{(\tree)}$ and nonlinear effects $\Psi_{\tree}(\tilde{\Mat{X}}) \hat{\Vec{\gamma}}^{(\tree)}$ of the original features $\Mat{X}$ and the network embedding features $\Mat{Z}$ on the response $\Vec{y}$.

In this form,  \method~implicitly incurs some ``double-dipping'' since the responses $\Vec{y}$ are used to both learn the decision stumps $\Psi_{\tree}$ and the coefficient estimates $\hat{\Vec{\alpha}}^{(\tree)}, \hat{\Vec{\beta}}^{(\tree)}$, and $\hat{\Vec{\gamma}}^{(\tree)}$.
Though this would pose a serious concern if we were interested in performing statistical inference on the model coefficients, we do not tackle such inferential tasks in this paper. 
For prediction tasks, this double-dipping is less of a concern since the primary goal of maximizing predictive performance on unseen data can be unbiasedly assessed via a held-out test set, regardless of how the model was trained.
At the same time, as seen with \rfplus~in \citet{agarwal2023mdi+}, the regularization penalties in \method~can help to mitigate potential overfitting, and we generally find that \method~yields better predictive performance when \eqref{eq:nerf} is fitted using the full training data $(\tilde{\Mat{X}}, \Vec{y})$ instead of the out-of-bag data.
We hence use this approach by default.
Nonetheless, there may be situations where one would like to be conservative with potential overfitting and avoid double-dipping altogether. In this case, one would fit \eqref{eq:nerf} using only the out-of-bag (OOB) samples for each tree $t$ since each tree $t$ in the RF was originally trained using a bootstrap (i.e., in-bag) sample of the training data. 

One may wonder whether it is necessary to incorporate network information through both the node effect $\Vec{\alpha}$ and the embedding features $\Mat{Z}$. It is certainly possible to simplify \method~to include only one of these components --- for example, if there is domain knowledge indicating one approach is much more appropriate than the other. We stress, however, that they can potentially capture different types of network effects: the node effects $\Vec{\alpha}$ capture cohesion in the observed network (nodes with a short network path between them are similar) whereas the embedding features $\Mat{Z}$ capture cohesion in the latent space (nodes close to each other in the latent space are similar). Depending on the network model and the relationship between the latent space distance and probability of edges, these two types of effects may be similar or very different (explored further in Sections~\ref{sec:sims} and \ref{sec:case_study}). Including both allows for maximum flexibility in prediction, at the expense of having to tune additional hyperparameters (discussed in Section~\ref{subsec:nerf_more}).

Finally, we highlight that \method~can be generalized to settings where multiple observations (e.g., repeated measurements) are observed at each node in the network. 
For such cases,
it is natural to have all observations belonging to the same node share the same node effect $\alpha$ and node embedding $\Vec{z}$.  Formally, if we have $n$ observations belonging to $q$ nodes in the network, the \method~optimization problem \eqref{eq:nerf} can be replaced with 
\begin{align*}
    \hat{\Vec{\alpha}}^{(\tree)}, \hat{\Vec{\beta}}^{(\tree)}, \hat{\Vec{\gamma}}^{(\tree)} =
    \argmin_{
        \substack{
            \Vec{\alpha} \in \R^q\\
            \Vec{\beta} \in \R^{p + r}\\
            \Vec{\gamma} \in \R^{m_t}
        }
    }\; \ell(\Vec{y}, \Mat{Q} \Vec{\alpha} + \tilde{\Mat{X}} \Vec{\beta} + \Psi_t(\tilde{\Mat{X}}) \Vec{\gamma}) + \lambda_{\alpha} \Vec{\alpha}^{\top} \bL \Vec{\alpha} + P_{\Vec{\beta}}(\Vec{\beta}) + P_{\Vec{\gamma}}(\Vec{\gamma}),
\end{align*}
where $\Mat{Q}$ is an $n \times q$ indicator matrix such that $Q_{ij} = 1$ if observation $i$ belongs to node $j$ and $0$ otherwise, $\tilde{\Mat{X}} = [\Mat{X}, \Mat{Q} \Mat{Z}] \in \R^{n \times (p + r)}$, and $\Mat{Z} \in \R^{q \times r}$ is the network embedding matrix for the $q$ nodes in the network.
This generalization is particularly useful in spatiotemporal applications (as we will see in Section~\ref{subsec:case_study_philly_crime}), where each node in the network corresponds to a spatial location and multiple observations are collected at each location over time.
For simplicity of notation, we focus on the case of one observation per node in the remainder of the paper.

\subsection{Prediction for test samples}\label{subsec:predictions}

Suppose now we want to make predictions on new test data points, which were not available at training time.
Suppose also that we have access to the network containing both the connections within the test set and between the test and the training sets at test time (but not at training time, when we only have the connections within the training set available).
To make predictions on the test points using \method, we will need estimates of both the network embedding features and individual node effects for the new test points. 

To estimate the network embedding features for these test points, we leverage existing out-of-sample extensions of the training embedding approach (e.g., using \citet{levin2018out} for Laplacian and adjacency spectral embedding approaches). These approaches are designed to embed new test points into the previously-trained embedding space. If the network embedding model does not admit an out-of-sample extension or allow for inductive prediction, \method~similarly cannot be used to make predictions on new test nodes. 

To estimate the individual node effects for the new test points, note that we cannot directly optimize \eqref{eq:nerf} with the test data since the test responses $\Vec{y}$ are unknown.
We thus follow the RNC approach \citep{li2019prediction}, which leverages cohesion to estimate the test node effects.

Formally, for a given tree $t$, let $\hat{\Vec{\alpha}}_1^{(\tree)}$ denote the training node effects for the $n_1 = n$  training points. To estimate the individual node effects $\hat{\Vec{\alpha}}_2^{(\tree)}$ for the $n_2$ new test points, we look for the maximally cohesive effects, minimizing 
\begin{align*}
    \hat{\Vec{\alpha}}_2^{(\tree)}  = \argmin_{\Vec{\alpha}_2 \in \R^{n_2}} (\hat{\Vec{\alpha}}_1^{(\tree)}, \Vec{\alpha}_2)^{\top} \Mat{L}' (\hat{\Vec{\alpha}}_1^{(\tree)}, \Vec{\alpha}_2),
\end{align*}
where $\Mat{L}'$ is the Laplacian corresponding to the network with the $n_1 + n_2$ nodes combining the training and test data, written as
\begin{align*}
    \Mat{L}' = 
    \begin{bmatrix}
        \Mat{L}_{11} & \Mat{L}_{12} \\
        \Mat{L}_{21} & \Mat{L}_{22}
    \end{bmatrix}.
\end{align*}
Here, $\Mat{L}_{11} = \Mat{L}$ corresponds to the $n_1$ training nodes and $\Mat{L}_{22}$ corresponds to the $n_2$~test~nodes. 

The estimated test network embedding features and test node effects $\hat{\Vec{\alpha}}_2^{(\tree)}$ can then be plugged into \eqref{eq:nerf_prediction} to obtain the test predictions.
One can also obtain valid prediction intervals for these test predictions using the split-conformal approach from \citet{lunde2023conformal}, though this is not the focus of our paper. We provide additional details on these prediction intervals and show that they both achieve the desired coverage and have similar widths relative to comparison methods in Appendix~\ref{app:conformal}.

\subsection{Choosing the network embeddings, penalties, and regularization parameters}\label{subsec:nerf_more}

 The \method~framework has the flexibility to make several modeling choices; that allows us to tailor \method~to the characteristics of the data at hand.

For constructing the network embedding features $\Mat{Z} \in \R^{n \times r}$, we default to the Laplacian spectral embedding given its popularity and reliability,  but any network embedding method which admits an out-of-sample extension for test data (i.e., inductive predictions) can be used with \method.
There is a large literature on selecting an appropriate embedding method and the number of latent dimensions $r$ given various network structures \citep[e.g., ][]{chen2018network, tang2018limit, li2020network}.
One can alternatively treat the embedding method, latent dimension $r$, and other network embedding hyperparameters as tuning parameters to be selected via cross-validation (CV).

Beyond the network embedding, we have the option to incorporate regularization penalties $P_{\Vec{\beta}}$ and $P_{\Vec{\gamma}}$ in \method, which may be helpful if there is a large number of parameters (e.g., if the tree $t$ is deep, or if $p$ is large) and/or strong correlations between the raw features $\tilde{\Mat{X}}$ and the decision stump features $\Psi_{\tree}(\tilde{\Mat{X}})$.   Numerous penalties have been designed for linear regression, with the most popular being the $\ell_1$ (LASSO) \citep{tibshirani1996regression}, $\ell_2$ (ridge) \citep{hoerl1970ridge}, and elastic net penalties \citep{zou2005regularization}. These or any other such regularization penalties can be used in \method~to induce their usual properties (e.g., $\ell_1$ induces sparsity in the estimated coefficients, etc.). Solving \eqref{eq:nerf} with these penalties is straightforward using existing optimization algorithms since the additional network cohesion penalty can be written as a generalized ridge penalty \citep{li2019prediction}. In our experience, the $\ell_2$ penalties (i.e., $P_{\Vec{\beta}}(\Vec{\beta}) = \lambda_{\beta} \norm{\Vec{\beta}}_2^2$ and $P_{\Vec{\gamma}}(\Vec{\gamma}) = \lambda_{\gamma} \norm{\Vec{\gamma}}_2^2$) work very well across a wide range of problems; we use them by default for the remainder of the paper. We also recommend allowing for separate regularization parameters $\lambda_{\beta}$ and $\lambda_{\gamma}$ since  the scales of $\tilde{\Mat{X}}$ and $\Psi_{\tree}(\tilde{\Mat{X}})$ and the relative strength of the linear versus nonlinear components are in general different. 

These regularization parameters $\lambda_{\beta}$ and $\lambda_{\gamma}$ can be selected by conventional cross-validation.   For computational reasons, the CV data splitting is typically applied to $(\tilde{\Mat{X}}, \Vec{y})$ after constructing the network embedding features $\Mat{Z}$ from the full training data. Though this introduces a small amount of data leakage in each held-out CV fold, we will demonstrate that the tuned regularization parameters using this approach yield good generalization performance on new test data in Sections~\ref{sec:sims} and \ref{sec:case_study}.

\section{Interpreting \method}\label{sec:interpret}

\begin{figure}
    \spacingset{1}
    \centering
    \includegraphics[width=1\linewidth]{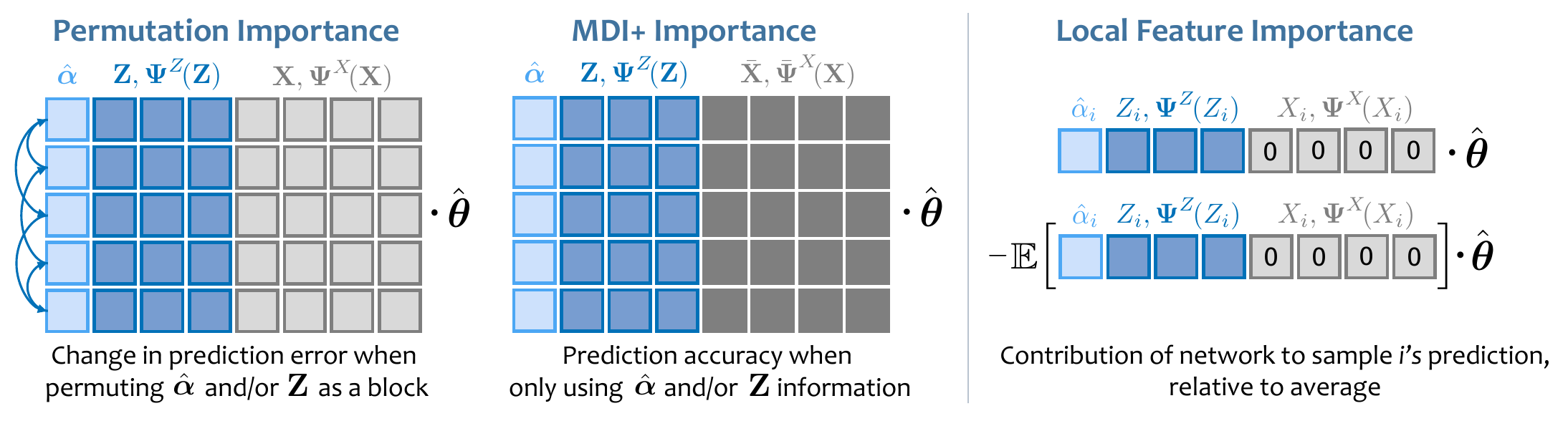}
    \caption{A schematic of global (left) and local (right) network importance measures for \method. To compute the permutation network importance, the network features (blue rows) are permuted. To compute the \mdiplus~network importance, the non-network features are replaced by their column means. To compute the local network importance, sample $i$'s prediction is computed using the network features only and compared to the average network contribution across the population.}
    \label{fig:importances}
\end{figure}

A key advantage of \method~is its dual nature, bridging two simple models: linear regression and RFs. 
This enables us to develop a suite of interpretability tools based on model-intrinsic measures that are derived directly from the fitted model parameters and structure.
While there is substantial literature discussing different definitions and criteria for what ``interpretability'' means in machine learning \citep[e.g.,][]{doshi2017towards, lipton2018mythos, murdoch2019definitions, molnar2020interpretable, allen2023interpretable}, we aim not to tackle these broader questions here, but instead focus on developing several interpretability tools that are commonly used and accepted by practitioners --- namely, feature importance, both global and local (Figure~\ref{fig:importances}), as well as sample influences for \method.

First, we introduce some additional notation. By construction, predictions made by a fitted \method~model $\fh$ take the linear form
\begin{align}\label{eq:nerf_pred}
    \fh(\tilde{\Mat{X}})
    = \frac{1}{T} \sum_{t = 1}^{T} \tilde{\Psi}_{\tree} (\tilde{\Mat{X}}) \hat{\Vec{\theta}}^{(\tree)}    = \frac{1}{T} \tilde{\Psi}_{\sT}(\tilde{\Mat{X}}) \hat{\Vec{\theta}} ,
\end{align}
where $(\hat{\Vec{\theta}}^{(\tree)})^{\top} := [1, (\hat{\Vec{\beta}}^{(\tree)})^{\top}, (\hat{\Vec{\gamma}}^{(\tree)})^{\top}]$, $\tilde{\Psi}_{\tree}(\tilde{\Mat{X}}) := [\hat{\Vec{\alpha}}^{(\tree)}, \tilde{\Mat{X}}, \Psi_{\tree}(\tilde{\Mat{X}})]$, $\hat{\Vec{\alpha}}^{(\tree)}$, $\hat{\Vec{\beta}}^{(\tree)}$, and $\hat{\Vec{\gamma}}^{(\tree)}$ are the solutions to \eqref{eq:nerf}, $T$ is the number of trees in the forest,  
$\tilde{\Psi}_{\sT}(\tilde{\Mat{X}}) := [\tilde{\Psi}_{1} (\tilde{\Mat{X}}), \ldots, \tilde{\Psi}_{T} (\tilde{\Mat{X}})]$ and $\hat{\Vec{\theta}}^{\top} := [(\hat{\Vec{\theta}}^{(1)})^{\top}, \ldots, (\hat{\Vec{\theta}}^{(T)})^{\top}]$.
Furthermore, let $\Psi_{\tree}^{k}(\tilde{\Mat{X}})$ denote the subset of columns in $\Psi_{\tree}(\tilde{\Mat{X}})$ corresponding to splits on feature $X_k$, and let $\tilde{\Psi}_{\tree}^{k}(\tilde{\Mat{X}}) = [\Mat{X}_k, \Psi_{\tree}^{k}(\tilde{\Mat{X}})]$. 
Similarly, let $\Psi_{\tree}^{Z}(\tilde{\Mat{X}})$ denote the subset of columns in $\Psi_{\tree}(\tilde{\Mat{X}})$ corresponding to splits on any network embedding feature from $\Mat{Z}$, and let $\tilde{\Psi}_{\tree}^{Z}(\tilde{\Mat{X}}) = [\Mat{Z}, \Psi_{\tree}^{Z}(\tilde{\Mat{X}})]$.

\subsection{Global Feature Importance}\label{subsec:global_feature_importance}

Global feature (or variable) importance is a classic measure of a feature's contribution to the model's predictions across all training observations (hence the term ``global'').  
For linear models, a typical global important measure is the $t$-statistic associated with the estimated coefficient in ordinary least squares.    
For RFs, mean decrease in impurity (MDI)  \citep{breiman1984classification} and permutation importance \citep{breiman2001random} are the most commonly used global feature importance measures.
Briefly,  MDI quantifies a feature's importance by measuring the decrease in node impurity across splits that were made using that particular feature. Permutation importance measures the change in the model's prediction performance when the feature's values are randomly permuted across all training samples. 

Despite their popularity in practice, both approaches have well-known shortcomings, such as MDI's bias towards high entropy and uncorrelated features \citep{strobl2007bias, nicodemus2011stability} and permutation importance's forced extrapolation and unpredictable performance in the presence of correlated features \citep{hooker2021unrestricted}. Many variants have since been proposed to help mitigate these challenges \citep[e.g.,][]{sandri2008bias, strobl2008conditional, li2019debiased, zhou2021unbiased, loecher2022unbiased, agarwal2023mdi+}. Similar ideas can be adapted to the global feature importance measures we develop below; however, we defer this to future work and focus first on developing the core permutation- and MDI-based global importance measures for \method.

Through the development of these global feature importance methods, we highlight a unique and key task of interpreting network-assisted prediction models --- that is, disentangling importance of individual node features from the importance of their network connections, which \method~can handle in a natural and intuitive way. 

\paragraph{Permutation importance for \method.} Similar to RF permutation importance \citep{breiman2001random}, the permutation importance for \method~can be measured by randomly permuting a feature's values across samples and measuring the resulting change in the model's accuracy. Specifically, given a fitted \method~model $\fh$, scoring metric $\sL$, and $B$ permutations, we compute the global permutation importance of feature $X_k$, denoted $\permimp{\fh}{k}$, as 
\begin{align}\label{eq:perm_imp}
    \permimp{\fh}{k}(\tilde{\Mat{X}}, \Vec{y}) = \frac{1}{B} \sum_{b = 1}^{B} \underbrace{\sL \left(\Vec{y}, \frac{1}{T} \tilde{\Psi}_{\sT}^{\setminus k, b}(\tilde{\Mat{X}}) \hat{\Vec{\theta}} \right)}_{\substack{\text{prediction performance using}\\\text{permuted data matrix}}} - \underbrace{\sL \left(\Vec{y}, \frac{1}{T} \tilde{\Psi}_{\sT}(\tilde{\Mat{X}}) \hat{\Vec{\theta}} \right)}_{\substack{\text{prediction performance using}\\\text{original data matrix}}},
\end{align}
where $\tilde{\Psi}_{\sT}^{\setminus k, b}(\tilde{\Mat{X}})$ is the matrix $\tilde{\Psi}_{\sT}(\tilde{\Mat{X}})$ after the rows of $[\tilde{\Psi}_{1}^{k}(\tilde{\Mat{X}}), \ldots, \tilde{\Psi}_{T}^{k}(\tilde{\Mat{X}})]$ have been randomly permuted. In other words, \eqref{eq:perm_imp} is the change in prediction performance when we jointly permute all features involving $X_k$, both linear and those resulting from the tree.  
The scoring metric $\sL$ is usually taken to be a dissimilarity measure (e.g., root mean squared error or classification error), and thus a higher $\permimp{\fh}{k}$ indicates a more important feature. In practice, it is also generally recommended to evaluate this permutation importance using held-out data \citep{fisher2019all, molnar2020interpretable}.

While permutation importance is a well-established idea, the opportunity to naturally extend it to network importance is unique to \method.   For general network-assisted methods, it is often unclear how to permute a network; one could, for example, rewire a random subset of all edges, rewire edges of a random subset of nodes, or use more sophisticated schemes, and it is unclear what impact the choice of permutation scheme would have on the importance measure. In contrast, to permute the network in \method, we can directly permute the values of the network-related features $\Vec{\alpha}^{(\tree)}$ and $\tilde{\Psi}_{\tree}^{Z}(\tilde{\Mat{X}})$. We thus compute the global permutation importance of the network, denoted $\permimp{\fh}{(\alpha, Z)}$, 
as
\begin{align}
    \permimp{\fh}{(\alpha, Z)}(\tilde{\Mat{X}}, \Vec{y}) = \frac{1}{B} \sum_{b = 1}^{B} \sL \left(\Vec{y}, \frac{1}{T} \tilde{\Psi}_{\sT}^{\setminus (\alpha, Z), b}(\tilde{\Mat{X}}) \hat{\Vec{\theta}} \right) - \sL \left(\Vec{y}, \frac{1}{T} \tilde{\Psi}_{\sT}(\tilde{\Mat{X}}) \hat{\Vec{\theta}} \right),
\end{align}
where $\tilde{\Psi}_{\sT}^{\setminus (\alpha, Z), b}(\tilde{\Mat{X}})$ is the matrix $\tilde{\Psi}_{\sT}(\tilde{\Mat{X}})$ after the rows of $[\hat{\Vec{\alpha}}^{(1)}, \tilde{\Psi}_{1}^{Z}(\tilde{\Mat{X}}), \ldots, \hat{\Vec{\alpha}}^{(T)}, \tilde{\Psi}_{T}^{Z}(\tilde{\Mat{X}})]$ have been permuted. Furthermore, we can separately assess the importance stemming from network cohesion, denoted $\permimp{\fh}{\alpha}$, and from the network embedding, denoted $\permimp{\fh}{Z}$, by respectively permuting the $[\hat{\Vec{\alpha}}^{(1)}, \ldots, \hat{\Vec{\alpha}}^{(T)}]$ or $[\tilde{\Psi}_{1}^{Z}(\tilde{\Mat{X}}), \ldots, \tilde{\Psi}_{T}^{Z}(\tilde{\Mat{X}})]$ blocks separately.

\paragraph{\mdiplus~importance for \method.} 
Since \rfplus~is no longer a purely tree-based model, the popular MDI measure for RF is not directly applicable to \rfplus. \citet{agarwal2023mdi+} thus built upon the linear regression interpretation of decision trees to develop a generalization of MDI called \mdiplus~--- a feature importance framework which is equivalent to MDI when applied to an ordinary RF but is more generally applicable to \rfplus.
Specifically, the \mdiplus~for a feature $X_k$ in \rfplus~is measured by evaluating the similarity between $\Vec{y}$ and the \rfplus~predictions using the data matrix, where all features that do not involve (or split on) $X_k$ have been imputed by their column means. This framework can be extended to define the \mdiplus~importance of feature $X_k$ in a \method~model $\fh$ as 
\begin{align}\label{eq:mdiplus}
    \mdi{\fh}{k}(\tilde{\Mat{X}}, \Vec{y}) 
    = \frac{1}{T} \sum_{t = 1}^{T} m \big( 
        \Vec{y}, \;
        \underbrace{\bar{\tilde{\Psi}}^{\setminus k}_{\tree}(\tilde{\Mat{X}}) \hat{\Vec{\theta}}^{(\tree)}}_{\substack{k^{th} \text{partial model}\\\text{predictions}}} 
    \big)
\end{align}
where $m$ is some similarity metric (e.g., $R^2$ for regression problems and negative log-loss for classification problems), $\hat{\Vec{\theta}}^{(\tree)}$ is that used in \eqref{eq:nerf_pred}, and $\bar{\tilde{\Psi}}^{\setminus k}_{\tree}(\tilde{\Mat{X}})$ is the matrix $\tilde{\Psi}_{\tree}(\tilde{\Mat{X}})$ but with all columns corresponding to features that do not involve (or split on) $X_k$ replaced by their column means, i.e., $$\bar{\tilde{\Psi}}^{\setminus k}_{\tree}(\tilde{\Mat{X}}) = [ \bar{\hat{\Vec{\alpha}}}^{(\tree)}, \bar{\tilde{\Psi}}_{\tree}^{Z}(\tilde{\Mat{X}}), \bar{\tilde{\Psi}}_{\tree}^{1}(\tilde{\Mat{X}}), \ldots, \bar{\tilde{\Psi}}_{\tree}^{k-1}(\tilde{\Mat{X}}), \tilde{\Psi}_{\tree}^{k}(\tilde{\Mat{X}}), \bar{\tilde{\Psi}}_{\tree}^{k+1}(\tilde{\Mat{X}}), \ldots, \bar{\tilde{\Psi}}_{\tree}^{p}(\tilde{\Mat{X}}) ] . $$
Note that as in linear regression, categorical features in \method~are dummy-coded and thus imputed using the mean of the dummy-coded features.
Intuitively, the $k^{th}$ partial model prediction is the prediction we would have gotten had we only had access to the information from feature $X_k$ and ignored all other features (by setting them to their column means).
If the $k^{th}$ partial model prediction is close to the observed response $\Vec{y}$, as measured by $m$, then feature $k$ is considered important and is reflected in a high \mdiplus~value.

To assess the global \mdiplus~importance of the network in \method, we further define
\begin{align}
    \mdi{\fh}{(\alpha, Z)}(\tilde{\Mat{X}}, \Vec{y}) = \frac{1}{T} \sum_{t = 1}^{T} m \big( 
        \Vec{y}, \;
        \underbrace{\bar{\tilde{\Psi}}^{\setminus (\alpha, Z)}_{\tree}(\tilde{\Mat{X}}) \hat{\Vec{\theta}}^{(\tree)}}_{\substack{\text{network partial model}\\\text{predictions}}} 
    \big)
\end{align}
where $\bar{\tilde{\Psi}}^{\setminus (\alpha, Z)}_{\tree}(\tilde{\Mat{X}})$ is the matrix $\tilde{\Psi}_{\tree}(\tilde{\Mat{X}})$ but with all columns corresponding to features that do not involve (or split on) the network replaced by their column means, i.e., $$\bar{\tilde{\Psi}}^{\setminus (\alpha, Z)}_{\tree}(\tilde{\Mat{X}}) = [ \hat{\Vec{\alpha}}^{(\tree)}, \tilde{\Psi}_{\tree}^{Z}(\tilde{\Mat{X}}), \bar{\tilde{\Psi}}_{\tree}^{1}(\tilde{\Mat{X}}), \ldots, \bar{\tilde{\Psi}}_{\tree}^{p}(\tilde{\Mat{X}}) ] . $$
The network partial model predictions can similarly be interpreted as the best we can do using only the network information, ignoring the contributions of all other non-network features.  A high value suggests that the model's predictions are driven primarily by the network information, and the network is important. We can again separate the contributions of network cohesion $\mdi{\fh}{\alpha}$ and the network embedding $\mdi{\fh}{Z}$ by computing the partial model predictions using only the $\hat{\Vec{\alpha}}^{(\tree)}$'s or only the $\tilde{\Psi}_{\tree}^{Z}(\tilde{\Mat{X}})$'s.

\subsection{Local Feature Importance}\label{subsec:local_feature_importance}

Moving beyond global feature importance, local importance measures have been proposed to quantify how much each feature contributes to each sample point's prediction. Arguably the most popular local feature importance measures are based on Shapley values \citep{shapley1953value} from cooperative game theory. Exact computation of Shapley values, however, is intractable for most machine learning models and necessitates approximations --- most notably, via SHAP \citep{lundberg2017unified} or variants thereof, e.g., treeSHAP \citep{lundberg2020local}. However, linear models are an exception. For a given linear model, Shapley values can be directly read off from the summands corresponding to each feature (see \citet{molnar2020interpretable}, Chapter 17). Similar to \citet{liang2025local}, we exploit the convenient linear form of \method~to define the local feature importance of feature $k$ for sample $i$ in the \method~model $\fh$ as
\begin{align}
    \lfi{\fh}{k}(\tilde{\Vec{x}}_i) = \frac{1}{T} \sum_{t = 1}^{T} \left( \tilde{\Psi}_{\tree}^{k}(\tilde{\Vec{x}}_i) \hat{\Vec{\theta}}^{(\tree)}_k - \E_n \left[ \tilde{\Psi}_{\tree}^{k}(\tilde{X}) \right] \hat{\Vec{\theta}}^{(\tree)}_k \right),
\end{align}
where $\hat{\Vec{\theta}}^{(\tree)}_k$ is the subset of coefficients in $\hat{\Vec{\theta}}^{(\tree)}$ that correspond to $\tilde{\Psi}_{\tree}^{k}(\tilde{\Vec{x}}_i)$, and the expectation is taken over the empirical distribution of the training sample.
The first term in the summation is simply the additive portion of the total prediction $\hat{y}_i$ coming from features involving $X_k$ while the second term is a normalization term. There are many possible normalization choices; this choice allows us to interpret $\lfi{\fh}{k}(\tilde{\Vec{x}}_i)$ as the contribution of $X_k$ to sample $i$'s prediction relative to the average contribution of $X_k$ in the training sample. Importantly, the magnitude indicates the strength of the importance while the sign indicates the direction of the importance (i.e., whether the feature contributes to a higher or lower predicted value compared to the average).

Similarly, we define the local network importance for sample $i$ in the \method~model $\fh$~as
\begin{align}\label{eq:lfi_network}
    \lfi{\fh}{(\alpha, Z)}(\tilde{\Vec{x}}_i) = \frac{1}{T} \sum_{t = 1}^{T} \left( \hat{\alpha}^{(\tree)}_i + \tilde{\Psi}_{\tree}^{Z}(\tilde{\Vec{x}}_i) \hat{\Vec{\theta}}^{(\tree)}_Z - \E[\hat{\Vec{\alpha}}^{(\tree)}] - \E \left[ \tilde{\Psi}_{\tree}^{Z}(\tilde{X}) \right] \hat{\Vec{\theta}}^{(\tree)}_Z \right),
\end{align}
where $\hat{\Vec{\theta}}^{(\tree)}_Z$ is the subset of coefficients in $\hat{\Vec{\theta}}^{(\tree)}$ that correspond to $\tilde{\Psi}_{\tree}^{Z}(\tilde{\Vec{x}}_i)$. As before, we interpret this quantity as the difference between the network's contribution to sample $i$'s prediction and the average of such contributions across all samples.  Further,  \eqref{eq:lfi_network} can be again separated into two components: $\lfi{\fh}{\alpha}(\tilde{\Vec{x}}_i) = \sum_{t = 1}^{T} \left( \hat{\alpha}^{(\tree)}_i - \E[\hat{\Vec{\alpha}}^{(\tree)}] \right)$ giving the local importance of network cohesion and $\lfi{\fh}{Z}(\tilde{\Vec{x}}_i) = \sum_{t = 1}^{T} \left( \tilde{\Psi}_{\tree}^{Z}(\tilde{\Vec{x}}_i) \hat{\Vec{\theta}}^{(\tree)}_Z - \E \left[ \tilde{\Psi}_{\tree}^{Z}(\tilde{X}) \right] \hat{\Vec{\theta}}^{(\tree)}_Z \right)$ giving the local importance of the network embedding.

\subsection{Sample Influence} \label{subsec:influences}

To complement the feature importance measures, we lastly develop sample influence scores for \method, which quantify the change in model fit or predictions when leaving out each sample point during training. Sample influences are typically challenging to compute for most modern machine learning settings given that a brute-force approach would require re-fitting the model for each held-out sample. 
While techniques such as those developed by \citet{koh2017understanding} have made it possible to compute sample influences for models with twice-differentiable loss functions without the need to re-fit the model, these techniques cannot be directly applied to tree-based models such as \method, where the tree splits are inherently discrete and non-differentiable. 
However, the linear formulation of \method~provides a unique opportunity to leverage the extensive classical statistical literature on sample influences \citep{hampel1974influence, cook1977detection, cook1982residuals}.
In particular, using the linear formulation of \method, if we keep all features (from the constructed trees $\Psi_t(\tilde{\Mat{X}})$ and the network embeddings $\Mat{Z}$) fixed, we can derive closed-form solutions for approximate leave-one-out (LOO) parameter estimates for \method. Since each tree in \method~is fit independently, we provide this LOO derivation for a single tree in \method~for simplicity.

Let $\hat{\Vec{\nu}}^{\top} = [\hat{\Vec{\alpha}}^{\top}, \hat{\Vec{\beta}}^{\top}, \hat{\Vec{\gamma}}^{\top}]$ be the estimate from \eqref{eq:nerf} for tree $t$, fitted on $n$ training samples. For now, consider the \method~model with $\ell_2$ penalties for $P_{\beta}$ and $P_{\gamma}$. For a fixed tree $t$, if we leave out the $i^{th}$ sample from the training data and re-fit the \method~coefficients, then the resulting estimate $\hat{\Vec{\nu}}^{(-i)}$ can be efficiently computed for all samples $i = 1, \dots, n$ via:
\begin{align}
    \hat{\Vec{\nu}}^{(-i)}
    = \left[ 
        \hat{\Vec{\nu}}
        + \frac{(\Mat{B}^{(i)})^{-1} \Vec{w}_i}{1 - h_i} (\hat{y}_i - y_i)
        - \frac{\hat{\alpha}_i}{[\Mat{B}^{-1}]_{ii}} \left( 
            [\Mat{B}^{-1}]_{\cdot i} + \frac{(\Mat{B}^{(i)})^{-1} \Vec{w}_i \Vec{w}_i^{\top} [\Mat{B}^{-1}]_{\cdot i}}{1 - h_i}
        \right)
    \right]_{-i},\label{eq:loo}
\end{align}
where
\begin{align*}
    \Mat{W} &= [\Mat{I}_n, \tilde{\Mat{X}}, \Psi_t(\tilde{\Mat{X}})],
    \quad \Mat{M} = \begin{bmatrix} 
        \lambda_{\alpha} \Mat{L} & \Mat{0} & \Mat{0} \\ 
        \Mat{0} & \lambda_{\beta} \Mat{I}_{p + r} & \Mat{0} \\
        \Mat{0} & \Mat{0} & \lambda_{\gamma} \Mat{I}_{m_t}
    \end{bmatrix},
    \quad \Mat{B}^{-1} = ( \Mat{W}^{\top} \Mat{W} + \Mat{M} )^{-1},\\
    \quad (\Mat{B}^{(i)})^{-1} &= \Mat{B}^{-1} - \frac{1}{[\Mat{B}^{-1}]_{ii}} [\Mat{B}^{-1}]_{\cdot i} [\Mat{B}^{-1}]_{\cdot i}^{\top},
    \qquad h_i = \Vec{w}_i^{\top} (\Mat{B}^{(i)})^{-1} \Vec{w}_i.
\end{align*}
This derivation (details in Appendix~\ref{app:sample_influences}) relies on the observation that \eqref{eq:nerf} is a generalized ridge problem. For \method~models with non-ridge penalties, the LOO estimates $\hat{\Vec{\nu}}^{(-i)}$ can be approximated using techniques developed in \citet{koh2017understanding}. These LOO parameter estimates $\hat{\Vec{\nu}}^{(-i)}$ can then be plugged into \eqref{eq:nerf_pred} to efficiently compute the LOO predictions and estimate the prediction error had we left out sample $i$ from the training set. In particular,  we define  influence of a training sample $i$ to be the mean squared difference between the original test predictions (computed using $\hat{\Vec{\nu}}$) and the test predictions when leaving out sample $i$ during training (computed using $\hat{\Vec{\nu}}^{(-i)}$).

Note however \eqref{eq:loo} is not a true LOO estimate, since we do not take into account how leaving out the sample impacts the fitted tree and the resulting stump features $\Psi_t(\tilde{\Mat{X}})$, the network embedding $\Mat{Z}$, the Laplacian matrix $\Mat{L}$ used in the cohesion penalty, or the choice of regularization parameters. 
Despite this limitation, we show in Appendix~\ref{app:sims} that the estimates in \eqref{eq:loo} provide good approximations to the oracle, obtained by leaving out each sample and recomputing everything from scratch (the tree, network embeddings, Laplacian matrix, and regularization parameters).  When computational efficiency is a priority, this closed-form approximation can be a good practical compromise.   

\section{Simulations}\label{sec:sims}

We next demonstrate the strong performance of \method~on both prediction and interpretability across a wide range of simulations. We compare \method~with two network-assisted prediction methods: its linear counterpart, RNC \citep{li2019prediction}, and an alternative tree-based method, Network BART \citep{deshpande2022flexbart}. We also included their non-network-assisted counterparts as baselines: \rfplus~\citep{agarwal2023mdi+}, ordinary linear regression, and BART \citep{chipman2010bart}. Tuning and implementation details are provided in Appendix~\ref{subsec:implementation_details}. Regarding interpretations, we focus specifically on the global feature importances and sample influences in these simulations and leave the investigation of local feature importances to the case studies in Section~\ref{sec:case_study}. 

\paragraph{Simulation Setup.} We generated $n = 300$ samples with $p = 20$ covariates sampled i.i.d.\ from $N(0,1)$ connected by an unweighted, undirected network drawn from a stochastic block model (SBM) with three blocks and edge probabilities 0.2 and 0.02 within and between blocks, respectively. Responses $\Vec{y} \in \R^n$ were then drawn from one of two network effect models, with the function $f$ of covariates to be specified later and noise variables drawn as $\epsilon_i \stackrel{iid}{\sim} N(0, \sigma_{\epsilon}^2)$:
\begin{enumerate}
    \item[(1)] \textit{Additive blockwise network effect}: 
    $
        y_i = \alpha_i + f(\Vec{x}_i) + \epsilon_i,
    $
    where $\alpha_i = -\eta, 0, \eta$ for samples in SBM blocks 1, 2, and 3, respectively. This network effect model is better captured by including network embedding features that reflect the SBM block structure, as opposed to encouraging network cohesion, since the presence or absence of a connection between two nodes within the same block has no effect on their responses' similarity.  
    \item[(2)] \textit{Network autocorrelation effect}:
    $
        y_i = \frac{\omega}{d_i} \sum_{j: A_{ij} = 1} y_j + f(\Vec{x}_i) + \epsilon_i,
    $
    where $d_i$ is the degree for sample $i$. This network effect model creates a direct relationship between a node's response and its neighbors' responses, and thus should benefit greatly from encouraging network cohesion in the fitted  model.
\end{enumerate}
We examined three different forms of $f$:
 \begin{itemize}[noitemsep]
     \item[] \textit{Linear}: $f(\Vec{x}_i) = x_{i1} + x_{i2}$; 
     \item[] \textit{Polynomial}: $f(\Vec{x}_i) = x_{i1} + x_{i1} x_{i2} + x_{i3} + x_{i3} x_{i4} + x_{i5} + x_{i5} x_{i6}$; 
     \item[] \textit{Locally Spiky Sparse (LSS)}: $f(\Vec{x}_i)$ = $\mathbf{1}_{\{x_{i1} > 0\}} \mathbf{1}_{\{x_{i2} > 0\}} + \mathbf{1}_{\{x_{i3} > 0\}} \mathbf{1}_{\{x_{i4} > 0\}} + \mathbf{1}_{\{x_{i5} > 0\}} \mathbf{1}_{\{x_{i6} > 0\}}$, introduced by \cite{behr2021provable}. 
 \end{itemize}
These functional forms capture a wide array of structures, including linear and nonlinear regimes as well as smooth (linear and polynomial) and non-smooth (LSS) scenarios.

We also considered varying levels of signal-to-noise ratio and network effect strength. Specifically, the noise variance $\sigma_{\epsilon}^2$ was chosen such that the proportion of variance explained (PVE) in $\Vec{y}$ by $\Mat{X}$ varied across $\{0.2, 0.4, 0.6, 0.8\}$. To vary the network effect strength, we varied $\eta \in \{0, 0.5, 1, 1.5\}$ in the additive blockwise network effect model and $\omega \in \{0.1, 0.3, 0.5, 0.7, 0.9\}$ in the network autocorrelation effect model. In each simulation, we randomly split the data into training (80\%) and test (20\%) sets. The training set was used to fit the prediction models while the test set was used to evaluate the prediction performance and compute the feature importances \citep{fisher2019all, molnar2020interpretable}. We replicated each simulation scenario 100 times.

\subsection{Prediction Performance Results}\label{subsec:sims-predictions}

\begin{figure}[t]
    \spacingset{1}
    \centering
    \includegraphics[width=\linewidth]{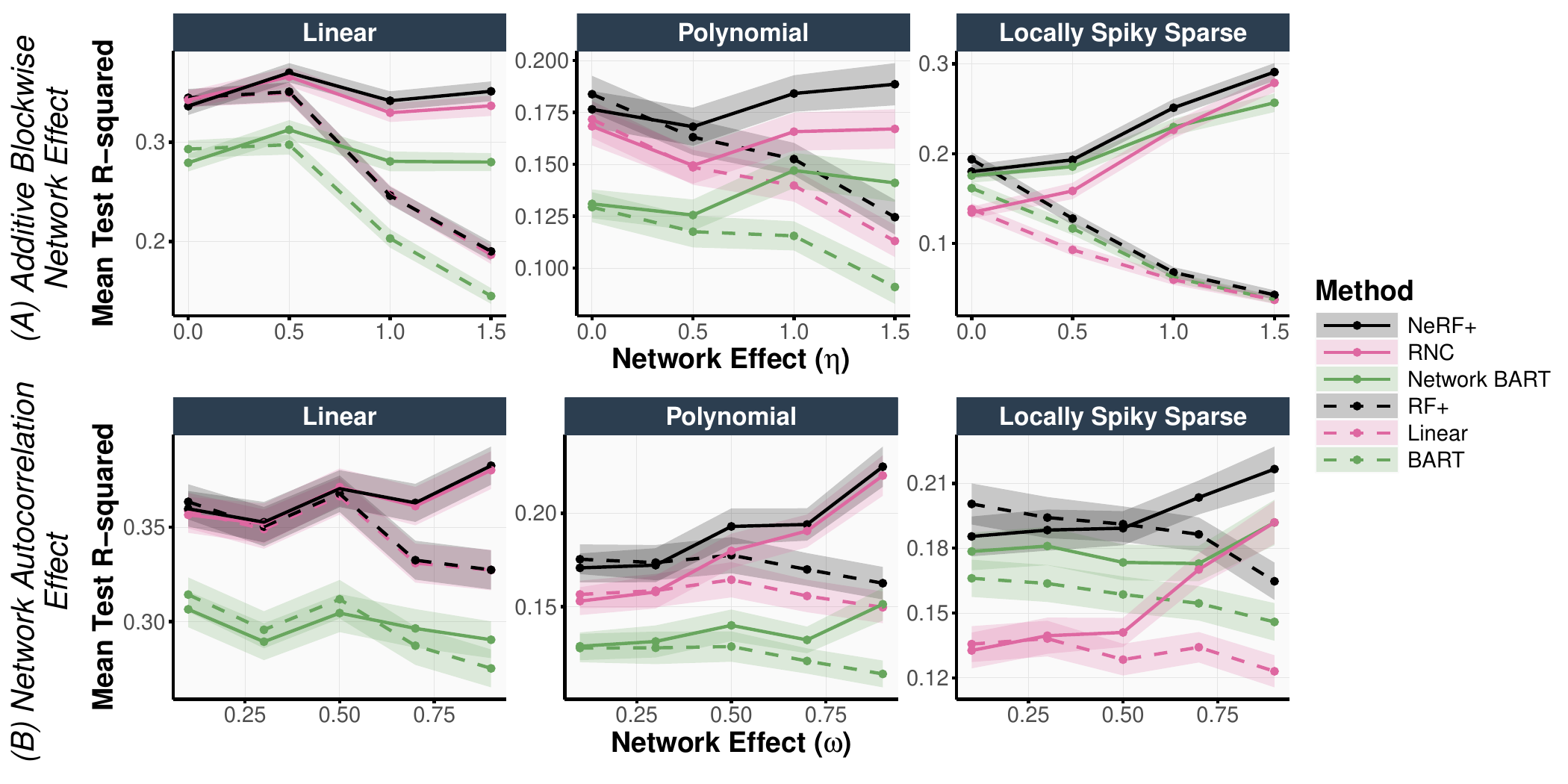}
    \caption{Prediction accuracy measured by the test $R^2$, for different simulated network effect models (rows), underlying functional forms (columns), and varying strengths of the network effect (x-axis) at a moderately-low signal-to-noise ratio ($\mathrm{PVE} = 0.4$). Results are averaged across 100 replications with shaded regions denoting $\pm 1 SE$.}
    \label{fig:sim_predictions}
\end{figure}

Figure~\ref{fig:sim_predictions} summarizes each method's test prediction accuracy, measured via $R^2$ on the test data, when the signal-to-noise was moderately-low ($\mathrm{PVE} = 0.4$).
There are a few general observations we can make. When there is a strong network effect (larger values of $\eta$ and $\omega$), network-assisted methods perform better than their non-network versions (solid vs.\ dashed lines). Moreover, the network-assisted and non-network methods are comparable where there is little influence from the network, suggesting that tuning works as it should to detect the network effect and there is no overfitting happening. When the underlying true model is linear, the linear methods (linear regression and RNC) perform well, as expected, and \method~performs similarly, successfully adapting to the linear functional form.  BART and network BART, on the other hand, both perform poorly on the linear model, a well-known issue with trees.   On the polynomial function, the linear methods start deteriorating relative to \rfplus/\method, but still  outperform their BART-based counterparts.  Finally, in the non-linear, non-smooth setting, RNC does poorly when there is no network effect, but recovers with the strong network effect, as most of the predictive power no longer comes from the features (and therefore the misspecified functional form of their dependence is less relevant).  \method, in contrast, performs very well across the entire range of settings, due to its ability to both leverage the network information and flexibly fit the relationship between the response and covariates. 

Similar patterns, showing that \method~consistently yields improved prediction performance over existing methods, are observed under other signal-to-noise regimes (PVE = $0.2, 0.6, 0.8$), correlated designs, and classification settings; see Appendix~\ref{subsec:main_sims}, \ref{subsec:correlated_sims}, and \ref{subsec:classification_sims} respectively. 
For example, in the classification setting with strong network effects, we, on average, observe a 2--5\% improvement in test accuracy for \method~over the next-best method.
We also demonstrate that the performance of \method~is robust to the choice of network embedding method in Appendix~\ref{subsec:robustness_sims}, yielding similar strong prediction performance across a variety of network embedding choices.
Finally, we report the computational costs of fitting \method~in Appendix~\ref{subsec:runtime_sims}, showing that for datasets of varying sizes up to $n = 1500$, \method~can be fit efficiently in under 3 minutes while Network BART requires over an hour.

\subsection{Global Feature Importance Results}\label{subsec:sims-globalfi}

\begin{figure}[t]
    \spacingset{1}
    \centering
    \includegraphics[width=1\linewidth]{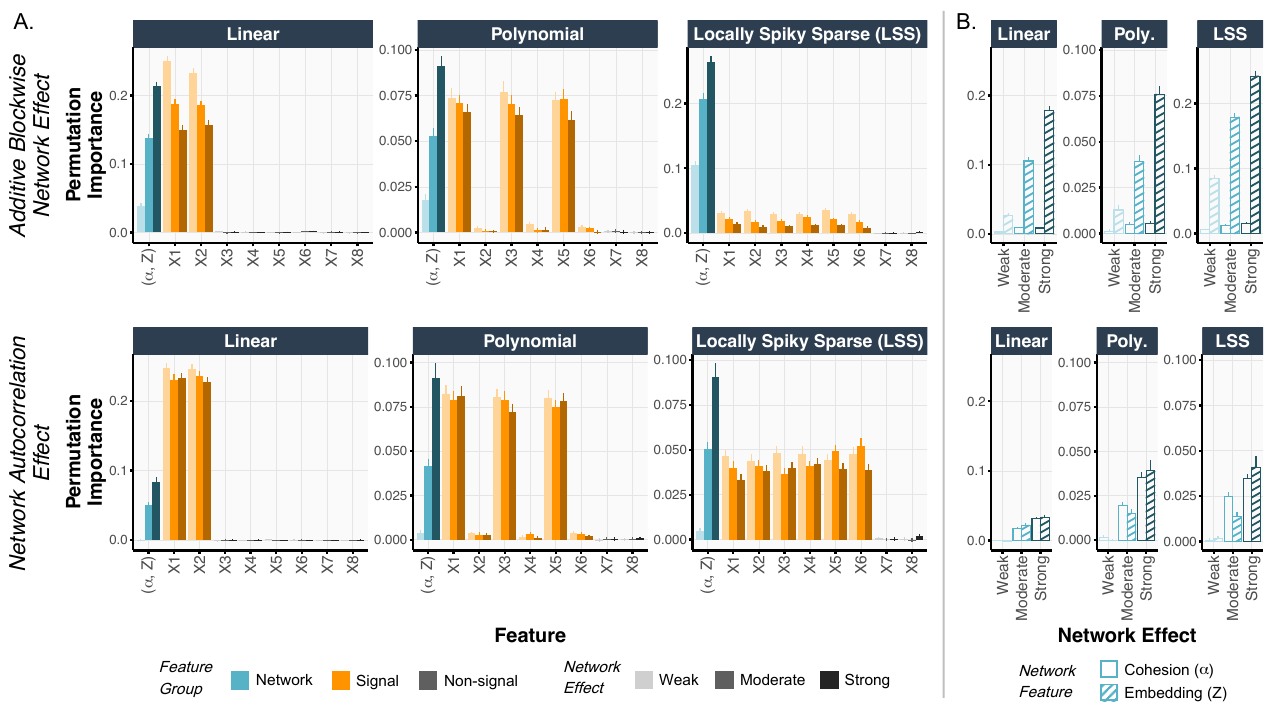}
    \caption{
        \method~global feature importances, measured via permutation importance at a moderately-low signal-to-noise ratio ($\mathrm{PVE} = 0.4$). 
        (A) Signal features exhibit higher importance than the non-signal features, and as the network effect increases (i.e., as the fill shade darkens), the estimated global importance of the network increases. Note: only the first 8 features are shown since the remaining features are non-signal and exhibit similarly low importances as the non-signal features shown. (B) The network embedding features have greater importance under the additive blockwise network effect model compared to the network autocorrelation effect model, while the opposite is true for the network cohesion component. Results are averaged across 100 replicates with error bars denoting $\pm 1$ SE.
    }
    \label{fig:sim_importances_permutation}
\end{figure}

In Figure~\ref{fig:sim_importances_permutation}A, we examine permutation feature importances computed by \method. We see that (1) the signal features in each model are correctly identified as more important than the non-signal features, and (2) the overall network's importance increases as the strength of the network effect increases as desired. The increasing network importance also agrees with the prediction accuracy results (Figure~\ref{fig:sim_predictions}), which show prediction methods that do not use the network perform worse.  

We can also investigate whether the importance of the network stems from the network cohesion $\Vec{\alpha}$ and/or the network embedding features $\Mat{Z}$. As seen in Figure~\ref{fig:sim_importances_permutation}B, the network embedding features have higher importance than the network cohesion in the additive blockwise network effect setting, since the only way the network is affecting the response is through the block membership, and the eigenvectors of the Laplacian capture that structure well.  On the other hand, though network cohesion has little to no importance in the additive blockwise network effect setting, the network cohesion importance is much higher in the network autocorrelation effect setting, since there each node is directly affected by its neighbors, and the global importance measures pick that up correctly.
By allowing both network embedding features and network cohesion, \method~can effectively capture and adapt to different types of network effects while disentangling their respective contributions to the model.  

Similar findings hold for \mdiplus, different PVEs, and when we have correlated features or classification problems; see Appendix~\ref{subsec:main_sims}, \ref{subsec:correlated_sims}, and \ref{subsec:classification_sims} respectively. In Appendix~\ref{subsec:bamdt}, we also examine the global feature importances from a different network-assisted BART model, BAMDT \citep{luo2022bamdt}, as feature importances are not currently implemented for Network BART; this investigation illustrates the current limitations of network-assisted BART models when comparing the importance of the network to that of other predictor features.

\subsection{Sample Influence Results}\label{subsec:sims-influences}

We assess the approximate LOO coefficient estimates from Section~\ref{subsec:influences} according to both (i) how well this approximation matches the oracle LOO coefficients obtained by manually leaving out each sample and refitting \method~(including refitting the trees and network embeddings for each LOO) and (ii) its practical utility for identifying influential samples, such as outliers. On task (i), we observe strong concordance between the approximate and oracle LOO coefficients; see detailed results in Appendix~\ref{subsec:loo_sims}.  Here we report on (ii), assessing the effectiveness of the approximate sample influence measures for detecting outliers.

To this end, we generated data according to the previously described simulation setup and altered the response of a randomly selected training sample to create an outlier.   Specifically, for a randomly selected training sample $i^*$ in each simulation replicate, we added a constant to its response $y_{i^*}$, either adding $\kappa$ times the standard deviation of the response $\Vec{y}$ if $y_{i^*} > 0$, or subtracting this constant if $y_{i^*} \leq 0$.  We varied $\kappa \in \{1, 2, 3, 4\}$ with larger values corresponding to more extreme outliers. Under each scenario, we then computed the approximate leave-one-out (LOO) coefficients using \eqref{eq:loo} and used them to estimate each training sample's influence, defined as the mean squared difference between the original test predictions and the test predictions after leaving out that sample from training.

\begin{figure}
    \spacingset{1}
    \centering
    \includegraphics[width=0.85\linewidth]{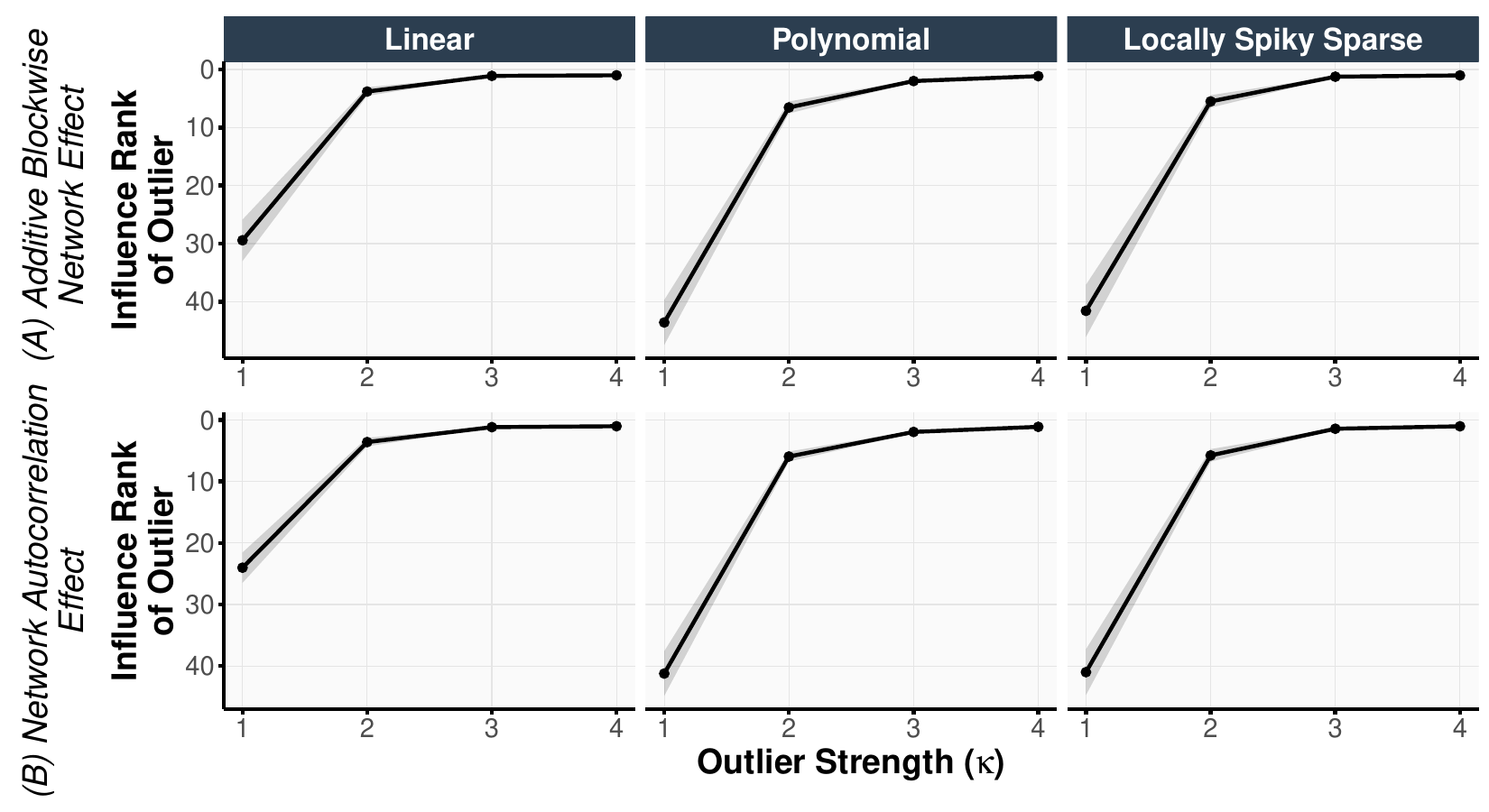}
    \caption{Rank of the outlier's influence (out of 240 total samples) for different simulated network effect models (rows), underlying functional forms (columns), and varying strengths of the outlier (x-axis). Rank 1 corresponds to the most influential sample. Results are averaged across 100 replications with shaded regions denoting $\pm 1$ SE.}
    \label{fig:influences}
\end{figure}

Figure~\ref{fig:influences} shows that as the perturbation strength ($\kappa$) increases, the influence of the outlier also increases, as it should. Moreover, the outlier was almost always estimated to have the highest influence (i.e., rank 1) when $\kappa \geq 3$. Influence measured by using the change in the estimated model parameters behaves similarly;  we omit these results for brevity. Overall, these results confirm the ability of our sample influence measure to identify influential samples, such as outliers, that have a disproportionate impact on the fitted model and its resulting predictions.

\section{Case Studies}\label{sec:case_study}

In this section, we illustrate \method~on two case studies on reducing school conflicts and predicting crime rates in Philadelphia.   While we do not know the mechanism of network influence in case studies, our results are intuitive and illustrate well the types of insights that can be obtained from \method~and the accompanying interpretability tools.  

\subsection{Case study 1:  School conflict intervention}\label{subsec:case_study_school}

\paragraph{Data.}   This dataset comes from an experiment originally conducted to assess the impact of anticonflict intervention workshops on reducing student conflicts within middle schools \citep{paluck2016changing}. In this study, half of the schools were randomly assigned to receive an anticonflict intervention, in which a small number of students (on average 26 per school) were selected to attend biweekly anticonflict trainings throughout the school year. Additionally, as part of the study, all students were asked to fill out surveys at the beginning and end of the school year. These surveys asked about (i) the student's demographics and background, (ii) who the students chose to spend the most time with in the last few weeks (i.e., their friends), and (iii) their perception of their school's ``friendliness''. Setting aside the original question of the study about effectiveness of the anticonflict training intervention, we focus on a prediction question:  can we predict each student's perception of their school's friendliness at the end of the school year from what we know about them at the start of the year, and how much of this perception stems from their social network?  

Following \citet{le2022linear}, we define the response variable as the average of 13 questions (with responses ranging from 0--5) related to the student's perception of their school's friendliness at the end of the school year.  A higher score indicates that the student perceives the school to be more friendly. To predict this end-of-year friendliness score, we use the following covariates: the student's grade, gender, ethnicity, whether or not they live with both parents, whether or not they are a returning student, whether or not they participated in the anticonflict training (i.e., the treatment), and the student's beginning-of-year school friendliness score. We also constructed a friendship network from the survey responses as in previous work \citep{paluck2016changing, le2022linear}. Specifically, for each of the beginning- and end-of-year surveys, we constructed an undirected, unweighted network, where $A_{ij} = 1$ as long as student $i$ lists student $j$ as a friend or vice versa. We then averaged the adjacency matrices from the two surveys to produce the final undirected, weighted network used in the prediction modeling.

\paragraph{Methods.} For each of the 28 schools that received intervention in the study, we trained separate prediction models (\method, RNC, Network BART, \rfplus, RF, linear regression, and BART) using the aforementioned covariates and friendship network to predict the students' end-of-year friendliness scores. Each model was trained on the largest connected component of students (after removing students with missing survey responses) using 75\% of the data for training and evaluated on the remaining 25\%. The number of training samples ranged between 62 and 467, depending on the school (full list in Appendix~\ref{app:case_study}). We repeated this random training-test split 100 times for each school and averaged the results. 

\paragraph{Results.}   This case study presents a very noisy prediction problem:  no matter the school or prediction method, the test $R^2$ never exceeds 0.4 and averages around 0.17 (see Table~\ref{tab:school_conflict_data} in Appendix~\ref{app:case_study}). Given this low signal and the heterogeneity across schools, it is no surprise that different methods perform best for different schools on prediction accuracy, but on average, RNC performs the best, yielding an average test $R^2$ of 0.190 across all schools, followed closely by \method~with an average test $R^2$ of 0.188 (Table~\ref{tab:school_conflict_data}).   However, for all schools, the test $R^2$ values from \method~lie within one standard error of the best-performer, indicating that \method~is generally competitive with the best-performing method.
This competitiveness of \method~is noteworthy given that on average, RF yielded a lower test $R^2$ than ordinary linear regression for the large majority (i.e., 20 out of 28) of schools. Despite the heterogeneity between schools, \method~exhibited the most consistent prediction performance, being always ranked in the top-3 performing methods regardless of the school whereas RNC appeared in the bottom-4 performing methods for 3 schools (see Table~\ref{tab:school_conflict_ranks}). This consistency of performance is a key advantage of \method~and results from its data-driven adaptability to both nonlinear and linear structures.

\begin{figure}[t]
    \spacingset{1}
    \centering
    \includegraphics[width=0.98\linewidth]{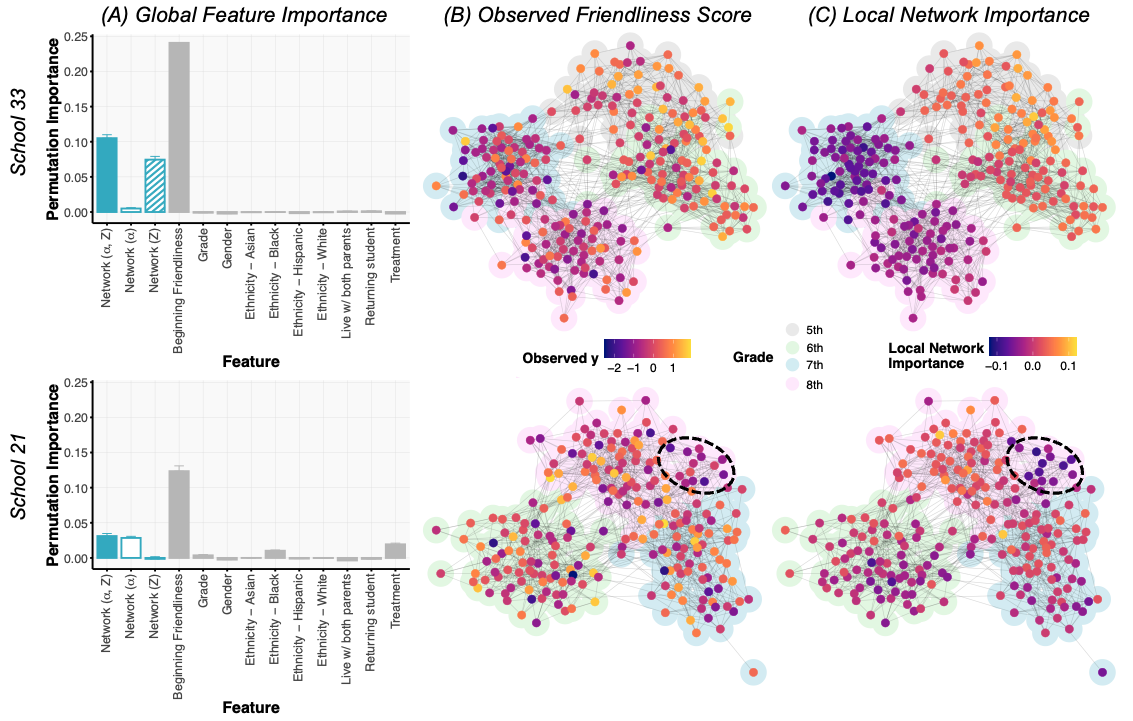}
    \caption{For two schools in the conflict intervention study, we show (A) the global permutation feature importances from \method, (B) the observed end-of-year school friendliness scores for each student, and (C) the local network importances from \method, which quantify the importance of the network effect for each student.    In School 21, we highlight a tight-knit group of friends (circled) in the 8th grade whose friendships, or peer-to-peer influences, are associated with a below-average perception of the school's friendliness atmosphere.}
    \label{fig:school_conflict_importances}
\end{figure}

Beyond prediction, we examined \method's feature importances to compare different student characteristics and types of network influences' effects on predicting the students' perceived school friendliness scores. While, unsurprisingly, the student's perceived school friendliness score at the beginning of the school year was by far the most important predictor of their end-of-year school friendliness score, Figure~\ref{fig:school_conflict_globalfi_nerfplus} also shows that the network was moderately important in many schools and that this  was driven by the network cohesion component for some schools and by the network embedding for others. 

We visualize the network importances for two such representative schools in Figure~\ref{fig:school_conflict_importances}. In School 33 (\method~test $R^2$: 0.292), where the network embedding component was globally more important than the network cohesion, the local network importances reveal a relatively strong community influence (mostly corresponding to their grade).   Note that importances are centered, so both large negative and large positive values indicate strong local importance, and we see that in this school, there is a substantial contribution from the network, with two grades (or communities) consistently above average and the other two consistently below.  
On the other hand, in School 21 (\method~test $R^2$: 0.111), where the network cohesion component was globally more important than the network embeddings, the local network importances do not exhibit a strong community structure, but provide evidence for strong peer-to-peer influence.
For example, there is a tight-knit friend group (circled) in the 8th grade whose connections are associated with a strong negative perception of the school's atmosphere.   Examining the observed school friendliness scores for this 8th grade friend group (Figure~\ref{fig:school_conflict_importances}(B)), we indeed see that their school friendliness scores are below-average compared to the other 8th graders, but without access to the local network importances, identifying this friend group solely from the observed friendliness network would have been challenging. Being able to identify subsets of the network where connections are especially important could be helpful for tailoring interventions.

\subsection{Case study 2:  Crime in Philadelphia} \label{subsec:case_study_philly_crime}

\paragraph{Data.} This dataset from \url{opendataphilly.org}, on monthly crime rates between January 2006 and December 2021 for different census tracts in Philadelphia, was studied by  \citet{deshpande2022flexbart} and hence included for comparison.  The response variable is the crime rate for a particular month and census tract, measured as the number of crimes per square mile per month. Following \citet{balocchi2023crime} and \citet{deshpande2022flexbart}, we transformed these crime rates (heavily skewed) by applying an inverse hyperbolic sine transformation.   To predict the crime rates, \citet{deshpande2022flexbart} used only the time variable indicating the month ($t = 1, \dots, 192$). We added weather-related covariates for that time (from \url{https://scacis.rcc-acis.org/}) including the average temperature per month, total amount of precipitation per month, number of days with precipitation per month, and the total amount of snowfall per month in Philadelphia. These covariates are expected to be only mildly related to the crime rates (and exhibit univariate correlations no stronger than 0.1), thereby providing an opportunity to examine feature selection and focus our attention on the network effect. 
The observed network is unweighted and undirected with $384$ nodes representing the census tracts in Philadelphia and edges connecting spatially adjacent tracts. 
Though a distance-weighted network could be used instead, we focus on the unweighted network for simplicity as the choice of weights can be challenging to tune in practice. 
In total, there are $n = 73,728$ observations, where each observation corresponds to a particular census tract and month. 

\paragraph{Methods.} Using the census tract network and covariates ($p=5$), we trained \method, RNC, Network BART, and their non-network-assisted counterparts (\rfplus, linear regression, and BART) to predict the monthly crime rates. Since multiple observations belong to each node (i.e., census tract), we adapted the \method~(and RNC) optimization problem \eqref{eq:nerf} so that all observations belonging to the same census tract shared the same $\alpha$ and $\Vec{z}$ as described in Section~\ref{subsec:nerf}. 

To evaluate performance, we split the data into a training and test set as follows.
First, we randomly selected $75\%$ of the tracts to be ``training'' tracts and $25\%$ of the tracts to be ``test'' tracts.
Then, pooling together observations from all the training tracts, we randomly sampled $\pi\%$ of these observations, with $\pi = \{0.1, 0.5, 1, 5, 10\}$, to form the training dataset.

The remaining observations from the training tracts, combined with the observations from the test tracts, form the test dataset. 
This design enables us to evaluate the test prediction error both on tracts that were seen in training and those that were never seen. Results are shown over 100 random data splits.

\paragraph{Results.}

\begin{figure}[t]
    \spacingset{1}
    \centering
    \includegraphics[width=0.95\linewidth]{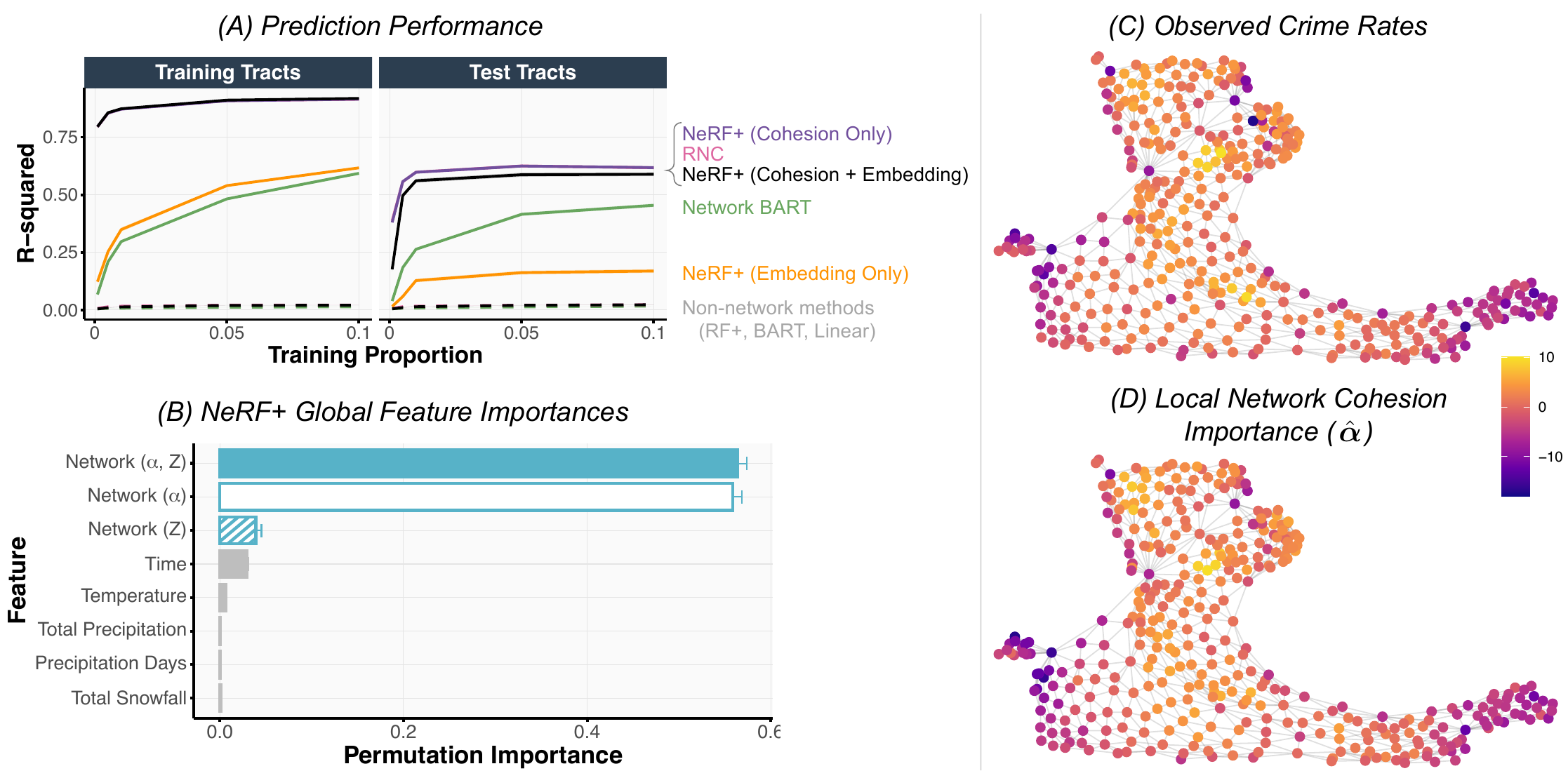}
    \caption{Philadelphia crime study. (A) Test prediction performance, measured via $R^2$, using varying amounts of data for training. Results are stratified by test observations from census tracts used in training (left) and census tracts not used in training (right). 
    (B) Permutation feature importances for the \method~fit (using 1\% of the data for training and both the network cohesion and network embeddings) indicate that the network cohesion is by far the most important contributor to the model predictions. Results in (A-B) are averaged across 100 training-test splits with error bars denoting $\pm 1 SE$. (C-D) Average observed crime rates and local network cohesion importances for each census tract are shown for an example run using 1\% of the data for training.
    The observed crime rates and local network cohesion importances are highly concordant with each other and reveal a smoothly varying structure.}
    \label{fig:philly_crime}
\end{figure}

Figure~\ref{fig:philly_crime}A summarizes the prediction performance results for the Philadelphia crime case study. We observe that \method~alongside RNC yield the best test prediction performance and can even achieve this optimal level of accuracy using $\sim 1\%$ of the data. 
More generally, it is also worth highlighting that: (1) the network-assisted methods (solid lines) substantially outperform the non-network-assisted methods (dashed lines), which show little-to-no predictive power, and (2) of the network-assisted methods, the network cohesion-based methods perform the best. This is expected given the spatial smoothness (translating to network cohesion) in the observed crime rates (Figure~\ref{fig:philly_crime}C), which makes the network information by far the most relevant.  

Both the global and local feature importances from \method~yield similar insights into the underlying network effect. In particular, the global feature importances from \method~(Figure~\ref{fig:philly_crime}B) reveal that (1) the network exhibited by far the highest global feature importance while the non-network features had close to zero importance, and (2) most of the overall network importance can be attributed to the network cohesion component, rather than the network embedding.   We also observe high concordance between the estimated local network cohesion importances from \method~and the observed crime rates (Figure~\ref{fig:philly_crime}C and D). These local importances clearly depict the strong network cohesion effect between adjacent census tracts and exemplify \method's ability to provide valuable, granular insights into the underlying network effects.

\section{Discussion}\label{sec:discussion}

\method~provides a flexible family of models, which can exploit dependencies created by a network when they help in prediction and capture both linear and nonlinear structures. The functional flexibility stems from \method's roots in \rfplus~while the network dependencies are captured through the network cohesion penalty, which effectively learns smoothly varying structures across the network, and through augmenting with network embedding features, which can be chosen to encode a variety of network properties.

An important contribution of \method~lies in its interpretation tools, which are directly derived from the model and are intrinsic to it, rather than post-hoc.  They can be used to identify the components that drive the model's predictions, examine the model for potential biases, provide key insights on how much predictive power is derived from the network connections compared to the features themselves, and suggest ways to further refine and improve the model. To facilitate the use of these interpretability tools in practice, we have developed an R package available at 
\if1\blind
  [\textit{redacted}],
\else
  \url{https://github.com/tiffanymtang/nerfplus},
\fi%
which includes a user-friendly Shiny application for interactive visualizations.  Finally, even our initial exploration of  local feature and network importances  shows that \method~can distill and illuminate granular insights and identify interesting or unusual nodes and features in a number of ways.   

When using \method~in practice, it should be emphasized that all interpretations are with respect to the fitted \method~model. That is, the feature importances (both global and local) measure how much a feature contributes to the prediction \textit{using the given \method~model}, and the sample influences measure how leaving out a sample impacts the fit \textit{of the given \method~model}.  Extrapolations of these interpretations from the model to the underlying scientific process require further careful and critical domain-specific examination of the model \citep{yu2020veridical}. 
Moreover, while the proposed feature importance measures serve as valuable starting points,  correlated features and other real-world challenges \citep{strobl2007bias, nicodemus2011stability, hooker2021unrestricted, alvarez2018robustness} can complicate the interpretation of feature importances, and they are not designed to provide evidence of truly causal relationships.  Causal inference on networks is an active area of research, complicated vastly by the concept of network inference, and is beyond the scope of this paper;  see for example \citet{aronow2017estimating} and \citet{forastiere2021identification}. 

There are a number of ways for building or improving upon \method.  They include learning  ``optimal'' network embeddings and fitting the model jointly (e.g., possibly incorporating ideas from \citet{luo2022bamdt} and \citet{deshpande2022flexbart}); leveraging a more sophisticated cross-validation scheme to account for network dependencies when tuning hyperparameters; extending \method~to more complex network structures such as dynamic or time-varying networks; and exploiting the community structure in networks to obtain community-level importances, which in a number of scientific applications, for example, in brain imaging, is the most scientifically relevant scale.  Ultimately,  \method~expands the scope and applicability of network-assisted machine learning, especially for problems where interpretability and transparency are essential.

\section*{Data and Code Availability}\label{app:data_codes}
Code to reproduce the results in this work can be found at 
\if1\blind
  [\textit{redacted}].
\else
  \url{https://github.com/tiffanymtang/nerfplus}.
\fi%
This repository also includes an R package and Shiny application to run and interpret \method~for custom datasets. Datasets used in this work are available via previous publications: \citet{paluck2016changing} for the school conflict intervention data (available from the authors by request); \citet{deshpande2022flexbart} for the Philadelphia crime data (see supplementary materials); and \citet{li2020stability} for the CCLE protein data (also made available at 
\if1\blind
  [\textit{redacted}]).
\else
  \url{https://github.com/tiffanymtang/nerfplus}).
\fi%

\section*{Acknowledgments}
This research was conducted when the first author was a postdoctoral fellow at the University of Michigan and is partially supported by NSF DMS grants 2210439 and 2610168 to E. Levina and J. Zhu.  





\printbibliography

@article{agarwal2023mdi+,
  title={Integrating Random Forests and Generalized Linear Models for Improved Accuracy and Interpretability}, 
  author={Agarwal, Abhineet and Kenney, Ana M and Tan, Yan Shuo and Tang, Tiffany M and Yu, Bin},
  journal={arXiv preprint arXiv:2307.01932},
  year={2025}
}

@InProceedings{luo2022bamdt,
  title = 	 {{BAMDT}: {B}ayesian Additive Semi-Multivariate Decision Trees for Nonparametric Regression},
  author =       {Luo, Zhao Tang and Sang, Huiyan and Mallick, Bani},
  booktitle = 	 {Proceedings of the 39th International Conference on Machine Learning},
  pages = 	 {14509--14526},
  year = 	 {2022},
  volume = 	 {162},
  publisher =    {PMLR}
}

@InProceedings{koh2017understanding,
  title = 	 {Understanding Black-box Predictions via Influence Functions},
  author =       {Pang Wei Koh and Percy Liang},
  booktitle = 	 {Proceedings of the 34th International Conference on Machine Learning},
  pages = 	 {1885--1894},
  year = 	 {2017},
  volume = 	 {70},
  publisher =    {PMLR}
}

@article{breiman2001random,
	title = {Random Forests},
	volume = {45},
	number = {1},
	journal = {Machine Learning},
	author = {Breiman, Leo},
	year = {2001},
	pages = {5--32},
}

@inproceedings{kipf2016semi,
  title={Semi-Supervised Classification with Graph Convolutional Networks},
  author={Kipf, Thomas N. and Welling, Max},
  booktitle={International Conference on Learning Representations (ICLR)},
  year={2017}
}

@inproceedings{hamilton2017inductive,
  title={Inductive representation learning on large graphs},
  author={Hamilton, Will and Ying, Zhitao and Leskovec, Jure},
  booktitle={Advances in Neural Information Processing Systems},
  volume={30},
  year={2017}
}

@article{lunde2023conformal,
  title={Conformal prediction for network-assisted regression},
  author={Lunde, Robert and Levina, Elizaveta and Zhu, Ji},
  journal={Journal of the American Statistical Association},
  volume = {0},
  number = {0},
  pages = {1--12},
  year={2025},
  publisher={Taylor \& Francis}
}

@article{le2022linear,
  title={Linear regression and its inference on noisy network-linked data},
  author={Le, Can M and Li, Tianxi},
  journal={Journal of the Royal Statistical Society Series B: Statistical Methodology},
  volume={84},
  number={5},
  pages={1851--1885},
  year={2022},
  publisher={Oxford University Press}
}

@article{klusowski2024large,
  title={Large scale prediction with decision trees},
  author={Klusowski, Jason M and Tian, Peter M},
  journal={Journal of the American Statistical Association},
  volume={119},
  number={545},
  pages={525--537},
  year={2024},
  publisher={Taylor \& Francis}
}

@article{li2019prediction,
  title={Prediction models for network-linked data},
  author={Li, Tianxi and Levina, Elizaveta and Zhu, Ji},
  volume = {13},
  journal = {The Annals of Applied Statistics},
  number = {1},
  publisher = {Institute of Mathematical Statistics},
  pages = {132--164},
  year={2019}
}

@inproceedings{grinsztajn2022tree,
  title={Why do tree-based models still outperform deep learning on typical tabular data?},
  author={Grinsztajn, L{\'e}o and Oyallon, Edouard and Varoquaux, Ga{\"e}l},
  booktitle={Advances in Neural Information Processing Systems},
  volume={35},
  year={2022}
}

@article{hampel1974influence,
  title={The influence curve and its role in robust estimation},
  author={Hampel, Frank R},
  journal={Journal of the American Statistical Association},
  volume={69},
  number={346},
  pages={383--393},
  year={1974},
  publisher={Taylor \& Francis}
}

@article{deshpande2022flexbart,
author = {Sameer K. Deshpande},
title = {{flexBART}: {F}lexible {B}ayesian Regression Trees with Categorical Predictors},
journal = {Journal of Computational and Graphical Statistics},
volume = {34},
number = {3},
pages = {1117--1126},
year = {2025}
}

@article{paluck2016changing,
  title={Changing climates of conflict: A social network experiment in 56 schools},
  author={Paluck, Elizabeth Levy and Shepherd, Hana and Aronow, Peter M},
  journal={Proceedings of the National Academy of Sciences},
  volume={113},
  number={3},
  pages={566--571},
  year={2016},
  publisher={National Acad Sciences}
}

@InProceedings{levin2018out,
  title = 	 {Out-of-sample extension of graph adjacency spectral embedding},
  author =       {Levin, Keith and Roosta, Fred and Mahoney, Michael and Priebe, Carey},
  booktitle = 	 {Proceedings of the 35th International Conference on Machine Learning},
  pages = 	 {2975--2984},
  year = 	 {2018},
  volume = 	 {80},
  publisher =    {PMLR}
}

@article{tibshirani1996regression,
  title = {{Regression Shrinkage and Selection Via the Lasso}},
  author={Tibshirani, Robert},
  journal={Journal of the Royal Statistical Society: Series B (Methodological)},
  volume={58},
  number={1},
  pages={267--288},
  year={1996},
  publisher={Wiley Online Library}
}

@article{hoerl1970ridge,
  title={Ridge regression: Biased estimation for nonorthogonal problems},
  author={Hoerl, Arthur E and Kennard, Robert W},
  journal={Technometrics},
  volume={12},
  number={1},
  pages={55--67},
  year={1970},
  publisher={Taylor \& Francis}
}

@book{breiman1984classification,
  title={Classification and Regression Trees},
  author={Breiman, Leo and Friedman, Jerome H. and Olshen, Richard A. and Stone, Charles J.},
  year={1984},
  publisher={Wadsworth International Group},
  address={Belmont, CA}
}

@article{behr2021provable,
  title={{Provable Boolean interaction recovery from tree ensemble obtained via random forests}},
  author={Behr, Merle and Wang, Yu and Li, Xiao and Yu, Bin},
  journal={Proceedings of the National Academy of Sciences},
  volume={119},
  number={22},
  pages={e2118636119},
  year={2022},
  publisher={National Academy of Sciences}
}

@article{kunzel2019linear,
  title={Linear Aggregation in Tree-based Estimators},
  author={K{\"u}nzel, S{\"o}ren R and Saarinen, Theo F and Liu, Edward W and Sekhon, Jasjeet S},
  journal={Journal of Computational and Graphical Statistics},
  volume={31},
  number={3},
  pages={917--934},
  year={2022},
  publisher={Taylor \& Francis}
}

@article{murdoch2019definitions,
  title={Definitions, methods, and applications in interpretable machine learning},
  author={Murdoch, W James and Singh, Chandan and Kumbier, Karl and Abbasi-Asl, Reza and Yu, Bin},
  journal={Proceedings of the National Academy of Sciences},
  volume={116},
  number={44},
  pages={22071--22080},
  year={2019},
  publisher={National Acad Sciences}
}

@article{rudin2019stop,
  title={Stop explaining black box machine learning models for high stakes decisions and use interpretable models instead},
  author={Rudin, Cynthia},
  journal={Nature Machine Intelligence},
  volume={1},
  number={5},
  pages={206--215},
  year={2019},
  publisher={Nature Publishing Group}
}

@inproceedings{agarwal2022hierarchical,
  title = 	 {Hierarchical Shrinkage: Improving the accuracy and interpretability of tree-based models.},
  author =       {Agarwal, Abhineet and Tan, Yan Shuo and Ronen, Omer and Singh, Chandan and Yu, Bin},
  booktitle = 	 {Proceedings of the 39th International Conference on Machine Learning},
  pages = 	 {111--135},
  year = 	 {2022},
  volume = 	 {162},
  publisher =    {PMLR}
}

@inproceedings{lundberg2017unified,
  title={A unified approach to interpreting model predictions},
  author={Lundberg, Scott M and Lee, Su-In},
  booktitle={Advances in Neural Information Processing Systems},
  volume={30},
  year={2017}
}

@article{zhou2021unbiased,
  title={Unbiased measurement of feature importance in tree-based methods},
  author={Zhou, Zhengze and Hooker, Giles},
  journal={ACM Transactions on Knowledge Discovery from Data (TKDD)},
  volume={15},
  number={2},
  pages={1--21},
  year={2021},
  publisher={ACM New York, NY, USA}
}

@article{sandri2008bias,
  title={A bias correction algorithm for the Gini variable importance measure in classification trees},
  author={Sandri, Marco and Zuccolotto, Paola},
  journal={Journal of Computational and Graphical Statistics},
  volume={17},
  number={3},
  pages={611--628},
  year={2008},
  publisher={Taylor \& Francis}
}

@inproceedings{li2019debiased,
  title={{A debiased MDI feature importance measure for random forests}},
  author={Li, Xiao and Wang, Yu and Basu, Sumanta and Kumbier, Karl and Yu, Bin},
  booktitle={Advances in Neural Information Processing Systems},
  volume={32},
  year={2019}
}

@article{lundberg2020local,
  title={From local explanations to global understanding with explainable {AI} for trees},
  author={Lundberg, Scott M and Erion, Gabriel and Chen, Hugh and DeGrave, Alex and Prutkin, Jordan M and Nair, Bala and Katz, Ronit and Himmelfarb, Jonathan and Bansal, Nisha and Lee, Su-In},
  journal={Nature Machine Intelligence},
  volume={2},
  number={1},
  pages={56--67},
  year={2020},
  publisher={Nature Publishing Group}
}

@article{strobl2007bias,
  title={Bias in random forest variable importance measures: Illustrations, sources and a solution},
  author={Strobl, Carolin and Boulesteix, Anne-Laure and Zeileis, Achim and Hothorn, Torsten},
  journal={BMC Bioinformatics},
  volume={8},
  number={1},
  pages={1--21},
  year={2007},
  publisher={BioMed Central}
}

@article{yu2020veridical,
  title={Veridical data science},
  author={Yu, Bin and Kumbier, Karl},
  journal={Proceedings of the National Academy of Sciences},
  volume={117},
  number={8},
  pages={3920--3929},
  year={2020},
  publisher={National Acad Sciences}
}

@article{nicodemus2011stability,
  title={On the stability and ranking of predictors from random forest variable importance measures},
  author={Nicodemus, Kristin K},
  journal={Briefings in Bioinformatics},
  volume={12},
  number={4},
  pages={369--373},
  year={2011},
  publisher={Oxford University Press}
}

@article{strobl2008conditional,
  title={Conditional variable importance for random forests},
  author={Strobl, Carolin and Boulesteix, Anne-Laure and Kneib, Thomas and Augustin, Thomas and Zeileis, Achim},
  journal={BMC Bioinformatics},
  volume={9},
  number={1},
  pages={1--11},
  year={2008},
  publisher={Springer}
}

@article{barretina2012cancer,
  title={{The Cancer Cell Line Encyclopedia enables predictive modelling of anticancer drug sensitivity}},
  author = {Barretina, Jordi and Caponigro, Giordano and Stransky, Nicolas and others},
  journal={Nature},
  volume={483},
  number={7391},
  pages={603--607},
  year={2012},
  publisher={Nature Publishing Group}
}

@article{wang2020simple,
  title={A simple new approach to variable selection in regression, with application to genetic fine mapping},
  author={Wang, Gao and Sarkar, Abhishek and Carbonetto, Peter and Stephens, Matthew},
  journal={Journal of the Royal Statistical Society: Series B (Statistical Methodology)},
  volume={82},
  number={5},
  pages={1273--1300},
  year={2020},
  publisher={Wiley Online Library}
}

@article{loecher2022unbiased,
  title={Unbiased variable importance for random forests},
  author={Loecher, Markus},
  journal={Communications in Statistics-Theory and Methods},
  volume={51},
  number={5},
  pages={1413--1425},
  year={2022},
  publisher={Taylor \& Francis}
}

@article{duncan2024simchef, 
year = {2024}, 
publisher = {The Open Journal}, 
volume = {9}, 
number = {95}, 
pages = {6156}, 
author = {Duncan, James and Tang, Tiffany and Elliott, Corrine F. and Boileau, Philippe and Yu, Bin}, 
title = {{simChef: High-quality data science simulations in R}}, 
journal = {Journal of Open Source Software} 
}

@article{hooker2021unrestricted,
	title = {Unrestricted permutation forces extrapolation: variable importance requires at least one more model, or there is no free variable importance},
	volume = {31},
	number = {6},
	journal = {Statistics and Computing},
	author = {Hooker, Giles and Mentch, Lucas and Zhou, Siyu},
	year = {2021},
	pages = {82},
  publisher={Springer}
}

@article{wu2020comprehensive,
  title={A comprehensive survey on graph neural networks},
  author={Wu, Zonghan and Pan, Shirui and Chen, Fengwen and Long, Guodong and Zhang, Chengqi and Philip, S Yu},
  journal={IEEE Transactions on Neural Networks and Learning Systems},
  volume={32},
  number={1},
  pages={4--24},
  year={2020},
  publisher={IEEE}
}

@article{zhou2020graph,
	title = {Graph neural networks: {A} review of methods and applications},
	volume = {1},
	journal = {AI Open},
	author = {Zhou, Jie and Cui, Ganqu and Hu, Shengding and Zhang, Zhengyan and Yang, Cheng and Liu, Zhiyuan and Wang, Lifeng and Li, Changcheng and Sun, Maosong},
	year = {2020},
	pages = {57--81},
}

@article{wang2023epistasis,
	title = {Epistasis regulates genetic control of cardiac hypertrophy},
	volume = {4},
	number = {6},
	journal = {Nature Cardiovascular Research},
	author = {Wang, Qianru and Tang, Tiffany M. and Youlton, Michelle and Weldy, Chad S. and Kenney, Ana M. and Ronen, Omer and Hughes, J. Weston and Chin, Elizabeth T. and Sutton, Shirley C. and Agarwal, Abhineet and Li, Xiao and Behr, Merle and Kumbier, Karl and Moravec, Christine S. and Tang, W. H. Wilson and Margulies, Kenneth B. and Cappola, Thomas P. and Butte, Atul J. and Arnaout, Rima and Brown, James B. and Priest, James R. and Parikh, Victoria N. and Yu, Bin and Ashley, Euan A.},
	year = {2025},
	pages = {740--760}
}

@article{simon2023interpreting,
  title={Interpreting random forest analysis of ecological models to move from prediction to explanation},
  author={Simon, Sophia M and Glaum, Paul and Valdovinos, Fernanda S},
  journal={Scientific Reports},
  volume={13},
  number={1},
  pages={3881},
  year={2023},
  publisher={Nature Publishing Group UK London}
}

@article{loef2022using,
  title={Using random forest to identify longitudinal predictors of health in a 30-year cohort study},
  author={Loef, Bette and Wong, Albert and Janssen, Nicole AH and Strak, Maciek and Hoekstra, Jurriaan and Picavet, H Susan J and Boshuizen, HC Hendriek and Verschuren, WM Monique and Herber, Gerrie-Cor M},
  journal={Scientific Reports},
  volume={12},
  number={1},
  pages={10372},
  year={2022},
  publisher={Nature Publishing Group UK London}
}

@article{belkin2003laplacian,
  title={Laplacian eigenmaps for dimensionality reduction and data representation},
  author={Belkin, Mikhail and Niyogi, Partha},
  journal={Neural Computation},
  volume={15},
  number={6},
  pages={1373--1396},
  year={2003},
  publisher={MIT Press}
}

@article{hoff2002latent,
  title={Latent space approaches to social network analysis},
  author={Hoff, Peter D and Raftery, Adrian E and Handcock, Mark S},
  journal={Journal of the American Statistical Association},
  volume={97},
  number={460},
  pages={1090--1098},
  year={2002},
  publisher={Taylor \& Francis}
}

@article{sussman2012consistent,
  title={A consistent adjacency spectral embedding for stochastic blockmodel graphs},
  author={Sussman, Daniel L and Tang, Minh and Fishkind, Donniell E and Priebe, Carey E},
  journal={Journal of the American Statistical Association},
  volume={107},
  number={499},
  pages={1119--1128},
  year={2012},
  publisher={Taylor \& Francis}
}

@inproceedings{grover2016node2vec,
  title={node2vec: Scalable feature learning for networks},
  author={Grover, Aditya and Leskovec, Jure},
  booktitle={Proceedings of the 22nd ACM SIGKDD International Conference on Knowledge Discovery and Data Mining},
  pages={855--864},
  year={2016},
  publisher={ACM}
}

@inproceedings{ying2019gnnexplainer,
  title={{GNNExplainer}: Generating explanations for graph neural networks},
  author={Ying, Zhitao and Bourgeois, Dylan and You, Jiaxuan and Zitnik, Marinka and Leskovec, Jure},
  booktitle={Advances in Neural Information Processing Systems},
  volume={32},
  year={2019}
}

@inproceedings{duval2021graphsvx,
	title = {{GraphSVX}: {S}hapley Value Explanations for Graph Neural Networks},
	booktitle = {Machine {Learning} and {Knowledge} {Discovery} in {Databases}. {Research} {Track}},
	publisher = {Springer International Publishing},
	author = {Duval, Alexandre and Malliaros, Fragkiskos D.},
	year = {2021},
	pages = {302--318},
}

@book{GNNBook2022,
author = "Wu, Lingfei and Cui, Peng and Pei, Jian and Zhao, Liang",
title = "Graph Neural Networks: Foundations, Frontiers, and Applications",
publisher = "Springer Singapore",
address = "Singapore",
pages = "725",
year = "2022",
}

@article{doshi2017towards,
  title={Towards a rigorous science of interpretable machine learning},
  author={Doshi-Velez, Finale and Kim, Been},
  journal={arXiv preprint arXiv:1702.08608},
  year={2017}
}

@article{lipton2018mythos,
  author = {Lipton, Zachary C.},
  title = {The mythos of model interpretability},
  year = {2018},
  issue_date = {October 2018},
  publisher = {Association for Computing Machinery},
  address = {New York, NY, USA},
  volume = {61},
  number = {10},
  journal = {Communications of the ACM},
  pages = {36--43},
  numpages = {8}
}

@book{molnar2020interpretable,
  title={Interpretable Machine Learning},
  subtitle={A Guide for Making Black Box Models Explainable},
  author={Christoph Molnar},
  year={2025},
  edition={3},
  isbn={978-3-911578-03-5},
  url={https://christophm.github.io/interpretable-ml-book}
}

@article{allen2023interpretable,
	title = {Interpretable Machine Learning for Discovery: Statistical Challenges and Opportunities},
	volume = {11},
	journal = {Annual Review of Statistics and Its Application},
	author = {Allen, Genevera I. and Gan, Luqin and Zheng, Lili},
	year = {2024},
	pages = {97--121},
}

@article{alvarez2018robustness,
  title={On the robustness of interpretability methods},
  author={Alvarez-Melis, David and Jaakkola, Tommi S},
  journal={arXiv preprint arXiv:1806.08049},
  year={2018}
}

@article{cook1977detection,
  title={Detection of influential observation in linear regression},
  author={Cook, R Dennis},
  journal={Technometrics},
  volume={19},
  number={1},
  pages={15--18},
  year={1977},
  publisher={Taylor \& Francis}
}

@book{cook1982residuals,
  title={Residuals and Influence in Regression},
  author={Cook, R. Dennis and Weisberg, Sanford},
  year={1982},
  publisher={Chapman \& Hall},
  address={New York}
}

@article{zou2005regularization,
  title={Regularization and variable selection via the elastic net},
  author={Zou, Hui and Hastie, Trevor},
  journal={Journal of the Royal Statistical Society Series B: Statistical Methodology},
  volume={67},
  number={2},
  pages={301--320},
  year={2005},
  publisher={Oxford University Press}
}

@article{chipman2010bart,
author = {Hugh A. Chipman and Edward I. George and Robert E. McCulloch},
title = {{BART: Bayesian additive regression trees}},
volume = {4},
journal = {The Annals of Applied Statistics},
number = {1},
publisher = {Institute of Mathematical Statistics},
pages = {266--298},
year = {2010}
}

@article{li2024explainable,
  title={Explainable Graph Neural Networks Under Fire},
  author={Li, Zhong and Geisler, Simon and Wang, Yuhang and G{\"u}nnemann, Stephan and van Leeuwen, Matthijs},
  journal={arXiv preprint arXiv:2406.06417},
  year={2024}
}

@article{manski1993identification,
  title={Identification of endogenous social effects: The reflection problem},
  author={Manski, Charles F},
  journal = {The Review of Economic Studies},
  volume={60},
  number={3},
  pages={531--542},
  year={1993},
  publisher={Wiley-Blackwell}
}

@article{chen2018tutorial,
  title={A tutorial on network embeddings},
  author={Chen, Haochen and Perozzi, Bryan and Al-Rfou, Rami and Skiena, Steven},
  journal={arXiv preprint arXiv:1808.02590},
  year={2018}
}

@article{wang2016trend,
  title={Trend filtering on graphs},
  author={Wang, Yu-Xiang and Sharpnack, James and Smola, Alexander J and Tibshirani, Ryan J},
  journal={Journal of Machine Learning Research},
  volume={17},
  number={105},
  pages={1--41},
  year={2016}
}

@article{chen2010graph,
  title={Graph-structured multi-task regression and an efficient optimization method for general fused lasso},
  author={Chen, Xi and Kim, Seyoung and Lin, Qihang and Carbonell, Jaime G and Xing, Eric P},
  journal={arXiv preprint arXiv:1005.3579},
  year={2010}
}

@incollection{shapley1953value,
  author       = {Shapley, Lloyd S.},
  title        = {A Value for n-Person Games},
  booktitle    = {Contributions to the Theory of Games II},
  editor       = {Kuhn, Harold W. and Tucker, Albert W.},
  publisher    = {Princeton University Press},
  address      = {Princeton},
  year         = {1953},
  pages        = {307--317}
}

@article{sparapani2021nonparametric,
  title={Nonparametric machine learning and efficient computation with {Bayesian} additive regression trees: The {BART R} package},
  author={Sparapani, Rodney and Spanbauer, Charles and McCulloch, Robert},
  journal={Journal of Statistical Software},
  volume={97},
  number={1},
  pages={1--66},
  year={2021}
}

@article{chen2018network,
  title={Network cross-validation for determining the number of communities in network data},
  author={Chen, Kehui and Lei, Jing},
  journal={Journal of the American Statistical Association},
  volume={113},
  number={521},
  pages={241--251},
  year={2018},
  publisher={Taylor \& Francis}
}

@article{li2020network,
  title={Network cross-validation by edge sampling},
  author={Li, Tianxi and Levina, Elizaveta and Zhu, Ji},
  journal={Biometrika},
  volume={107},
  number={2},
  pages={257--276},
  year={2020},
  publisher={Oxford University Press}
}

@article{tang2018limit,
  author = {Minh Tang and Carey E. Priebe},
  title = {{Limit theorems for eigenvectors of the normalized Laplacian for random graphs}},
  volume = {46},
  journal = {The Annals of Statistics},
  number = {5},
  publisher = {Institute of Mathematical Statistics},
  pages = {2360--2415},
  year = {2018}
}

@inproceedings{golea1997generalization,
	title = {Generalization in Decision Trees and {DNF}: Does Size Matter?},
	volume = {10},
	booktitle = {Advances in {Neural} {Information} {Processing} {Systems}},
	publisher = {MIT Press},
	author = {Golea, Mostefa and Bartlett, Peter and Lee, Wee Sun and Mason, Llew},
	year = {1997}
}

@article{liang2025local,
  title={{Local MDI+}: Local Feature Importances for Tree-Based Models},
  author={Liang, Zhongyuan and Rewolinski, Zachary T and Agarwal, Abhineet and Tang, Tiffany M and Yu, Bin},
  journal={arXiv preprint arXiv:2506.08928},
  year={2025}
}

@article{aronow2017estimating,
author = {Peter M. Aronow and Cyrus Samii},
title = {{Estimating average causal effects under general interference, with application to a social network experiment}},
volume = {11},
journal = {The Annals of Applied Statistics},
number = {4},
publisher = {Institute of Mathematical Statistics},
pages = {1912--1947},
year = {2017}
}

@article{forastiere2021identification,
  title={Identification and estimation of treatment and interference effects in observational studies on networks},
  author={Forastiere, Laura and Airoldi, Edoardo M and Mealli, Fabrizia},
  journal={Journal of the American Statistical Association},
  volume={116},
  number={534},
  pages={901--918},
  year={2021},
  publisher={Taylor \& Francis}
}

@article{fisher2019all,
  title={All models are wrong, but many are useful: Learning a variable's importance by studying an entire class of prediction models simultaneously},
  author={Fisher, Aaron and Rudin, Cynthia and Dominici, Francesca},
  journal={Journal of Machine Learning Research},
  volume={20},
  number={177},
  pages={1--81},
  year={2019}
}

@article{li2020stability,
  title={{A stability-driven protocol for drug response interpretable prediction (staDRIP)}},
  author={Li, Xiao and Tang, Tiffany M and Wang, Xuewei and Kocher, Jean-Pierre A and Yu, Bin},
  journal={arXiv preprint arXiv:2011.06593},
  year={2020}
}

@inproceedings{quinlan1992learning,
  title={Learning with Continuous Classes},
  author={Quinlan, J. Ross},
  booktitle={Proceedings of the 5th Australian Joint Conference on Artificial Intelligence},
  pages={343--348},
  year={1992},
  publisher={World Scientific}
}

@article{chaudhuri1994piecewise,
  title={Piecewise-Polynomial Regression Trees},
  author={Chaudhuri, Probal and Huang, Min-Ching and Loh, Wei-Yin and Yao, Ruji},
  journal={Statistica Sinica},
  volume={4},
  number={1},
  pages={143--167},
  year={1994}
}

@article{balocchi2023crime,
  title={Crime in Philadelphia: Bayesian clustering with particle optimization},
  author={Balocchi, Cecilia and Deshpande, Sameer K and George, Edward I and Jensen, Shane T},
  journal={Journal of the American Statistical Association},
  volume={118},
  number={542},
  pages={818--829},
  year={2023},
  publisher={Taylor \& Francis}
}

\clearpage
\appendix
\begin{center} 
\Large
\textbf{Appendix: ``Interpretable Network-assisted Random Forest$\plus$''}
\end{center} 

\section{Derivation of Sample Influences}\label{app:sample_influences}
In this section, we derive the closed-form leave-one-out (LOO) parameter estimate given in \eqref{eq:loo} for \method~with $\ell_2$ penalties.   This derivation only needs to be done for a single tree $t$.  

Recall that $\hat{\Vec{\nu}}^{\top} = [\hat{\Vec{\alpha}}^{\top}, \hat{\Vec{\beta}}^{\top}, \hat{\Vec{\gamma}}^{\top}]$ denotes the parameter estimates obtained by solving \eqref{eq:nerf} for one tree $t$, fitted on all $n$ training samples. When fitting \method~with $\ell_2$ penalties for $P_{\Vec{\beta}}$ and $P_{\Vec{\gamma}}$, \method~\eqref{eq:nerf} reduces to a generalized ridge regression problem, so that
\begin{align}
    \hat{\Vec{\nu}} = \left( \Mat{W}^{\top} \Mat{W} + \Mat{M} \right)^{-1} \Mat{W}^{\top} \Vec{y} = \Mat{B}^{-1} \Mat{W}^{\top} \Vec{y},
\end{align}
where
\begin{align*}
    \Mat{W} = [\Mat{I}_n, \tilde{\Mat{X}}, \Psi_t(\tilde{\Mat{X}})] = \begin{bmatrix} 
        \Vec{w}_1^{\top} \\ 
        \cdots \\
        \Vec{w}_n^{\top}
    \end{bmatrix}, \quad 
    \Mat{B} = \Mat{W}^{\top} \Mat{W} + \Mat{M}, \quad \text{and} \quad
    \quad \Mat{M} = \begin{bmatrix} 
        \lambda_{\alpha} \Mat{L} & \Mat{0} & \Mat{0} \\ 
        \Mat{0} & \lambda_{\beta} \Mat{I}_{p + r} & \Mat{0} \\
        \Mat{0} & \Mat{0} & \lambda_{\gamma} \Mat{I}_{m_t}
    \end{bmatrix}.
\end{align*}

Now, suppose we leave out the $i^{th}$ training sample and refit \eqref{eq:nerf} in \method~without re-estimating the tree $t$, network embeddings $\Mat{Z}$, the node degrees in the Laplacian matrix $\Mat{L}$, or the regularization parameters $\lambda_{\alpha}, \lambda_{\beta}, \lambda_{\gamma}$. Then, the corresponding LOO \method~parameter estimate, denoted $\hat{\Vec{\nu}}^{(-i)}$, is given by
\begin{align}
    \hat{\Vec{\nu}}^{(-i)} = \left( \Mat{W}^{\top}_{-i, -i} \Mat{W}_{-i, -i} + \Mat{M}_{-i, -i} \right)^{-1} \Mat{W}^{\top}_{-i, -i} \Vec{y}_{-i},
\end{align}
where the subscript $-i$ denotes the removal of the $i^{th}$ row or column of the corresponding matrix (or vector).

Letting $\Mat{B}^{(-i)} = \Mat{W}_{\cdot, -i}^{\top} \Mat{W}_{\cdot, -i} + \Mat{M}_{-i, -i}$, we can write
\begin{align*}
    \hat{\Vec{\nu}}^{(-i)} &= (\Mat{B}^{(-i)} - \Vec{w}_{i, -i} \Vec{w}_{i, -i}^{\top})^{-1} \Mat{W}_{-i, -i}^{\top} \Vec{y}_{-i} \\
    &= \left\{
        (\Mat{B}^{(-i)})^{-1} + 
        \frac{
            (\Mat{B}^{(-i)})^{-1} \Vec{w}_{i, -i} \Vec{w}_{i, -i}^{\top} (\Mat{B}^{(-i)})^{-1}
        }{
            1 - \Vec{w}_{i, -i}^{\top} (\Mat{B}^{(-i)})^{-1} \Vec{w}_{i, -i}
        } \right\} \Mat{W}_{-i, -i}^{\top} \Vec{y}_{-i} \\
    &= \left\{
        (\Mat{B}^{(-i)})^{-1} + 
        \frac{
            (\Mat{B}^{(-i)})^{-1} \Vec{w}_{i, -i} \Vec{w}_{i, -i}^{\top} (\Mat{B}^{(-i)})^{-1}
        }{
            1 - \Vec{w}_{i, -i}^{\top} (\Mat{B}^{(-i)})^{-1} \Vec{w}_{i, -i}
        } \right\} 
        \left(
        \Mat{W}_{\cdot, -i}^{\top} \Vec{y} - \Vec{w}_{i, -i} y_i    
        \right) \\
    &= (\Mat{B}^{(-i)})^{-1} \Mat{W}_{\cdot, -i}^{\top} \Vec{y} 
        - (\Mat{B}^{(-i)})^{-1} \Vec{w}_{i, -i} y_i \\
        & \qquad + \frac{(\Mat{B}^{(-i)})^{-1} \Vec{w}_{i, -i}}{1 - \Vec{w}_{i, -i}^{\top} (\Mat{B}^{(-i)})^{-1} \Vec{w}_{i, -i}} \left\{ 
            \Vec{w}_{i, -i}^{\top} (\Mat{B}^{(-i)})^{-1} \Mat{W}_{\cdot, -i}^{\top} \Vec{y}
            - \Vec{w}_{i, -i}^{\top} (\Mat{B}^{(-i)})^{-1} \Vec{w}_{i, -i} y_i    
        \right\},
\end{align*}
where the second equality follows from the Sherman-Morrison formula.

Note that Sherman-Morrison can also be applied to show that
\begin{align*}
    (\Mat{B}^{(-i)})^{-1} 
    &= [\Mat{B}^{-1}]_{-i, -i} - \frac{1}{[\Mat{B}^{-1}]_{ii}} [\Mat{B}^{-1}]_{-i, i} [\Mat{B}^{-1}]_{-i, i}^{\top} \\
    &= \left[\Mat{B}^{-1} - \frac{1}{[\Mat{B}^{-1}]_{ii}} [\Mat{B}^{-1}]_{\cdot i} [\Mat{B}^{-1}]_{\cdot i}^{\top} \right]_{-i, -i}.
\end{align*}

Defining $(\Mat{B}^{(i)})^{-1} = \Mat{B}^{-1} - \frac{1}{[\Mat{B}^{-1}]_{ii}} [\Mat{B}^{-1}]_{\cdot i} [\Mat{B}^{-1}]_{\cdot i}^{\top}$, it follows that $(\Mat{B}^{(-i)})^{-1} = [(\Mat{B}^{(i)})^{-1}]_{-i, -i}$. Using this identity, we have the following equivalences: 
\begin{enumerate}[label=(\roman*)]
    \item $\; \Vec{w}_{i, -i}^{\top} (\Mat{B}^{(-i)})^{-1} \Vec{w}_{i, -i} = \Vec{w}_{i}^{\top} (\Mat{B}^{(i)})^{-1} \Vec{w}_{i} =: h_i$,  
    \item $\; (\Mat{B}^{(-i)})^{-1} \Vec{w}_{i, -i} = \left[ (\Mat{B}^{(i)})^{-1} \Vec{w}_{i} \right]_{-i}$, and
    \item $\; (\Mat{B}^{(-i)})^{-1} \Mat{W}_{\cdot, -i}^{\top} \Vec{y} = \hat{\Vec{\nu}}_{-i} - \frac{\hat{\Vec{\nu}}_{i}}{[\Mat{B}^{-1}]_{ii}} [\Mat{B}^{-1}]_{-i, i}$,
\end{enumerate}
where (iii) follows from the observation that
\begin{align*}
    (\Mat{B}^{(-i)})^{-1} \Mat{W}_{\cdot, -i}^{\top} \Vec{y} 
    &= \underbrace{[\Mat{B}^{-1}]_{-i, -i} \Mat{W}_{\cdot, -i}^{\top} \Vec{y}}_{= \hat{\Vec{\nu}}_{-i} - [\Mat{B}^{-1}]_{-i, i} y_i} 
    - \frac{1}{[\Mat{B}^{-1}]_{ii}} [\Mat{B}^{-1}]_{-i, i} \underbrace{[\Mat{B}^{-1}]_{-i, i}^{\top} \Mat{W}_{\cdot, -i}^{\top} \Vec{y}}_{= \hat{\nu}_{i} - [\Mat{B}^{-1}]_{ii} y_i} \\
    &= \hat{\Vec{\nu}}_{-i} - \frac{\hat{\nu}_{i}}{[\Mat{B}^{-1}]_{ii}} [\Mat{B}^{-1}]_{-i, i}.
\end{align*}
Note also that similar to the ordinary linear regression case, $h_i$ here can be interpreted as the leverage of sample $i$ in the given generalized ridge regression problem.

Plugging these into the expression for $\hat{\Vec{\nu}}^{(-i)}$, we obtain
\begin{align*}
    \hat{\Vec{\nu}}^{(-i)}
    &= \left(\hat{\Vec{\nu}}_{-i} - \frac{\hat{\nu}_{i}}{[\Mat{B}^{-1}]_{ii}} [\Mat{B}^{-1}]_{-i, i}\right) - \left[ (\Mat{B}^{(i)})^{-1} \Vec{w}_{i} \right]_{-i} y_i \\
        &\qquad + \frac{\left[ (\Mat{B}^{(i)})^{-1} \Vec{w}_{i} \right]_{-i}}{1 - h_i} \left\{
            \Vec{w}_{i, -i}^{\top} \left( \hat{\Vec{\nu}}_{-i} - \frac{\hat{\nu}_{i}}{[\Mat{B}^{-1}]_{ii}} [\Mat{B}^{-1}]_{-i, i} \right)
            - h_i y_i    
        \right\} \\
    &= \hat{\Vec{\nu}}_{-i} - \frac{\hat{\nu}_{i}}{[\Mat{B}^{-1}]_{ii}} [\Mat{B}^{-1}]_{-i, i} \\
        &\qquad + \frac{\left[ (\Mat{B}^{(i)})^{-1} \Vec{w}_{i} \right]_{-i}}{1 - h_i} \left\{
            (h_i - 1) y_i 
            + \Vec{w}_{i, -i}^{\top} \hat{\Vec{\nu}}_{-i} 
            - \frac{\hat{\nu}_{i}}{[\Mat{B}^{-1}]_{ii}} \Vec{w}_{i, -i}^{\top} [\Mat{B}^{-1}]_{-i, i}
            - h_i y_i
        \right\} \\
    &= \hat{\Vec{\nu}}_{-i} - \frac{\hat{\nu}_{i}}{[\Mat{B}^{-1}]_{ii}} [\Mat{B}^{-1}]_{-i, i} 
        + \frac{\left[ (\Mat{B}^{(i)})^{-1} \Vec{w}_{i} \right]_{-i}}{1 - h_i} \left\{
            - y_i 
            + \Vec{w}_{i, -i}^{\top} \hat{\Vec{\nu}}_{-i} 
            - \frac{\hat{\nu}_{i}}{[\Mat{B}^{-1}]_{ii}} \Vec{w}_{i, -i}^{\top} [\Mat{B}^{-1}]_{-i, i}
        \right\} \\
    &= \hat{\Vec{\nu}}_{-i} - \frac{\hat{\alpha}_{i}}{[\Mat{B}^{-1}]_{ii}} [\Mat{B}^{-1}]_{-i, i} 
        + \frac{\left[ (\Mat{B}^{(i)})^{-1} \Vec{w}_{i} \right]_{-i}}{1 - h_i} \left\{
            - y_i 
            + (\hat{y}_i - \hat{\alpha}_i)
            - \frac{\hat{\alpha}_{i}}{[\Mat{B}^{-1}]_{ii}} \Vec{w}_{i, -i}^{\top} [\Mat{B}^{-1}]_{-i, i}
        \right\},
\end{align*}
where the last equality follows because $\Vec{w}_{i, -i}^{\top} \hat{\Vec{\nu}}_{-i} = \hat{y}_i - \hat{\alpha}_i$ and $\hat{\nu}_i = \hat{\alpha}_i$.

Finally, since
\begin{align*}
    \Vec{w}_{i, -i}^{\top} [\Mat{B}^{-1}]_{-i, i} = \Vec{w}_i^{\top} [\Mat{B}^{-1}]_{\cdot i} - w_{ii} [\Mat{B}^{-1}]_{ii} = \Vec{w}_i^{\top} [\Mat{B}^{-1}]_{\cdot i} - [\Mat{B}^{-1}]_{ii},
\end{align*}
we have that
\begin{align*}
    \hat{\Vec{\nu}}^{(-i)} 
    &= \hat{\Vec{\nu}}_{-i} - \frac{\hat{\alpha}_{i}}{[\Mat{B}^{-1}]_{ii}} [\Mat{B}^{-1}]_{-i, i}
        + \frac{\left[ (\Mat{B}^{(i)})^{-1} \Vec{w}_{i} \right]_{-i}}{1 - h_i} \left\{
            - y_i 
            + (\hat{y}_i - \hat{\alpha}_i)
            - \frac{\hat{\alpha}_{i}}{[\Mat{B}^{-1}]_{ii}} \Vec{w}_{i}^{\top} [\Mat{B}^{-1}]_{\cdot i} - \hat{\alpha}_i
        \right\} \\
    &= \hat{\Vec{\nu}}_{-i} 
        + \frac{\left[ (\Mat{B}^{(i)})^{-1} \Vec{w}_{i} \right]_{-i}}{1 - h_i} (\hat{y}_i - y_i)
        - \frac{\hat{\alpha}_{i}}{[\Mat{B}^{-1}]_{ii}} \left( 
            [\Mat{B}^{-1}]_{-i, i}  + 
            \frac{\left[ (\Mat{B}^{(i)})^{-1} \Vec{w}_{i} \right]_{-i} \Vec{w}_{i}^{\top} [\Mat{B}^{-1}]_{\cdot i}}{1 - h_i}
        \right) \\
    &= \left[\hat{\Vec{\nu}}
        + \frac{(\Mat{B}^{(i)})^{-1} \Vec{w}_{i}}{1 - h_i} (\hat{y}_i - y_i)
        - \frac{\hat{\alpha}_{i}}{[\Mat{B}^{-1}]_{ii}} \left( 
            [\Mat{B}^{-1}]_{\cdot i}  + 
            \frac{(\Mat{B}^{(i)})^{-1} \Vec{w}_{i} \Vec{w}_{i}^{\top} [\Mat{B}^{-1}]_{\cdot i}}{1 - h_i}
        \right) \right]_{-i},
\end{align*}
completing the derivation of \eqref{eq:loo}.

\section{Simulation Details and Results}\label{app:sims}

In this section, we provide additional details and results relating to the simulation studies.
Specifically, we describe the choice of hyperparameters and implementation details for methods used in the simulations in Section~\ref{subsec:implementation_details}.
In Section~\ref{subsec:main_sims}, we provide additional simulation results to supplement those presented in Section~\ref{sec:sims}.
We also report results of additional simulations to demonstrate the effectiveness of \method~with correlated features (Section~\ref{subsec:correlated_sims}), multiple observations per node (Section~\ref{subsec:repeated_obs_sims}), various network embedding choices (Section~\ref{subsec:robustness_sims}), and performance on classification tasks (Section~\ref{subsec:classification_sims}).
We report comparisons between \method, Network BART, and Bayesian Additive Semi-Multivariate Decision Trees (BAMDT), another network-assisted variant of BART, in Section~\ref{subsec:bamdt}.
Finally, we provide results to validate our closed-form approximation for the sample influence measures in Section~\ref{subsec:loo_sims} and report the computational costs of fitting \method~in Section~\ref{subsec:runtime_sims}.

All methods were implemented using R, and the simulation study was conducted using the \texttt{simChef} R package \citep{duncan2024simchef} to facilitate reproducibility.

\subsection{Implementation Details}\label{subsec:implementation_details}

\paragraph{\method.} We fit \method~with $\ell_2$ penalties for $P_{\beta}$ and $P_{\gamma}$. Here, the initial RF within \method~was fitted using the \texttt{ranger} R package with the following hyperparameters: number of trees $= 500$, proportion of features to consider at each split $= \frac{1}{3}$, and all other default hyperparameter values. We set the Laplacian regularization parameter to $\lambda_L = 0.05$ and the network embeddings $\Mat{Z}$ to be the top two components from the regularized Laplacian spectral embedding of the network. To select the optimal regularization parameters in \method, we used 5-fold CV and performed a grid search over all possible combinations of the regularization parameters $\lambda_{\alpha}, \lambda_{\beta}, \lambda_{\gamma}$, considering 10 values logarithmically spaced between $10^{-4}$ and $10^{3}$ for each regularization parameter. Because tuning these regularization parameters for each tree in \method~can be computationally expensive for our repeated simulations, we tuned these regularization parameters for 10 (out of 500) trees in each forest and randomly chose from these tuned regularization parameters for the remaining trees in the forest. Finally, when computing the permutation feature importances, we used 50 permutations for each feature and evaluated the change in the test $R^2$ value. When computing \mdiplus, we used the $R^2$ value to assess the similarity between the observed responses and the partial model predictions.

\paragraph{RNC.} Similar to \method, we fit RNC with an $\ell_2$ penalty applied to the regression coefficients, set the Laplacian regularization parameter to $\lambda_L = 0.05$, and used 5-fold grid search CV to select the optimal regularization parameters $\lambda_{\alpha}$ and $\lambda_{\beta}$, considering 10 values logarithmically spaced between $10^{-4}$ and $1000$ for each regularization parameter.

\paragraph{Network BART.} We implemented Network BART using the \texttt{flexBART} R package \citep{deshpande2022flexbart}. Here, we allowed for non-uniform cutpoints (and instead, searched across all possible cutpoints for each feature) and set all other hyperparameters to their defaults.

\paragraph{\rfplus.} \rfplus~was fitted in a similar manner as \method, but with $\lambda_{\alpha} = 0$.

\paragraph{BART.} To fit BART, we used the \texttt{BART} R package \citep{sparapani2021nonparametric} with the following hyperparameters: number of trees $= 500$, number of burn-in iterations $= 1000$, number of posterior draws $= 1000$, and all other hyperparameters set to their default values.

\subsection{Main Simulations}\label{subsec:main_sims}

\begin{figure}
    \spacingset{1}
    \centering
    \includegraphics[width=1\linewidth]{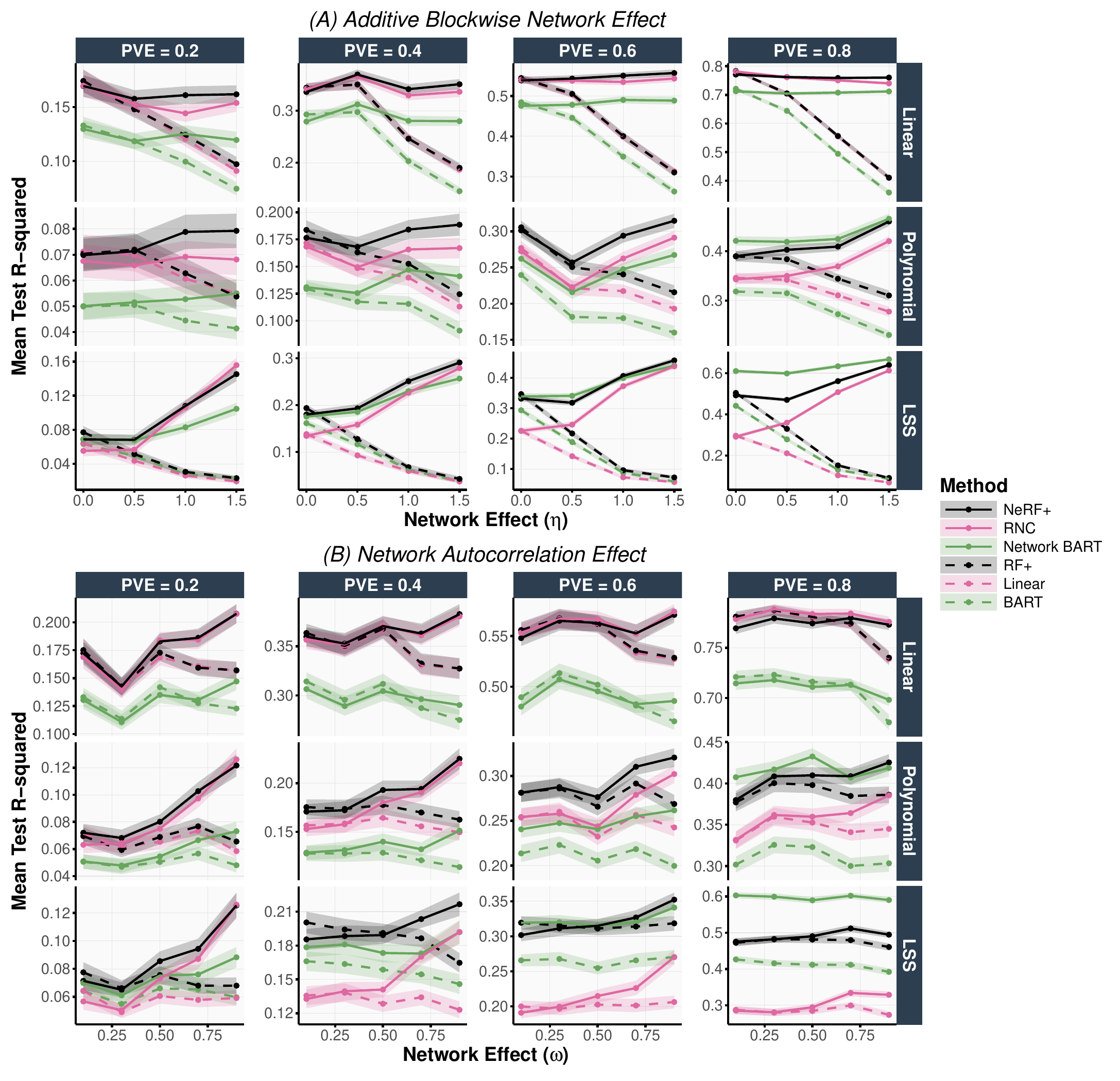}
    \caption{Prediction accuracy, as measured by the test $R^2$ value (y-axis), under (A) the additive blockwise network effect simulation and (B) the network autocorrelation effect simulation with i.i.d.\ standard normal features. Results are shown across different underlying functional forms (rows), noise levels (columns), and varying strengths of the network effect (x-axis). Test $R^2$ values are averaged across 100 simulation replicates with shaded regions denoting $\pm 1 SE$.}
    \label{fig:sim_predictions_supplement}
\end{figure}

\paragraph{Prediction Performance.} 
In Figure~\ref{fig:sim_predictions_supplement}, we provide the prediction accuracy of all methods across different functional forms, noise levels ($\mathrm{PVE}$), and strengths of the network effect under both the additive blockwise network effect and network autocorrelation effect simulation models with i.i.d.\ standard normal covariates (described in Section~\ref{sec:sims}). These results supplement those presented in Figure~\ref{fig:sim_predictions}. With the exception of the LSS simulation under very low noise levels ($\mathrm{PVE} = 0.8$), we see that \method~consistently yields the best or is on par with the best-performing prediction model.
Interestingly, under the LSS scenario when $\mathrm{PVE} = 0.8$, Network BART yields a higher $R^2$ than \method. Given that this gap is only observed under this very low-noise setting, we suspect that this may be due to Network BART's ability to more accurately and efficiently capture the interacting covariate terms in the LSS model, whereas the greedy nature of CART and inclusion of linear terms in \method~are less well-suited to this scenario. However, as Network BART unfortunately does not provide feature importance measures, we are not able to verify this hypothesis about the source of improvement.  Regardless, across most scenarios and particularly those under realistic noise levels observed in practice,\footnote{In many genomics applications, the PVE is estimated to be between 0.05 to 0.4 \citep{wang2020simple}.} \method~yields the highest test $R^2$ values, showcasing its ability to accurately and flexibly capture a wide variety of data-generating processes in practice. 

\begin{figure}
    \spacingset{1}
    \centering
    \includegraphics[width=1\linewidth]{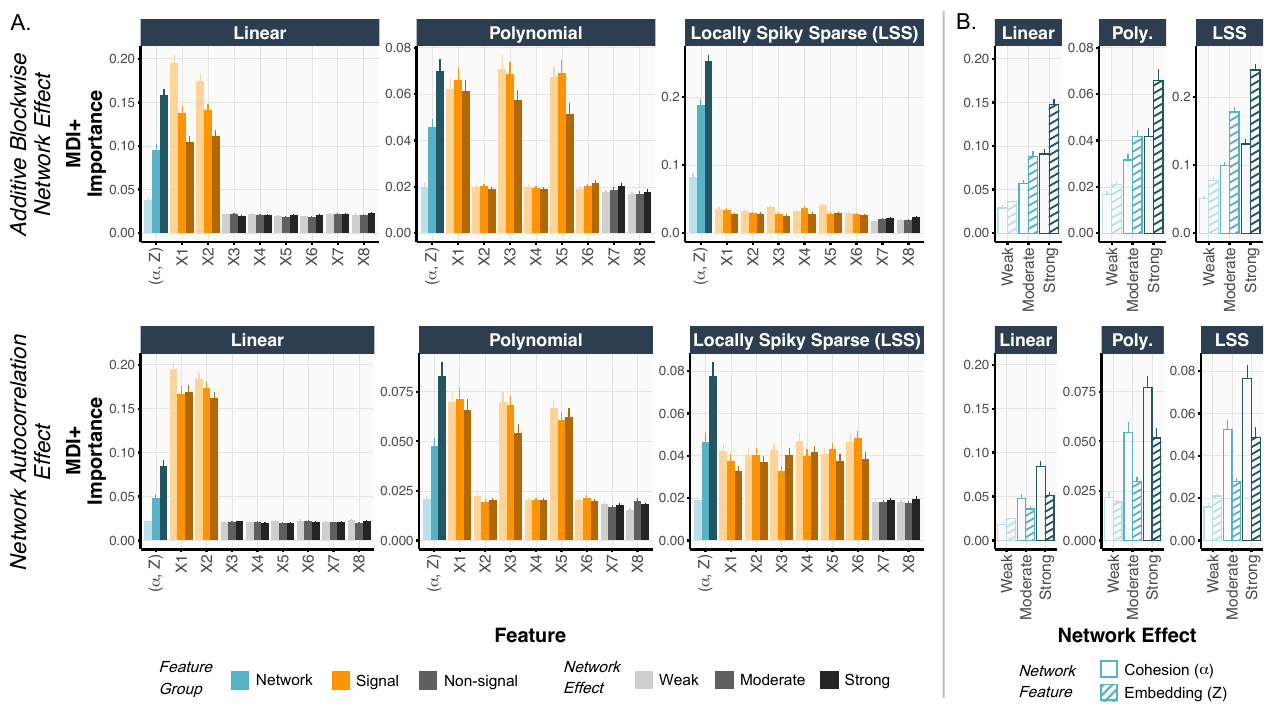}
    \caption{
        \method~global feature importances, measured via \mdiplus~importance at a moderately-low signal-to-noise ratio ($\mathrm{PVE} = 0.4$). 
        (A) Signal features exhibit higher importance than the non-signal features, and as the network effect increases (i.e., as the fill shade darkens), the estimated global importance of the network increases. Note: only the first 8 features are shown since the remaining features are non-signal and exhibit similarly low importances as the non-signal features shown. (B) The network embedding features have greater importance under the additive blockwise network effect model compared to the network autocorrelation effect model, while the opposite is true for the network cohesion component. Results are averaged across 100 replicates with error bars denoting $\pm 1 SE$.
    }
    \label{fig:sim_importances_mdiplus}
\end{figure}

\begin{figure}
    \spacingset{1}
    \centering
    \includegraphics[width=1\linewidth]{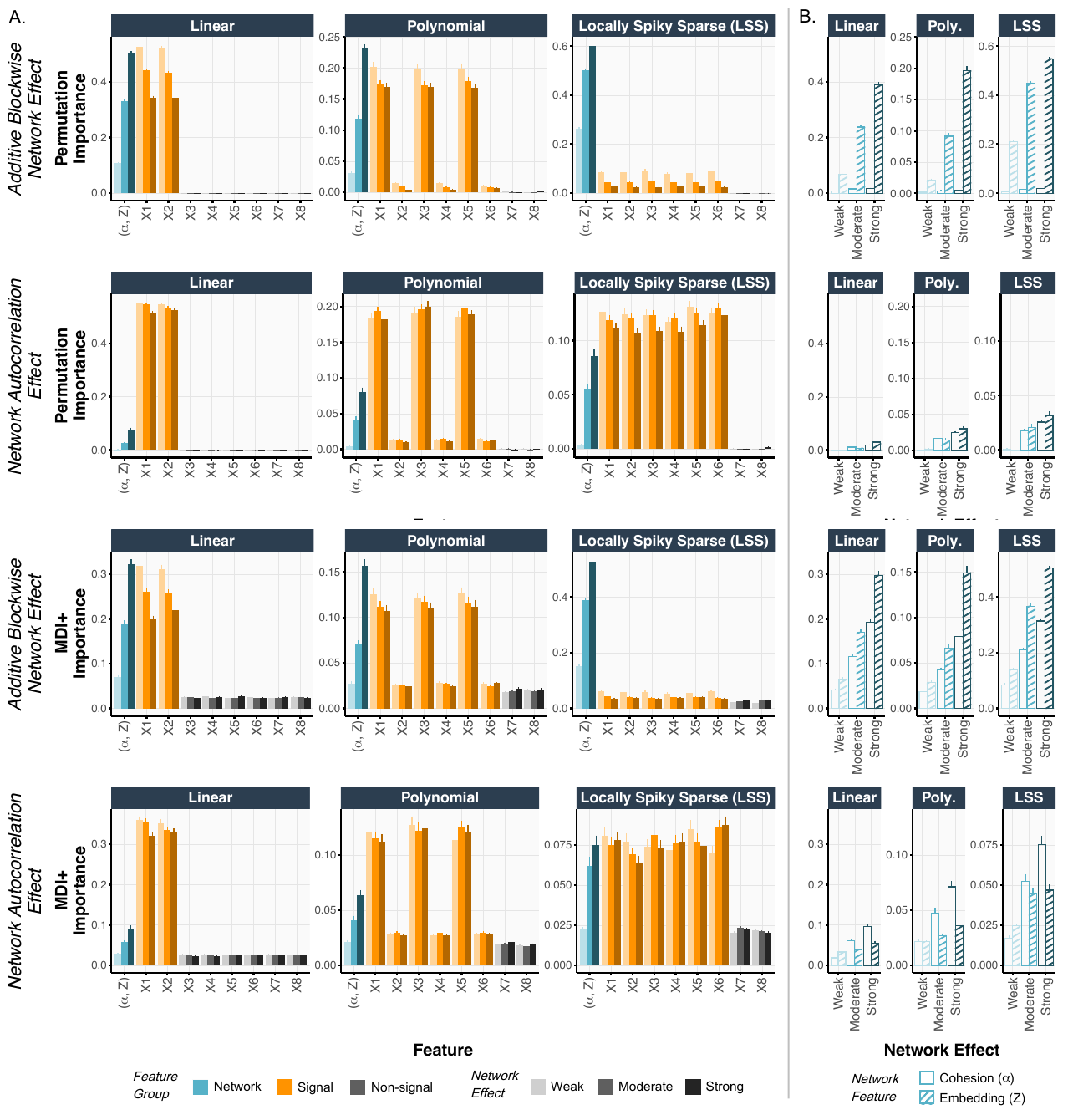}
    \caption{
        \method~global feature importances, measured via permutation and \mdiplus~importance at a high signal-to-noise ratio ($\mathrm{PVE} = 0.8$). 
        (A) Signal features exhibit higher importance than the non-signal features, and as the network effect increases (i.e., as the fill shade darkens), the estimated global importance of the network increases. Note: only the first 8 features are shown since the remaining features are non-signal and exhibit similarly low importances as the non-signal features shown. (B) The network embedding features have greater importance under the additive blockwise network effect model compared to the network autocorrelation effect model, while the opposite is true for the network cohesion component. Results are averaged across 100 replicates with error bars denoting $\pm 1 SE$.
    }
    \label{fig:sim_importances_0.8}
\end{figure}

\paragraph{Global Feature Importances.}
In Figure~\ref{fig:sim_importances_mdiplus}, we show the \mdiplus~global feature importances from \method~when $\mathrm{PVE} = 0.4$. Similar to the permutation feature importances presented in Figure~\ref{fig:sim_importances_permutation}, the \mdiplus~importances (i) effectively distinguish between signal and non-signal features, (ii) accurately show an increasing importance of the network as the simulated network effect strengthens, and (iii) reveal a higher importance from the network embedding features compared to the network cohesion component in the additive blockwise network effect model, while the opposite is true in the network autocorrelation effect model. These findings also hold under different signal-to-noise ratios (see Figure~\ref{fig:sim_importances_0.8} where $\mathrm{PVE} = 0.8$).

\subsection{Correlated Features Simulations}\label{subsec:correlated_sims}

\paragraph{Simulation Setup.} In the base simulations presented in Section~\ref{sec:sims}, we generated the covariate matrix $\Mat{X}$ with i.i.d.\ standard normal entries. To demonstrate \method's strong performance under more realistic correlated designs, we repeated the simulation study, keeping the same data-generating processes for the response and network but using a real dataset as the covariate matrix $\Mat{X}$, the proteomics data from the Cancer Cell Line Encyclopedia (CCLE) \citep{barretina2012cancer}, which is known to have highly correlated features. We processed the data as in previous work \citep{li2020stability}. This dataset contains 370 cell lines (samples) and 214 proteins (features). For each simulation replicate, we randomly sampled 80\% of the 370 samples for training and used the remaining 20\% for testing. We also randomly sampled different features to serve as the signal features in $f$ in each simulation replicate, but kept the full set of features ($p = 214$) in $\Mat{X}$ when fitting the models. Note that due to the correlated nature of these real-world features, there was extremely low variability in the Locally Spiky Sparse (LSS) scenario, and thus we omitted it from the correlated features simulation study.

\begin{figure}
    \spacingset{1}
    \centering
    \includegraphics[width=1\linewidth]{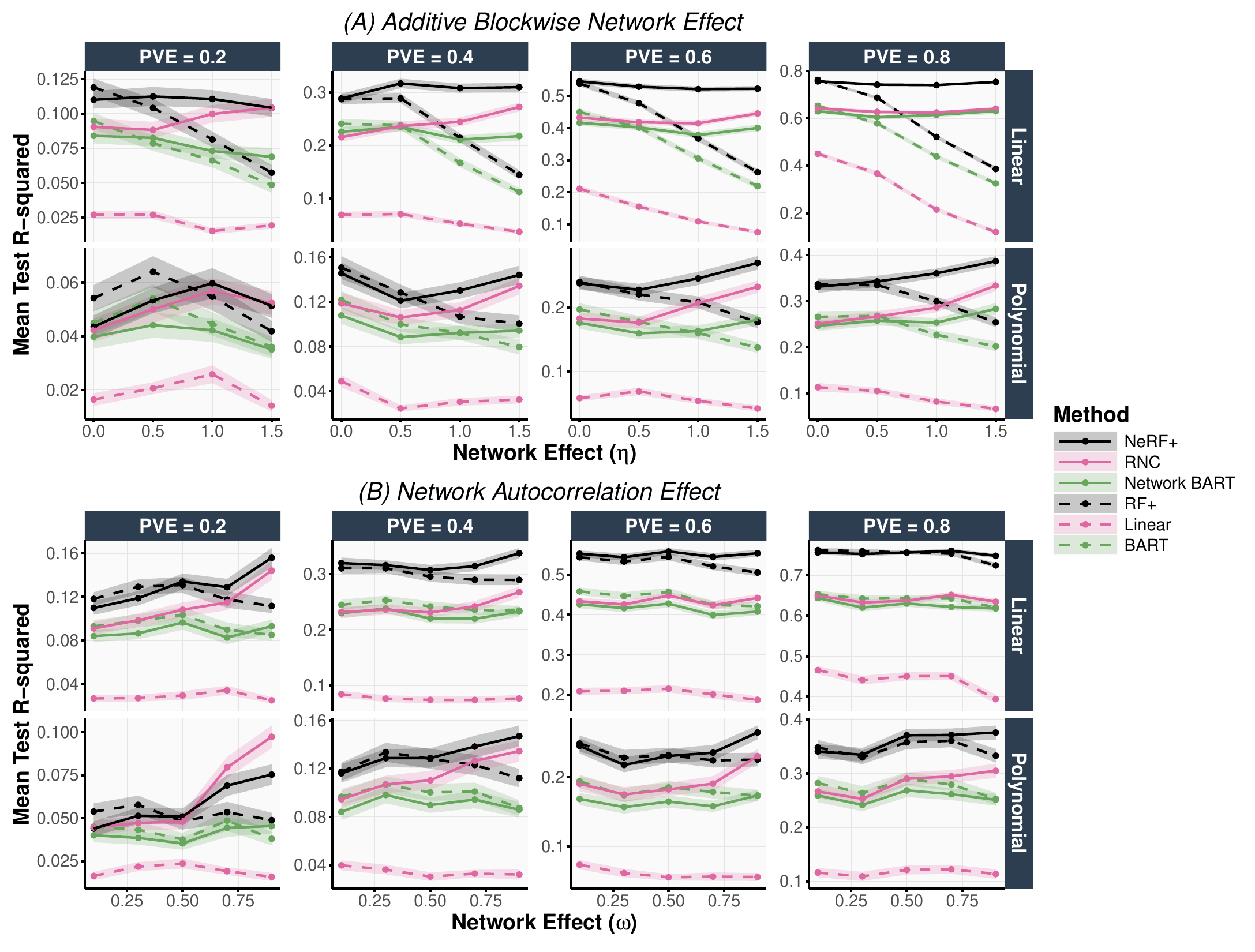}
    \caption{Prediction accuracy with correlated features, as measured by the test $R^2$ value (y-axis), under (A) the additive blockwise network effect simulation and (B) the network autocorrelation effect simulation. Results are shown across different underlying functional forms (rows), noise levels (columns), and varying strengths of the network effect (x-axis). Test $R^2$ values are averaged across 100 replicates with shaded regions denoting $\pm 1 SE$.}
    \label{fig:sim_predictions_real_data_supplement}
\end{figure}

\paragraph{Prediction Performance.} Figure~\ref{fig:sim_predictions_real_data_supplement} summarizes the prediction accuracies from the correlated features simulation study. As in the independent features setting, \method~consistently yields the best or is on par with the best-performing prediction model across both linear and nonlinear (i.e., polynomial) simulation settings. We also observe that under the linear simulation setting, \method~substantially outperforms RNC in terms of prediction accuracy; this is unlike the independent feature setting, where RNC and \method~yielded almost identical prediction accuracies under the linear simulation setting. We speculate this gap in performance is due to the sparse nature of the true linear model and the large number of features ($p = 214$) in the data. While RNC is a dense (non-sparse) model, \method~has two mechanisms for adapting to sparsity: (i) the trees in RF inherently perform feature selection as they may not split on all features, and (ii) as recommended in \citet{agarwal2023mdi+}, only those features used to split in the trees are included in the $\Mat{X} \Vec{\beta}$ term in \method.

\begin{figure}
    \spacingset{1}
    \centering
    \includegraphics[width=0.9\linewidth]{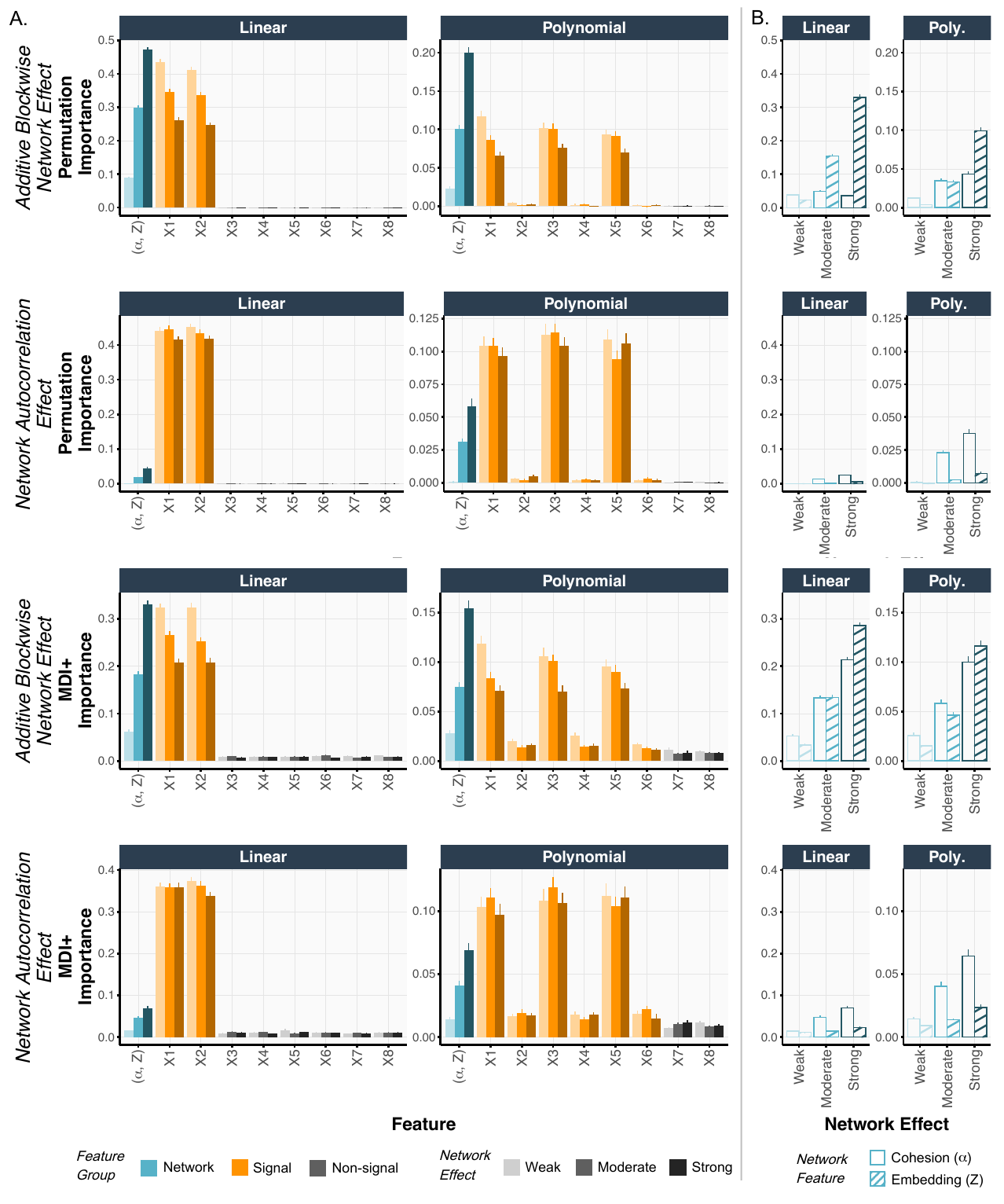}
    \caption{
        \method~global feature importances with correlated features, measured via permutation and \mdiplus~importance  at a high signal-to-noise ratio ($\mathrm{PVE} = 0.8$).  (A) Signal features exhibit higher importance than the non-signal features, and as the network effect increases (i.e., as the fill shade darkens), the estimated global importance of the network also increases. Note: only the first 8 features (out of 214) are shown since remaining features are non-signal and exhibit similarly low importances as the non-signal features shown here. (B) Network cohesion has greater importance than the network embedding features in the network autocorrelation effect model, while the opposite is true in the additive blockwise network effect model. Results are averaged across 100 replicates with error bars denoting $\pm 1 SE$.}
    \label{fig:sim_importances_real_data}
\end{figure}

\paragraph{Global Feature Importances.} In Figure~\ref{fig:sim_importances_real_data}, we provide the corresponding \method~global feature importances from the correlated features simulation study. Similar to the independent features setting, we show that the permutation and \mdiplus~importances satisfy several desirable properties even in the presence of correlated features. Namely, (i) the signal features correctly have higher importance than the non-signal features, (ii) the importance of the network increases as the simulated strength of the network effect increases, and (iii) the importance of network cohesion is generally higher than that of the network embedding features in the network autocorrelation effect model while the opposite is generally true in the additive blockwise network effect model.

\subsection{Simulations with Multiple Observations per Node}\label{subsec:repeated_obs_sims}

\paragraph{Simulation Setup.} As briefly mentioned in Section~\ref{subsec:nerf}, \method~can be extended to accommodate multiple observations per node. To demonstrate this, we simulated data from the additive blockwise network effect model, described in Section~\ref{sec:sims}, with $\mathrm{PVE} = 0.6$ and varying sample sizes $n \in \{150, 300, 450\}$. For each sample size, we compared the performance of \method~applied to the situation where there is only one observation per node versus the situation where there are only 150 nodes, each with $n / 150$ observations.

\begin{figure}[tb]
    \spacingset{1}
    \centering
    \includegraphics[width=0.9\linewidth]{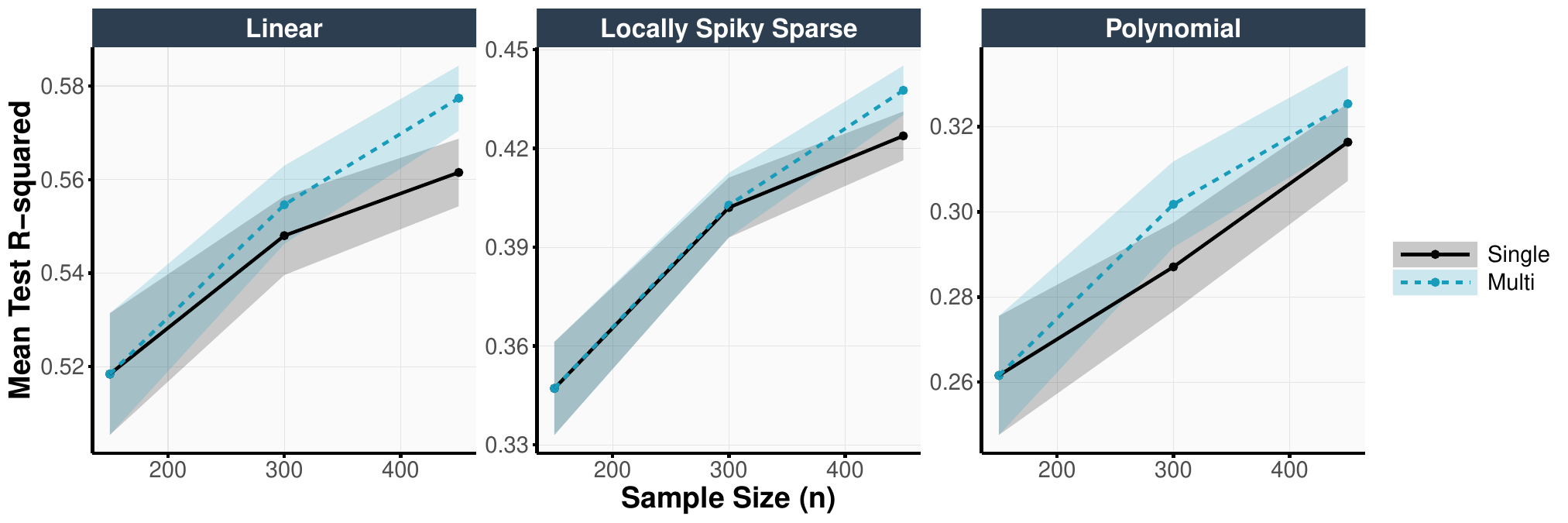}
    \caption{Comparison of \method~prediction performance under the additive blockwise network effect simulation with one observation per node (solid line) versus multiple observations per node (dashed line). In the setting with multiple observations per node, there are a total of 150 nodes. The x-axis shows the total number of samples $n$, and the y-axis shows the test $R^2$ value. Results are averaged across 100 replicates with shaded regions denoting $\pm 1 SE$.}
    \label{fig:sim_repeated_obs}
\end{figure}

\paragraph{Prediction Performance.}
Figure~\ref{fig:sim_repeated_obs} shows the prediction performance of \method~in these two scenarios. 
We see that when there are multiple observations per node, the prediction problem generally becomes easier for \method, as evidenced by the higher test $R^2$ values. 
This demonstrates that \method~can effectively leverage multiple observations per node to improve prediction performance.

\subsection{Comparing Network Embedding Choices}\label{subsec:robustness_sims}

\paragraph{Simulation Setup.} We next assess the robustness of \method's performance to the choices related to the network embedding, namely (i) the network embedding method, (ii) the number of embedding dimensions $r$, and (iii) the Laplacian regularization parameter $\lambda_{L}$. To this end, we repeated the base simulations described in Section~\ref{sec:sims} with different choices for each of these three components. 
Specifically, to assess the robustness of \method~to the choice of network embedding method, we fixed $r = 2$ and $\lambda_{L} = 0.05$ and considered three different network embedding choices: the Laplacian spectral embedding, the adjacency spectral embedding, and concatenating both the Laplacian and adjacency spectral embeddings for $\Mat{Z}$. 
Then, to assess the robustness of \method~to the number of embedding dimensions, we fixed the network embedding method to be the Laplacian spectral embedding and $\lambda_{L} = 0.05$ and considered $r \in \{1, 2, 3, 5\}$. 
Finally, to assess the robustness of \method~to the Laplacian regularization parameter, we fixed the network embedding method to be the Laplacian spectral embedding and $r = 2$ and considered $\lambda_{L} \in \{0.01, 0.05, 0.1\}$.

\begin{figure}[!tb]
    \spacingset{1}
    \centering
    \includegraphics[width=1\linewidth]{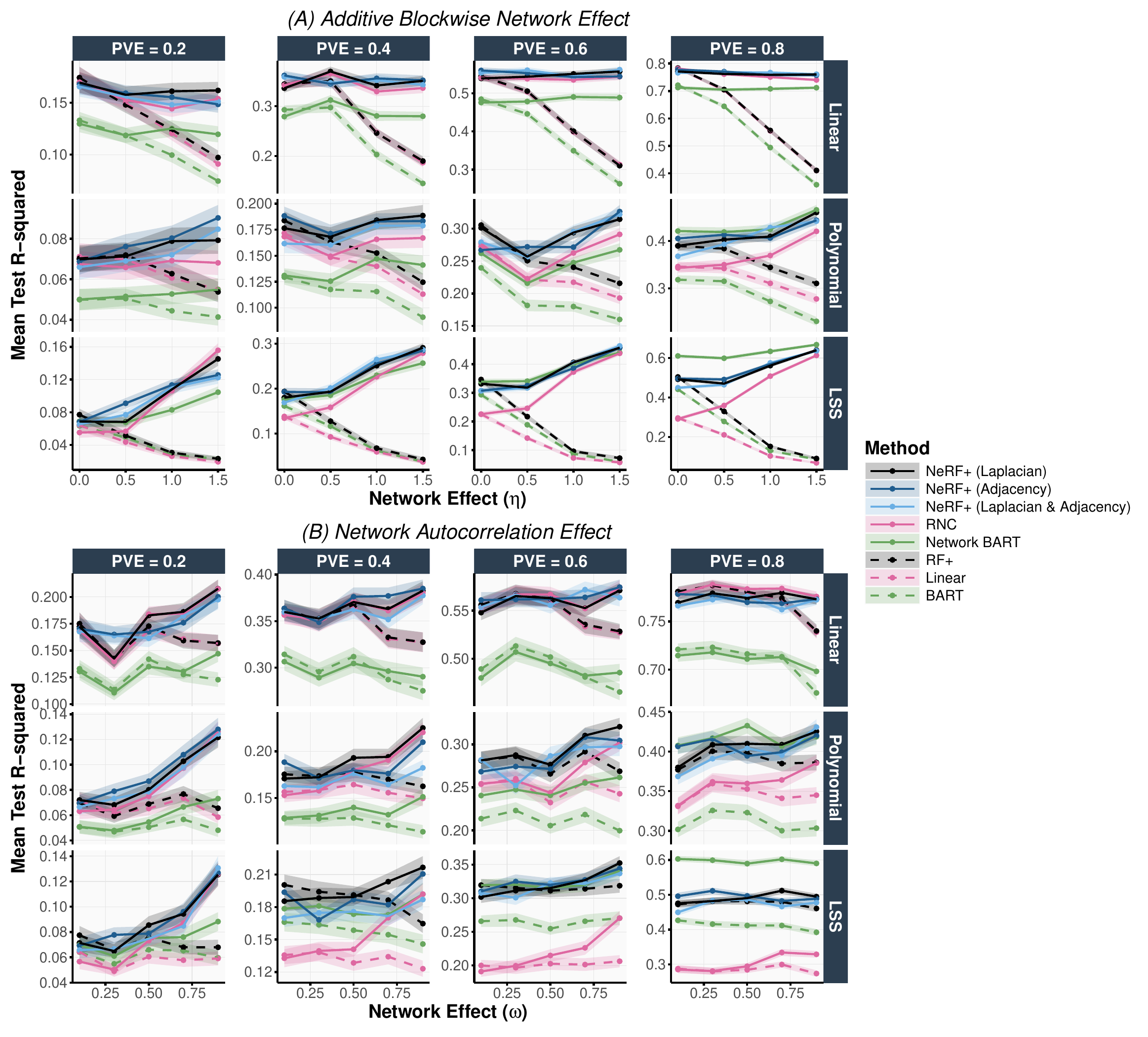}
    \caption{Prediction accuracy, as measured by the test $R^2$ value (y-axis), when varying the network embedding method under (A) the additive blockwise network effect simulation and (B) the network autocorrelation effect simulation with i.i.d.\ standard normal features. Results are shown across different underlying functional forms (rows), noise levels (columns), and varying strengths of the network effect (x-axis). Test $R^2$ values are averaged across 100 simulation replicates with shaded regions denoting $\pm 1 SE$.}
    \label{fig:sim_embedding_method}
\end{figure}

\begin{figure}[!tb]
    \spacingset{1}
    \centering
    \includegraphics[width=1\linewidth]{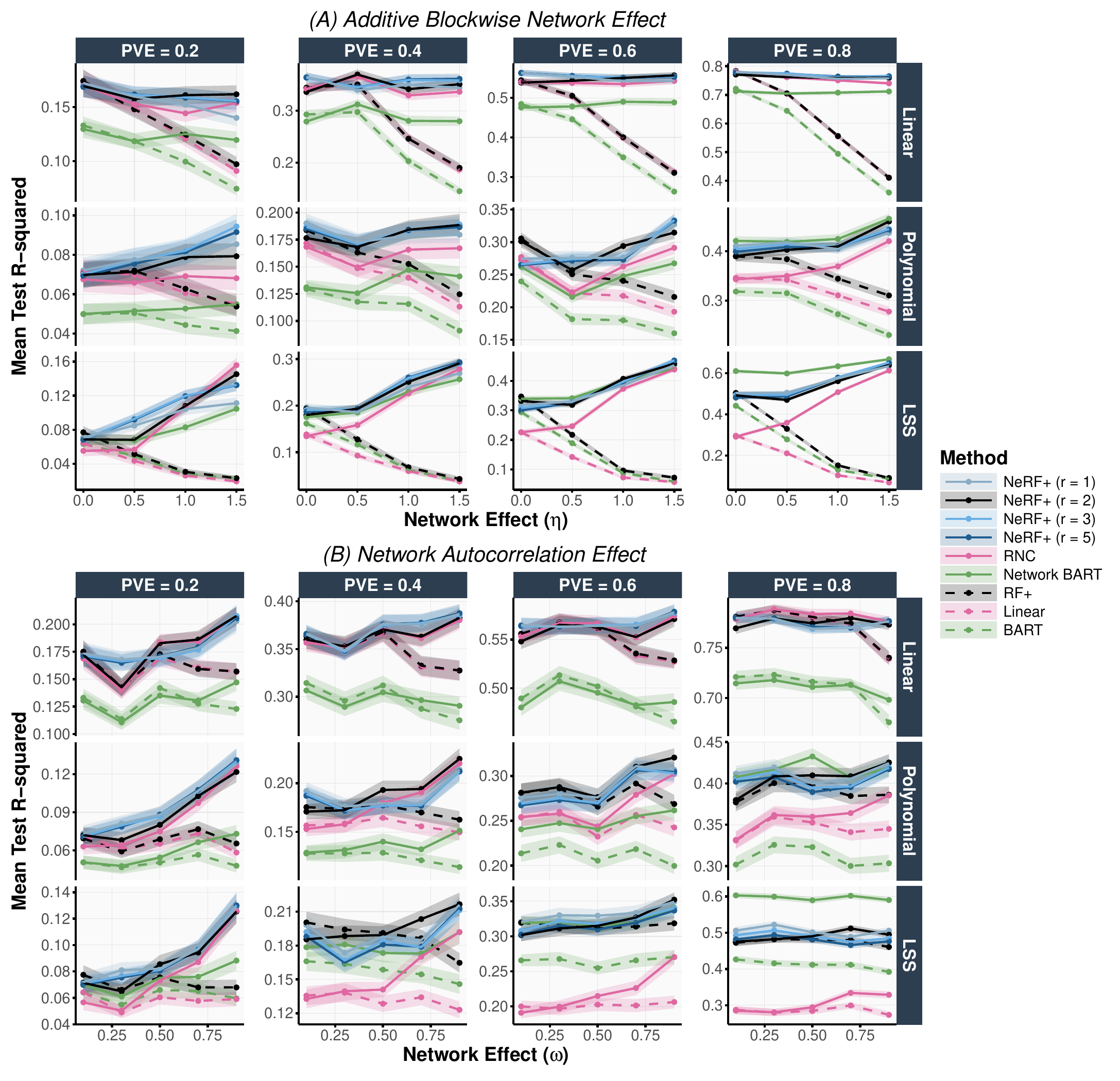}
    \caption{Prediction accuracy, as measured by the test $R^2$ value (y-axis), when varying the network embedding dimension ($r$) under (A) the additive blockwise network effect simulation and (B) the network autocorrelation effect simulation with i.i.d.\ standard normal features. Results are shown across different underlying functional forms (rows), noise levels (columns), and varying strengths of the network effect (x-axis). Test $R^2$ values are averaged across 100 simulation replicates with shaded regions denoting $\pm 1 SE$.}
    \label{fig:sim_embedding_dim}
\end{figure}

\begin{figure}[!tb]
    \spacingset{1}
    \centering
    \includegraphics[width=1\linewidth]{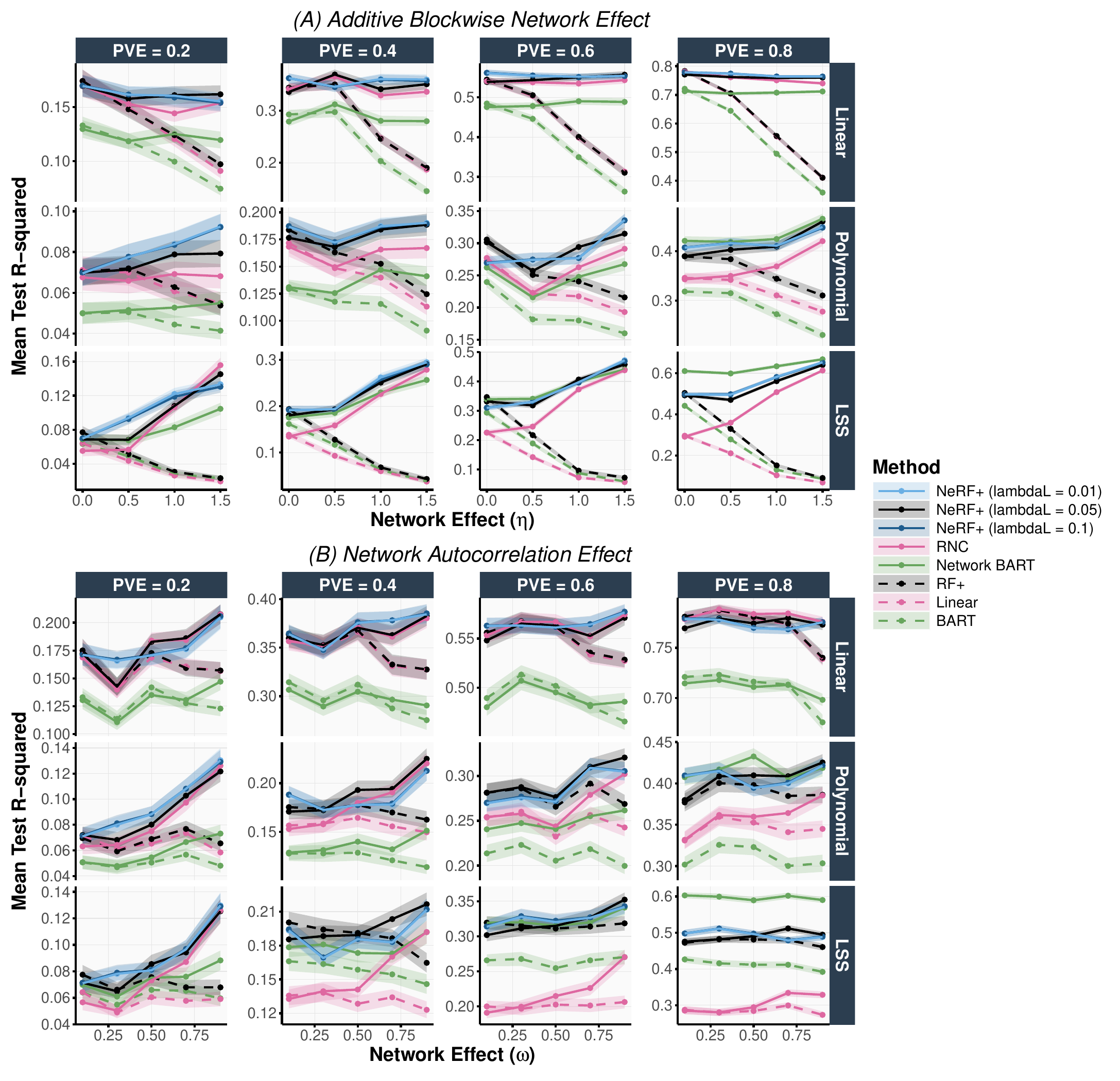}
    \caption{Prediction accuracy, as measured by the test $R^2$ value (y-axis), when varying the network Laplacian regularization ($\lambda_L$) under (A) the additive blockwise network effect simulation and (B) the network autocorrelation effect simulation with i.i.d.\ standard normal features. Results are shown across different underlying functional forms (rows), noise levels (columns), and varying strengths of the network effect (x-axis). Test $R^2$ values are averaged across 100 simulation replicates with shaded regions denoting $\pm 1 SE$.}
    \label{fig:sim_embedding_reg}
\end{figure}

\paragraph{Prediction results.}
Figures~\ref{fig:sim_embedding_method}--\ref{fig:sim_embedding_reg} show the prediction performance of \method~under these different choices for the network embedding method, dimension, and Laplacian regularization parameter.  
They show that \method~is relatively robust to the choice of network embedding method, the number of embedding dimensions, and the Laplacian regularization parameter, as evidenced by the similar test $R^2$ values across all choices. 
Importantly, the differences between NeRF+ with these different modeling choices are small relative to the differences between NeRF+ and the competing methods, and NeRF+ continues to consistently outperform existing methods across most simulation settings.

\clearpage
\subsection{Classification Simulations}\label{subsec:classification_sims}

\paragraph{Simulation Setup.} Up to this point, all simulations have focused on regression problems with continuous responses. \method, however, is a general framework that can accommodate a variety of loss functions and thus can be applied to a wide variety of supervised learning problems, including classification. 
To demonstrate this, we repeated the main simulations from Section~\ref{sec:sims} but with a binary response $y_i \in \{0, 1\}$ generated from a Bernoulli distribution with success probability $p_i = \text{logit}^{-1}(\alpha_i + f(\Vec{x}_i))$.
Here, the node effects $\alpha_i$ are generated as in the additive blockwise network effect (regression) simulations and $f$ can be a linear, polynomial, or Locally Spiky Sparse (LSS) function of the covariates $\Vec{x}_i$ as described in Section~\ref{sec:sims}.
This simulation is analogous to the additive blockwise network effect simulation described in Section~\ref{sec:sims}, but with a binary $y_i$ obtained by applying the logistic link function to the noiseless regression response $\alpha_i + f(\Vec{x}_i)$.
In this setting, we fit \method~with a logistic loss function and compared its performance to Network BART, RNC with a logistic loss, BART, and logistic regression.

\begin{figure}[!h]
    \spacingset{1}
    \centering
    \includegraphics[width=1\linewidth]{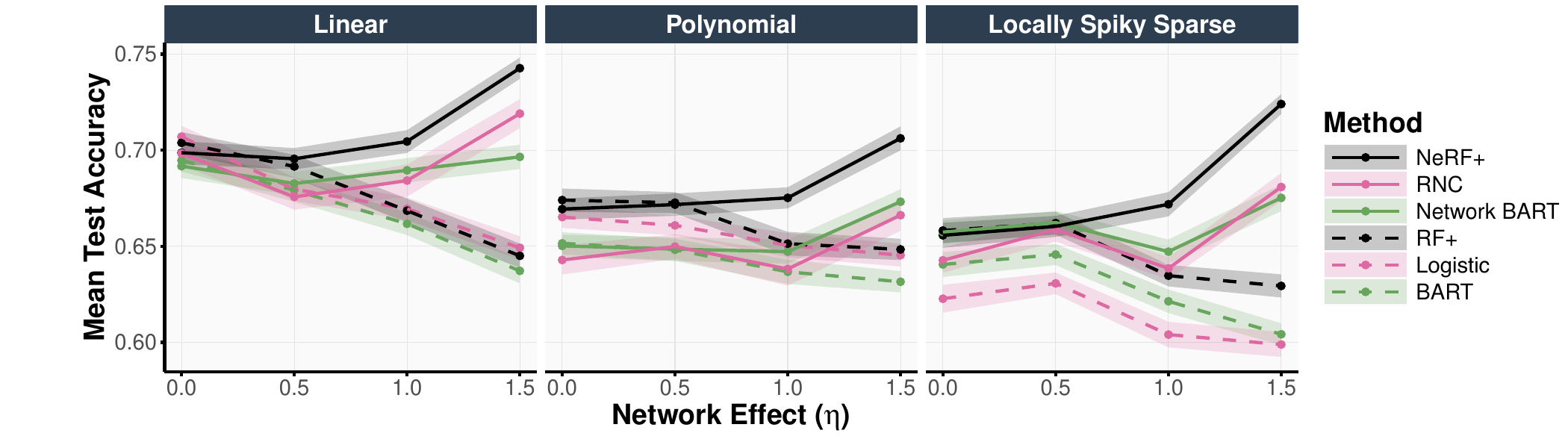}
    \caption{Test classification accuracy under logistic additive blockwise network effect simulation with i.i.d. standard normal features. Results are shown across different underlying functional forms (columns) and varying strengths of the network effect (x-axis). Classification accuracies are averaged across 100 simulation replicates with shaded regions denoting $\pm 1 SE$.}
    \label{fig:sim_logistic}
\end{figure}

\paragraph{Prediction Performance.}
As in the regression simulations, \method~consistently yields the best prediction performance, as measured via classification accuracy in Figure~\ref{fig:sim_logistic}, across all functional forms and strengths of the network effect.
Moreover, this gap in performance is particularly pronounced under the nonlinear (i.e., polynomial and LSS) settings, where \method~substantially outperforms all competing methods.

\begin{figure}[!h]
    \spacingset{1}
    \centering
    \includegraphics[width=\linewidth]{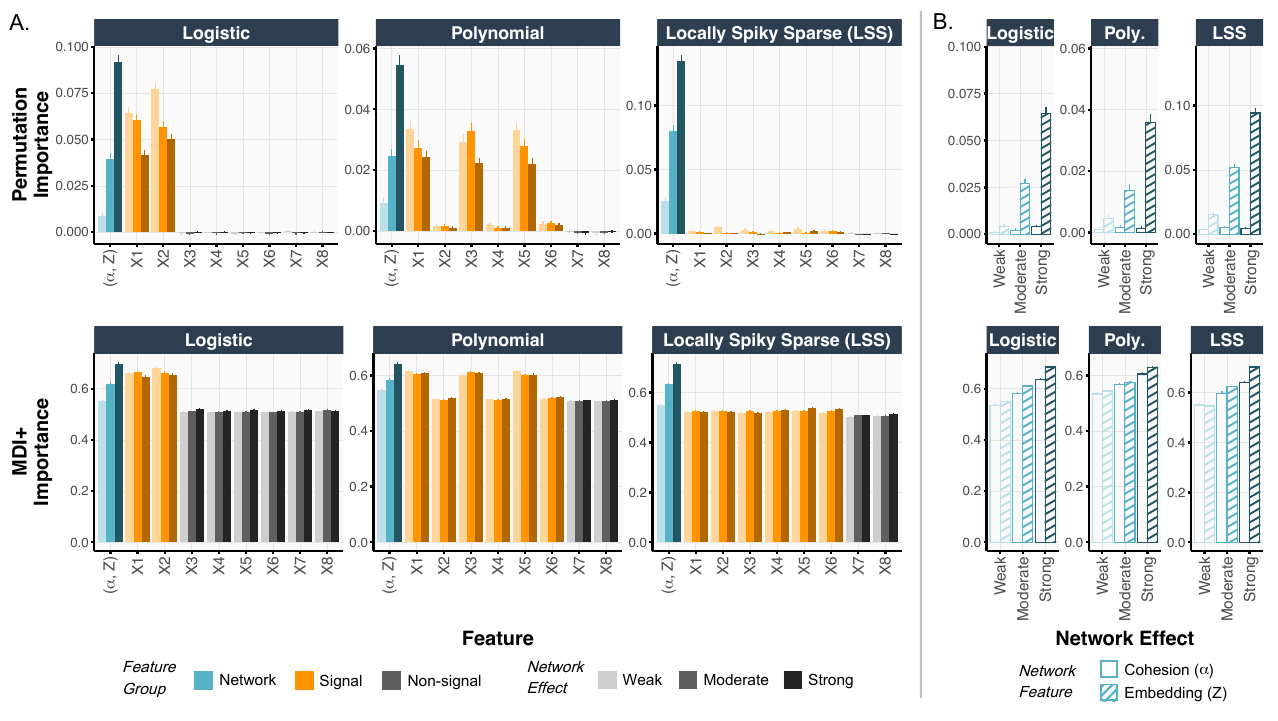}
    \caption{
        \method~global feature importances under the classification simulation setting. 
        (A)~Signal features exhibit higher importance than the non-signal features, and as the network effect increases (i.e., as the fill shade darkens), the estimated global importance of the network increases. Note: only the first 8 features are shown since the remaining features are non-signal and exhibit similarly low importances as the non-signal features shown. (B) Network embedding features have greater importance than the network cohesion in the additive blockwise network effect model. Results are averaged across 100 replicates with error bars denoting $\pm 1 SE$.
    }
    \label{fig:sim_importances_logistic}
\end{figure}

\paragraph{Global Feature Importances.}
In Figure~\ref{fig:sim_importances_logistic}, we show the corresponding global feature importances for \method~under the classification simulation setup. Similar to the regression simulations, we observe that both the permutation and \mdiplus~feature importances (i) effectively distinguish between signal and non-signal features, (ii) accurately show an increasing importance of the network as the simulated network effect strengthens, and (iii) reveal a higher importance from the network embedding features compared to the network cohesion component under the additive blockwise network effect model.

\clearpage

\subsection{Comparison with BAMDT}\label{subsec:bamdt}

Throughout this work, we have compared \method~to Network BART \citep{deshpande2022flexbart}, a network-assisted variant of BART which incorporates network-based decision splits in the BART trees by leveraging minimum spanning trees of the network.
There are, however, other network-assisted BART implementations, such as Bayesian Additive Semi-Multivariate Decision Trees (BAMDT) \citep{luo2022bamdt}, which similarly leverage minimum spanning trees of the network to allow for network-based decision splits in BART. 
Notably, BAMDT provides the ability to extract feature importances from the fitted model, which is not currently available in Network BART.
In what follows, we compare the prediction performance of \method, Network BART, and BAMDT, applied to the additive blockwise network effect simulation study from Section~\ref{sec:sims}. We also compare the global feature importances obtained from \method~and BAMDT, as Network BART does not currently provide a mechanism for extracting feature importances.

\paragraph{BAMDT Hyperparameters.} To match the hyperparameter settings used in Network BART, we fit BAMDT with 200 trees, 1000 burn-in iterations, 1000 posterior draws, and no thinning. 
Unlike Network BART, however, BAMDT requires the researcher to specify the probability of splitting on a covariate versus the network (\texttt{prob\_split\_by\_x}). 
We found that the prediction performance of BAMDT can be sensitive to this choice of \texttt{prob\_split\_by\_x}, and hence tried using the default value of \texttt{prob\_split\_by\_x} = 0.85, as well as two other values, \texttt{prob\_split\_by\_x} = 0.5 and \texttt{prob\_split\_by\_x} = 0.15.
All other BAMDT hyperparameters were set to their default values, shown in the demonstration code provided at \url{https://github.com/ztluostat/BAMDT}. 

\begin{figure}
    \spacingset{1}
    \centering
    \includegraphics[width=\linewidth]{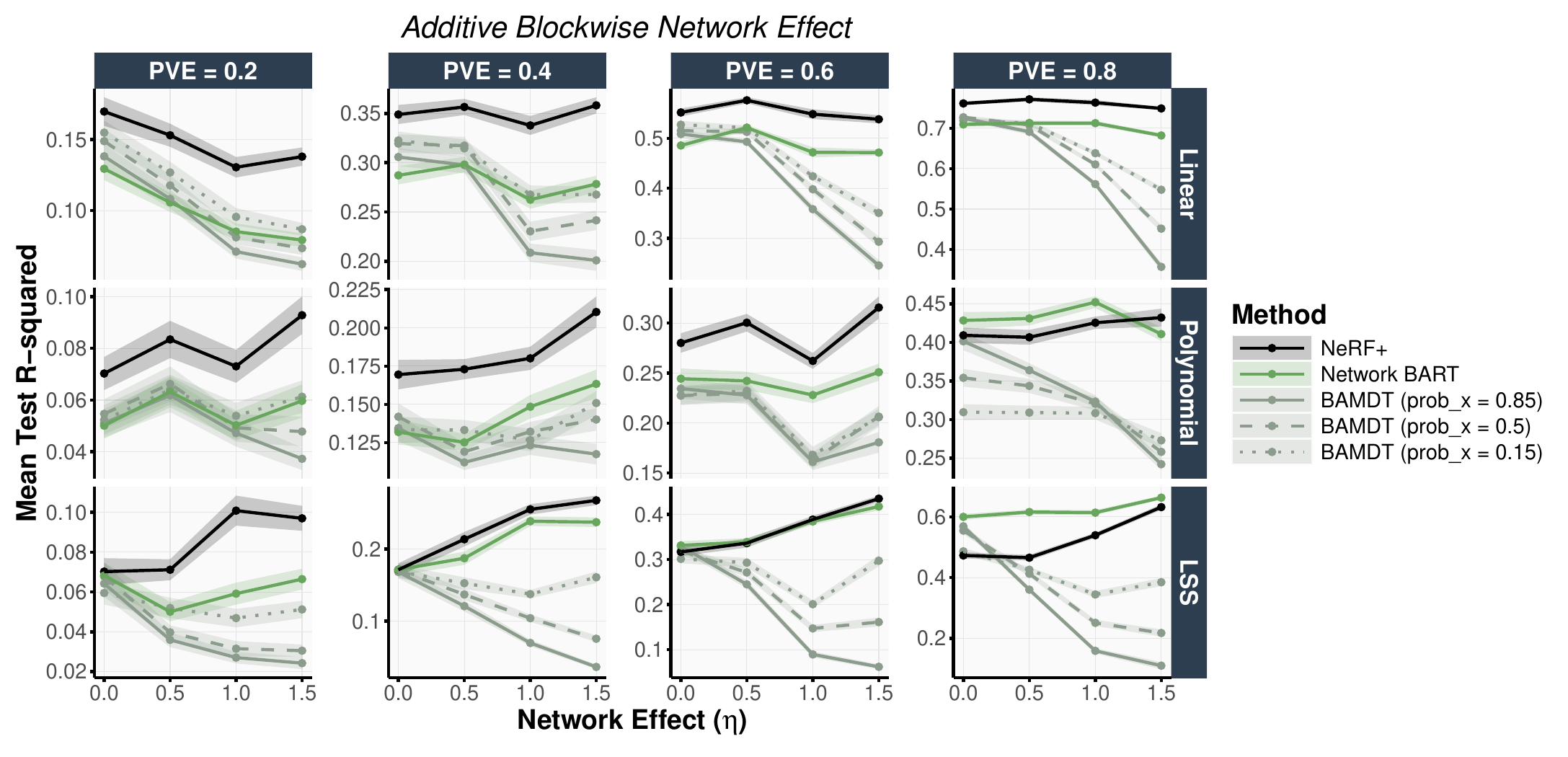}
    \caption{Prediction accuracy of \method, Network BART, and BAMDT, as measured by the test $R^2$ value (y-axis), under the additive blockwise network effect simulation. Results are shown across different underlying functional forms (rows), noise levels (columns), and varying strengths of the network effect (x-axis). Test $R^2$ values are averaged across 100 replicates with shaded regions denoting $\pm 1 SE$.}
    \label{fig:bamdt-prediction}
\end{figure}

\paragraph{Prediction Performance.} In Figure~\ref{fig:bamdt-prediction}, we compare the prediction performance of \method, Network BART, and BAMDT under the additive blockwise network effect simulation model from Section~\ref{sec:sims}. With the exception of the LSS simulation with very low noise levels ($\mathrm{PVE} = 0.8$), \method~consistently yields the best prediction performance. Network BART also generally outperforms BAMDT while the performance of BAMDT is often sensitive to the choice of \texttt{prob\_split\_by\_x}. In general, lower values of \texttt{prob\_split\_by\_x} in BAMDT yield better prediction performance as the network effect strengthens though its prediction performance is still generally worse than that of \method~and Network BART.

\begin{figure}
    \spacingset{1}
    \centering
    \includegraphics[width=\linewidth]{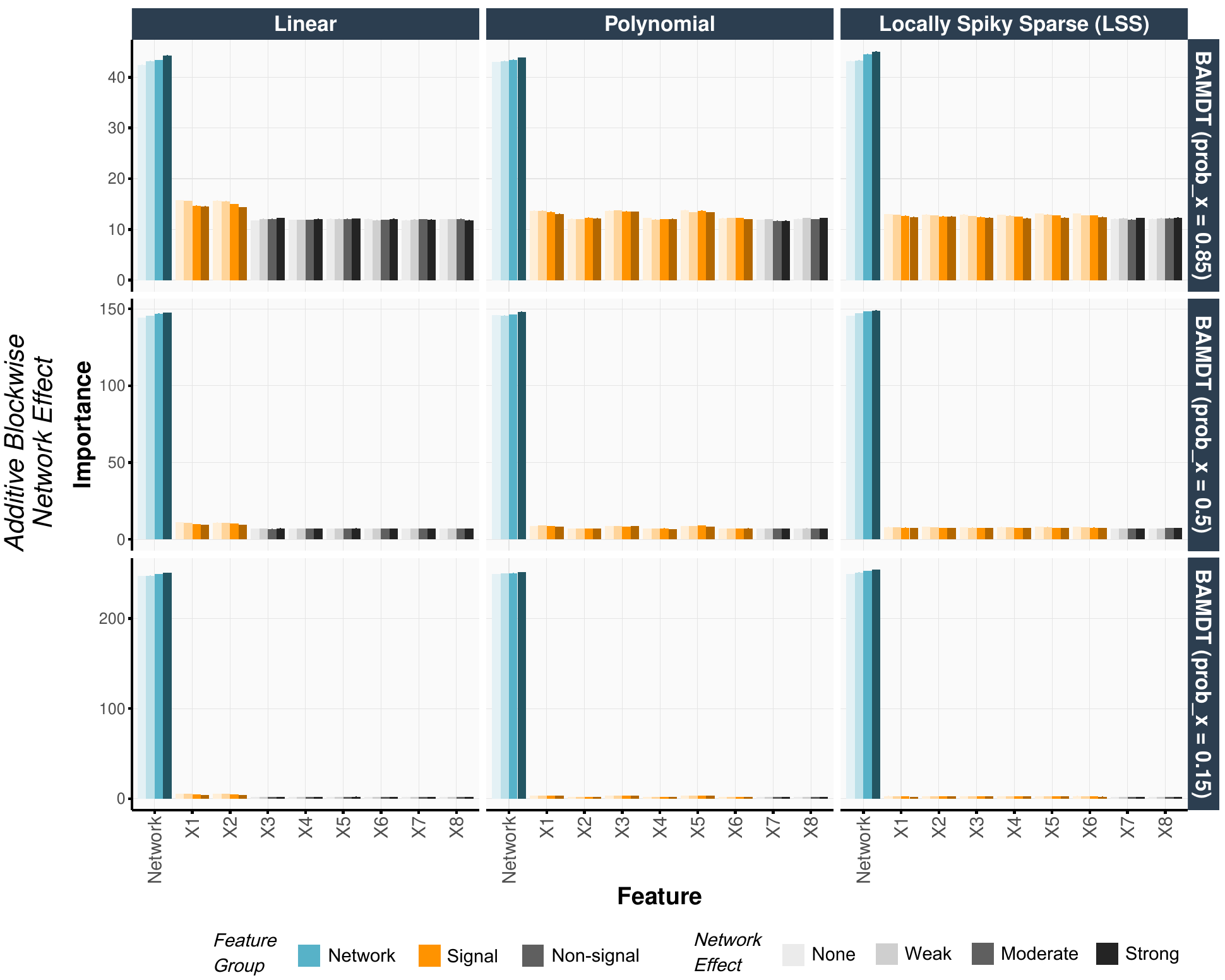}
    \caption{Global feature importances for BAMDT with varying \texttt{prob\_split\_by\_x} parameters under additive blockwise network effect simulation ($\mathrm{PVE} = 0.4$). Note: only the first 8 features are shown.  The remaining features are non-signal and have similar low importances to the non-signal features shown.  Results are averaged across 100 replicates with error bars denoting $\pm 1 SE$.}
    \label{fig:bamdt-fi}
\end{figure}

\paragraph{Global Feature Importances.} 
One of the advantages of using BAMDT over Network BART is the ability to easily extract global feature importances from the fitted model. 
These importances in BAMDT are computed as the proportion of times a feature is used to split in the trees. 
We summarize these BAMDT feature importances in Figure~\ref{fig:bamdt-fi} for the additive blockwise network effect simulation model ($\mathrm{PVE} = 0.4$) from Section~\ref{sec:sims}. 
Figure~\ref{fig:bamdt-fi} illustrates one of the main limitations of the BAMDT feature importance: even when the network effect is zero, the network appears to be the most important component in the model and exhibits by far the highest feature importance of any other predictor feature.
While this phenomenon is partially due to the choice of \texttt{prob\_split\_by\_x}, we see that even when \texttt{prob\_split\_by\_x} = 0.85, the network features are still much more important than any other predictor feature when the network effect is zero.
Another contributing factor to this inflated network importance is that there are often many more possible network-based splits than covariate-based splits in the BAMDT trees, which can lead to the network being selected more often for splitting even when it is not truly important \citep{strobl2007bias}.
While similar problems may also exist for random forest-based feature importances, the feature importances from random forests (and \method) account for \textit{how much} a feature contributes to the prediction accuracy of the model (e.g., through the decrease in impurity), rather than simply counting the number of times a feature is used in a split. Counting the number of splits, as done in BAMDT, can be misleading, as it may include many splits that do not substantially improve the prediction accuracy of the model.

\subsection{Approximating Sample Influences}\label{subsec:loo_sims}

\begin{figure}
    \spacingset{1}
    \centering
    \includegraphics[width=0.8\linewidth]{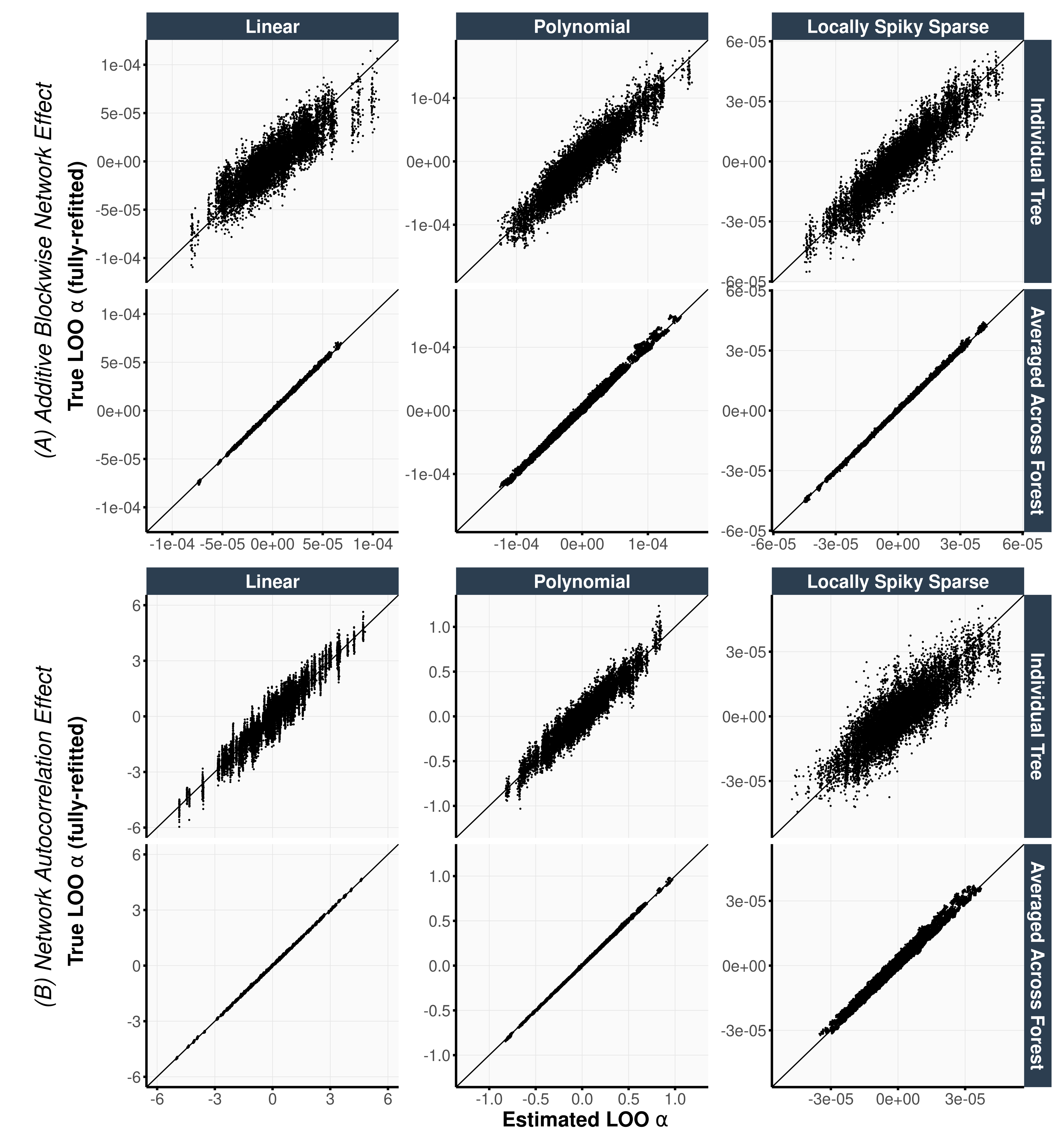}
    \caption{We compare the LOO estimates of $\Vec{\alpha}$ using the closed-form approximation \eqref{eq:loo} (x-axis) against the oracle obtained from leaving out each sample and refitting \method~(y-axis). In each subplot, the top row shows the coefficient estimates for a single tree while the bottom row shows the coefficient estimates, averaged across all trees in \method.}
    \label{fig:loo_alphas_supplement}
\end{figure}

\begin{figure}
    \spacingset{1}
    \centering
    \includegraphics[width=0.8\linewidth]{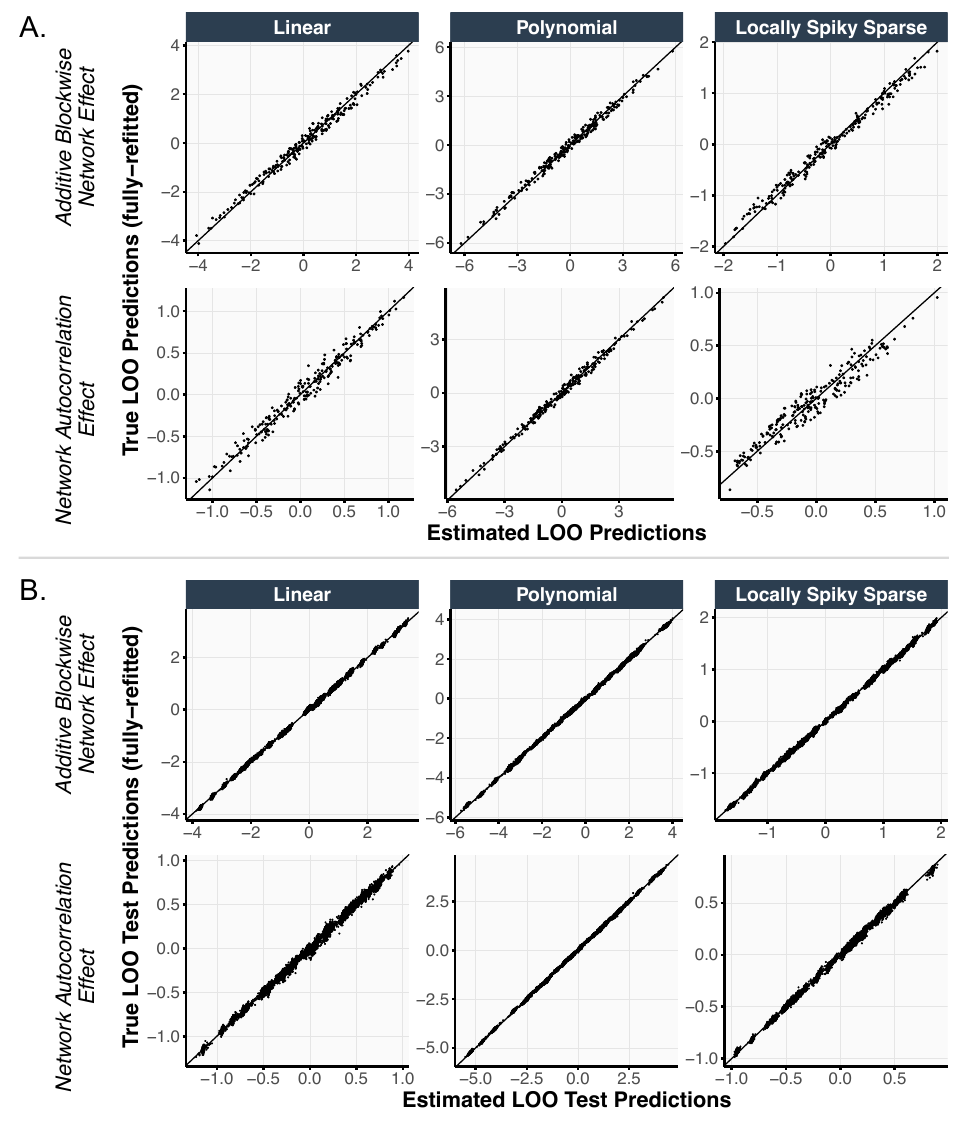}
    \caption{We show (A) the LOO training predictions and (B) the test predictions obtained using the estimated LOO coefficients \eqref{eq:loo} (x-axis) against the oracle test predictions obtained from leaving out each sample and refitting \method~(y-axis).}
    \label{fig:loo_predictions_supplement}
\end{figure}

In Figure~\ref{fig:loo_alphas_supplement}, we compare the LOO estimates of $\Vec{\alpha}$ using our closed-form approximation \eqref{eq:loo} against the oracle obtained from leaving out each sample and refitting \method. Similarly, in Figure~\ref{fig:loo_predictions_supplement}, we examine the LOO training predictions and the test predictions, obtained using our estimated LOO coefficients \eqref{eq:loo} against the oracle predictions obtained when leaving out each sample and refitting \method. The strong concordance between the estimated and true LOO coefficients and predictions helps to validate our closed-form approximation for efficiently estimating leave-one-out predictions without needing to refit \method~$n$ times.

\subsection{Computational Costs}\label{subsec:runtime_sims}

To provide practitioners with a sense of the computational costs of \method, we report the average runtime (in minutes) when applying \method~to regression datasets of varying sample sizes in Table~\ref{tab:runtimes}.
Here, we used the simulated data from the linear additive blockwise network effect model with $\mathrm{PVE} = 0.6$ and a network effect $\eta = 1$.
The runtimes are based on a single core of a 2023 MacBook Pro with Apple M3 Pro chip.

Though \method~is inevitably slower than an ordinary RF due to the additional steps of estimating the node effects $\Vec{\alpha} \in \R^n$ and fitting a penalized GLM on top of the RF tree structure, \method~is still relatively fast and scalable, compared to existing network-assisted methods such as Network BART.
As seen in Table~\ref{tab:runtimes}, when the sample size is 1000 or larger, Network BART (with 200 trees) takes more than an hour to run, while \method~(with 500 trees) takes less than 3 minutes.
We also note that both Network BART and \method~in this computational comparison were run with a fixed set of hyperparameters, and thus the reported runtimes do not include the additional time required for hyperparameter tuning via cross-validation.
However, given that these results are reported using only one core, we note that \method~can be easily parallelized across either trees in the forest or hyperparameter combinations to maintain a reasonable runtime even when hyperparameter tuning is performed.

\begin{table}
    \spacingset{1}
    \centering
    \begin{tabular}{ccc}
    \toprule
    \textbf{Sample Size} ($n$) & \textbf{\method} & \textbf{Network BART}\\
    \midrule
    200 & 0.10 (0.000) & 1.47 (0.004)\\
    300 & 0.13 (0.000) & 4.77 (0.014)\\
    500 & 0.24 (0.000) & 21.21 (0.061)\\
    1000 & 0.95 (0.001) & $>$ 60.00\\
    1500 & 2.67 (0.002) & $>$ 60.00\\
    \bottomrule
    \end{tabular}
    \caption{Average runtime (in minutes) for \method~and Network BART on simulated data with varying sample sizes. Results are averaged across 50 replicates. Standard errors are reported in parentheses.}
    \label{tab:runtimes}
\end{table}

\section{Case Study Details and Results}\label{app:case_study}
In this section, we provide additional details and results for the school conflict reduction and Philadelphia crime case studies.  

For both case studies, we tuned the regularization parameters in \method~by a grid search across 20 values, logarithmically spaced between $10^{-3}$ and $10^3$, for each regularization parameter $\lambda_{\alpha}, \lambda_{\beta}, \lambda_{\gamma}$. 
In the Philadelphia crime case study, we used the top two components from the regularized Laplacian spectral embedding of the network for $\Mat{Z}$, as in the simulations.
In the school conflict reduction case study, since many schools had four grades of students, we used the top four components from the regularized Laplacian spectral embedding of the network for $\mathbf{Z}$. 
Although not all schools had four grades, we found in practice that choosing a larger number of network components typically does not hurt performance. 
All other hyperparameters and implementation details were the same as those used in the simulations (see Appendix~\ref{app:sims}).

\subsection{School Conflict Reduction Case Study}

\begin{figure}
    \spacingset{1}
    \centering
    \includegraphics[width=1\textwidth]{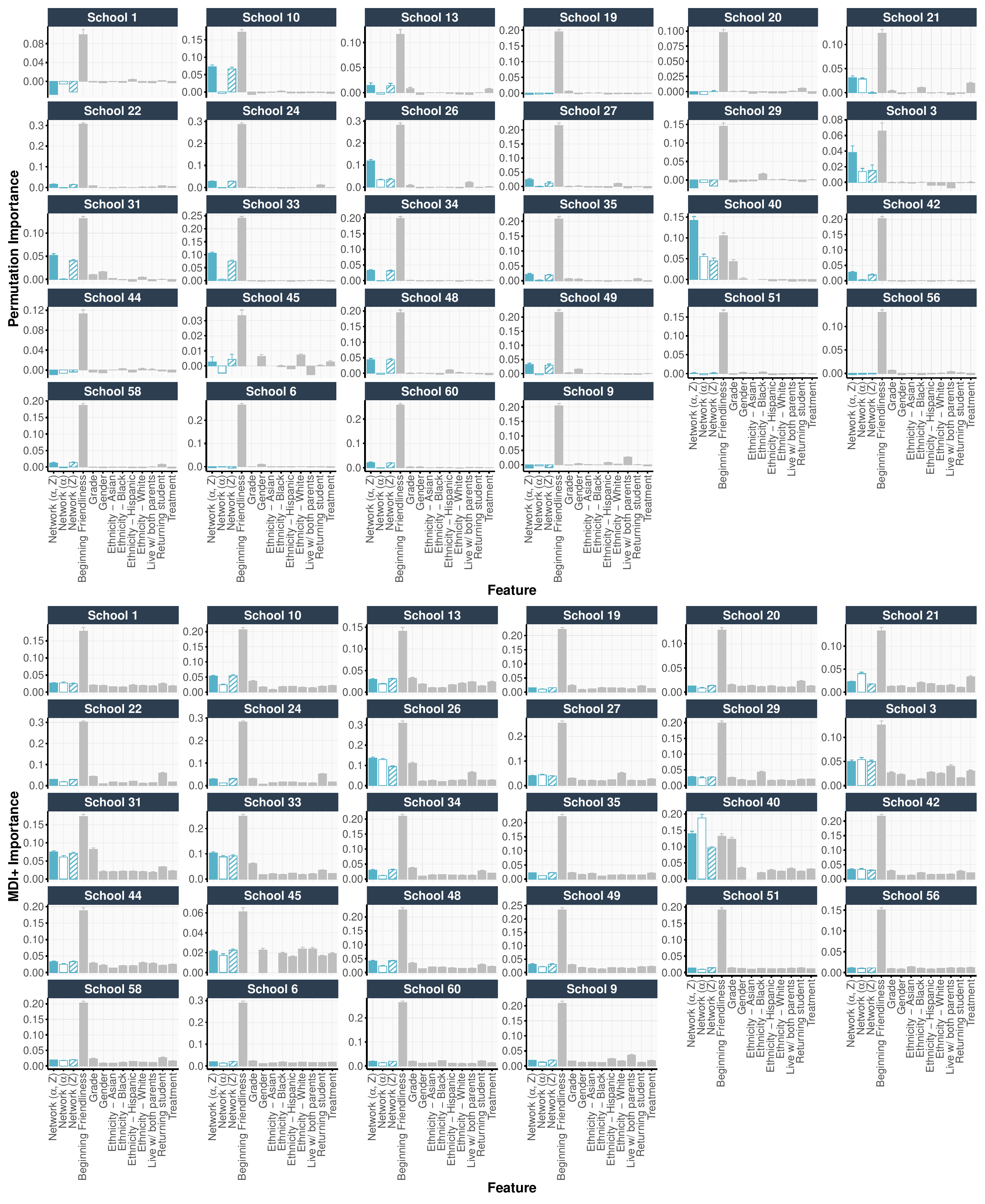}
    \caption{\method~global feature importances, as measured via permutation (top) and \mdiplus~(bottom) importance, for each school in the school conflict reduction case study.}
    \label{fig:school_conflict_globalfi_nerfplus}
\end{figure}

In Table~\ref{tab:school_conflict_data}, we provide the full prediction performance results for each school in the school conflict reduction case study. The table includes the number of samples $n$ and the test $R^2$ values from each method and school. We also provide the test $R^2$ averaged across all schools for each method in the last row. Table~\ref{tab:school_conflict_ranks} summarizes these prediction results, showing the number of times each method was the best (i.e., rank 1), second best (i.e., rank 2), etc. across the 28 schools. Here, we highlight that though RNC was the best method for the most schools (17), \method~was always in the top three best-performing methods. This consistently strong performance is a reflection of \method's flexibility to capture both linear and nonlinear structures unlike RNC, which fails to account for nonlinear relationships in the data. Lastly, for completeness, we provide the \method~global feature importances for each school in Figure~\ref{fig:school_conflict_globalfi_nerfplus}.

\begin{sidewaystable}
\spacingset{1}
\centering
\small
\begin{tabular}[t]{ccccccccc}
\toprule
\textbf{School} & \textbf{$n$} & \textbf{\method} & \textbf{RNC} & \makecell{\textbf{Network}\\\textbf{BART}} & \textbf{\rfplus} & \makecell{\textbf{Linear}\\\textbf{Regression}} & \textbf{BART} & \textbf{RF}\\
\midrule
\cellcolor{gray!10}{1} & \cellcolor{gray!10}{180} & \cellcolor{gray!10}{0.084 (0.020)} & \cellcolor{gray!10}{\textit{0.105 (0.021)}} & \cellcolor{gray!10}{0.056 (0.019)} & \cellcolor{gray!10}{0.052 (0.020)} & \cellcolor{gray!10}{0.024 (0.023)} & \cellcolor{gray!10}{0.028 (0.020)} & \cellcolor{gray!10}{0.050 (0.019)}\\
3 & 89 & \textit{0.011 (0.033)} & -0.009 (0.033) & -0.063 (0.031) & -0.054 (0.032) & -0.168 (0.040) & -0.118 (0.033) & -0.082 (0.033)\\
\cellcolor{gray!10}{6} & \cellcolor{gray!10}{342} & \cellcolor{gray!10}{0.265 (0.011)} & \cellcolor{gray!10}{\textit{0.280 (0.011)}} & \cellcolor{gray!10}{0.251 (0.012)} & \cellcolor{gray!10}{0.259 (0.012)} & \cellcolor{gray!10}{0.253 (0.012)} & \cellcolor{gray!10}{0.241 (0.012)} & \cellcolor{gray!10}{0.233 (0.012)}\\
9 & 296 & \textit{0.223 (0.014)} & 0.219 (0.015) & 0.212 (0.014) & 0.221 (0.014) & 0.207 (0.015) & 0.206 (0.014) & 0.197 (0.014)\\
\cellcolor{gray!10}{10} & \cellcolor{gray!10}{214} & \cellcolor{gray!10}{0.179 (0.014)} & \cellcolor{gray!10}{0.176 (0.014)} & \cellcolor{gray!10}{\textit{0.184 (0.015)}} & \cellcolor{gray!10}{0.156 (0.015)} & \cellcolor{gray!10}{0.147 (0.015)} & \cellcolor{gray!10}{0.157 (0.015)} & \cellcolor{gray!10}{0.132 (0.015)}\\
13 & 174 & \textit{0.094 (0.022)} & 0.069 (0.022) & 0.088 (0.021) & 0.087 (0.021) & 0.010 (0.025) & 0.068 (0.022) & 0.085 (0.022)\\
\cellcolor{gray!10}{19} & \cellcolor{gray!10}{390} & \cellcolor{gray!10}{0.198 (0.013)} & \cellcolor{gray!10}{\textit{0.213 (0.013)}} & \cellcolor{gray!10}{0.185 (0.013)} & \cellcolor{gray!10}{0.193 (0.013)} & \cellcolor{gray!10}{0.188 (0.013)} & \cellcolor{gray!10}{0.174 (0.013)} & \cellcolor{gray!10}{0.137 (0.013)}\\
20 & 365 & 0.109 (0.012) & 0.110 (0.012) & \textit{0.113 (0.013)} & 0.095 (0.012) & 0.087 (0.013) & 0.099 (0.012) & 0.062 (0.012)\\
\cellcolor{gray!10}{21} & \cellcolor{gray!10}{236} & \cellcolor{gray!10}{0.111 (0.015)} & \cellcolor{gray!10}{\textit{0.123 (0.015)}} & \cellcolor{gray!10}{0.093 (0.016)} & \cellcolor{gray!10}{0.105 (0.016)} & \cellcolor{gray!10}{0.108 (0.017)} & \cellcolor{gray!10}{0.077 (0.017)} & \cellcolor{gray!10}{0.026 (0.017)}\\
22 & 481 & 0.327 (0.008) & \textit{0.332 (0.009)} & 0.323 (0.008) & 0.327 (0.008) & 0.324 (0.008) & 0.305 (0.008) & 0.264 (0.008)\\
\cellcolor{gray!10}{24} & \cellcolor{gray!10}{623} & \cellcolor{gray!10}{0.305 (0.008)} & \cellcolor{gray!10}{0.305 (0.008)} & \cellcolor{gray!10}{0.289 (0.008)} & \cellcolor{gray!10}{\textit{0.305 (0.008)}} & \cellcolor{gray!10}{0.302 (0.008)} & \cellcolor{gray!10}{0.288 (0.008)} & \cellcolor{gray!10}{0.281 (0.008)}\\
26 & 129 & 0.365 (0.014) & \textit{0.387 (0.014)} & 0.346 (0.015) & 0.358 (0.015) & 0.340 (0.016) & 0.314 (0.016) & 0.306 (0.015)\\
\cellcolor{gray!10}{27} & \cellcolor{gray!10}{158} & \cellcolor{gray!10}{0.188 (0.017)} & \cellcolor{gray!10}{0.185 (0.017)} & \cellcolor{gray!10}{\textit{0.196 (0.017)}} & \cellcolor{gray!10}{0.152 (0.017)} & \cellcolor{gray!10}{0.145 (0.018)} & \cellcolor{gray!10}{0.151 (0.018)} & \cellcolor{gray!10}{0.159 (0.016)}\\
29 & 173 & 0.137 (0.016) & \textit{0.145 (0.016)} & 0.131 (0.016) & 0.126 (0.016) & 0.108 (0.017) & 0.104 (0.016) & 0.092 (0.016)\\
\cellcolor{gray!10}{31} & \cellcolor{gray!10}{267} & \cellcolor{gray!10}{0.207 (0.010)} & \cellcolor{gray!10}{\textit{0.213 (0.010)}} & \cellcolor{gray!10}{0.206 (0.010)} & \cellcolor{gray!10}{0.199 (0.010)} & \cellcolor{gray!10}{0.176 (0.011)} & \cellcolor{gray!10}{0.192 (0.011)} & \cellcolor{gray!10}{0.168 (0.012)}\\
33 & 262 & 0.292 (0.011) & \textit{0.301 (0.012)} & 0.284 (0.012) & 0.287 (0.012) & 0.279 (0.012) & 0.265 (0.012) & 0.256 (0.012)\\
\cellcolor{gray!10}{34} & \cellcolor{gray!10}{366} & \cellcolor{gray!10}{0.214 (0.011)} & \cellcolor{gray!10}{\textit{0.228 (0.011)}} & \cellcolor{gray!10}{0.213 (0.012)} & \cellcolor{gray!10}{0.213 (0.011)} & \cellcolor{gray!10}{0.212 (0.012)} & \cellcolor{gray!10}{0.200 (0.012)} & \cellcolor{gray!10}{0.177 (0.012)}\\
35 & 278 & \textit{0.207 (0.012)} & 0.200 (0.012) & 0.180 (0.013) & 0.183 (0.012) & 0.171 (0.014) & 0.171 (0.013) & 0.176 (0.012)\\
\cellcolor{gray!10}{40} & \cellcolor{gray!10}{83} & \cellcolor{gray!10}{\textit{0.222 (0.019)}} & \cellcolor{gray!10}{0.182 (0.023)} & \cellcolor{gray!10}{0.203 (0.020)} & \cellcolor{gray!10}{0.165 (0.023)} & \cellcolor{gray!10}{0.128 (0.028)} & \cellcolor{gray!10}{0.200 (0.020)} & \cellcolor{gray!10}{0.163 (0.022)}\\
42 & 296 & 0.202 (0.012) & \textit{0.220 (0.012)} & 0.188 (0.013) & 0.191 (0.012) & 0.194 (0.013) & 0.182 (0.013) & 0.135 (0.013)\\
\cellcolor{gray!10}{44} & \cellcolor{gray!10}{152} & \cellcolor{gray!10}{0.141 (0.019)} & \cellcolor{gray!10}{\textit{0.144 (0.019)}} & \cellcolor{gray!10}{0.120 (0.019)} & \cellcolor{gray!10}{0.114 (0.019)} & \cellcolor{gray!10}{0.054 (0.020)} & \cellcolor{gray!10}{0.091 (0.020)} & \cellcolor{gray!10}{0.119 (0.019)}\\
45 & 186 & 0.033 (0.020) & -0.008 (0.021) & 0.011 (0.020) & 0.037 (0.020) & -0.037 (0.021) & 0.015 (0.020) & \textit{0.050 (0.018)}\\
\cellcolor{gray!10}{48} & \cellcolor{gray!10}{223} & \cellcolor{gray!10}{0.202 (0.013)} & \cellcolor{gray!10}{\textit{0.211 (0.013)}} & \cellcolor{gray!10}{0.164 (0.014)} & \cellcolor{gray!10}{0.177 (0.013)} & \cellcolor{gray!10}{0.176 (0.014)} & \cellcolor{gray!10}{0.155 (0.014)} & \cellcolor{gray!10}{0.112 (0.014)}\\
49 & 219 & 0.213 (0.013) & \textit{0.237 (0.013)} & 0.199 (0.014) & 0.208 (0.013) & 0.201 (0.014) & 0.182 (0.014) & 0.140 (0.015)\\
\cellcolor{gray!10}{51} & \cellcolor{gray!10}{323} & \cellcolor{gray!10}{0.176 (0.012)} & \cellcolor{gray!10}{\textit{0.185 (0.012)}} & \cellcolor{gray!10}{0.152 (0.013)} & \cellcolor{gray!10}{0.169 (0.012)} & \cellcolor{gray!10}{0.161 (0.012)} & \cellcolor{gray!10}{0.141 (0.013)} & \cellcolor{gray!10}{0.143 (0.013)}\\
56 & 436 & 0.110 (0.011) & \textit{0.122 (0.011)} & 0.099 (0.011) & 0.109 (0.011) & 0.109 (0.011) & 0.085 (0.012) & 0.045 (0.012)\\
\cellcolor{gray!10}{58} & \cellcolor{gray!10}{442} & \cellcolor{gray!10}{0.185 (0.012)} & \cellcolor{gray!10}{\textit{0.197 (0.012)}} & \cellcolor{gray!10}{0.178 (0.012)} & \cellcolor{gray!10}{0.179 (0.012)} & \cellcolor{gray!10}{0.180 (0.012)} & \cellcolor{gray!10}{0.173 (0.012)} & \cellcolor{gray!10}{0.120 (0.012)}\\
60 & 508 & \textit{0.265 (0.008)} & 0.259 (0.009) & 0.254 (0.008) & 0.260 (0.008) & 0.252 (0.009) & 0.240 (0.008) & 0.234 (0.008)\\
\midrule
\cellcolor{gray!10}{\textbf{Mean}} & \cellcolor{gray!10}{--} & \cellcolor{gray!10}{\textbf{0.188}} & \cellcolor{gray!10}{\textbf{0.190}} & \cellcolor{gray!10}{\textbf{0.173}} & \cellcolor{gray!10}{\textbf{0.174}} & \cellcolor{gray!10}{\textbf{0.155}} & \cellcolor{gray!10}{\textbf{0.157}} & \cellcolor{gray!10}{\textbf{0.142}}\\
\bottomrule
\end{tabular}
\caption{Number of samples $n$ and test $R^2$ values from prediction methods applied to each school in the school conflict reduction case study. Test $R^2$ values are averaged across 100 training-test splits with standard errors in parentheses. Last row shows the test $R^2$ averaged across all schools.}
\label{tab:school_conflict_data}
\end{sidewaystable}

\begin{table}[h!]
\spacingset{1}
\centering
\small
\begin{tabular}[t]{cccccccc}
\toprule
\textbf{Method} & \textbf{Rank 1} & \textbf{Rank 2} & \textbf{Rank 3} & \textbf{Rank 4} & \textbf{Rank 5} & \textbf{Rank 6} & \textbf{Rank 7}\\
\midrule
\cellcolor{gray!10}{NeRF+} & \cellcolor{gray!10}{6} & \cellcolor{gray!10}{19} & \cellcolor{gray!10}{3} & \cellcolor{gray!10}{0} & \cellcolor{gray!10}{0} & \cellcolor{gray!10}{0} & \cellcolor{gray!10}{0}\\
RNC & 17 & 3 & 5 & 1 & 1 & 1 & 0\\
\cellcolor{gray!10}{Network BART} & \cellcolor{gray!10}{3} & \cellcolor{gray!10}{2} & \cellcolor{gray!10}{5} & \cellcolor{gray!10}{6} & \cellcolor{gray!10}{12} & \cellcolor{gray!10}{0} & \cellcolor{gray!10}{0}\\
RF+ & 1 & 4 & 11 & 7 & 5 & 0 & 0\\
\cellcolor{gray!10}{Linear Regression} & \cellcolor{gray!10}{0} & \cellcolor{gray!10}{0} & \cellcolor{gray!10}{3} & \cellcolor{gray!10}{8} & \cellcolor{gray!10}{6} & \cellcolor{gray!10}{3} & \cellcolor{gray!10}{8}\\
BART & 0 & 0 & 1 & 3 & 1 & 22 & 1\\
\cellcolor{gray!10}{RF} & \cellcolor{gray!10}{1} & \cellcolor{gray!10}{0} & \cellcolor{gray!10}{0} & \cellcolor{gray!10}{3} & \cellcolor{gray!10}{3} & \cellcolor{gray!10}{2} & \cellcolor{gray!10}{19}\\
\bottomrule
\end{tabular}
\caption{Frequency of each prediction method achieving each rank (1 = best method, 7 = worst method) across the 28 schools in the school conflict reduction case study.}
\label{tab:school_conflict_ranks}
\end{table}

\subsection{Philadelphia Crime Case Study}

\begin{figure}
    \spacingset{1}
    \centering
    \includegraphics[width=0.5\linewidth]{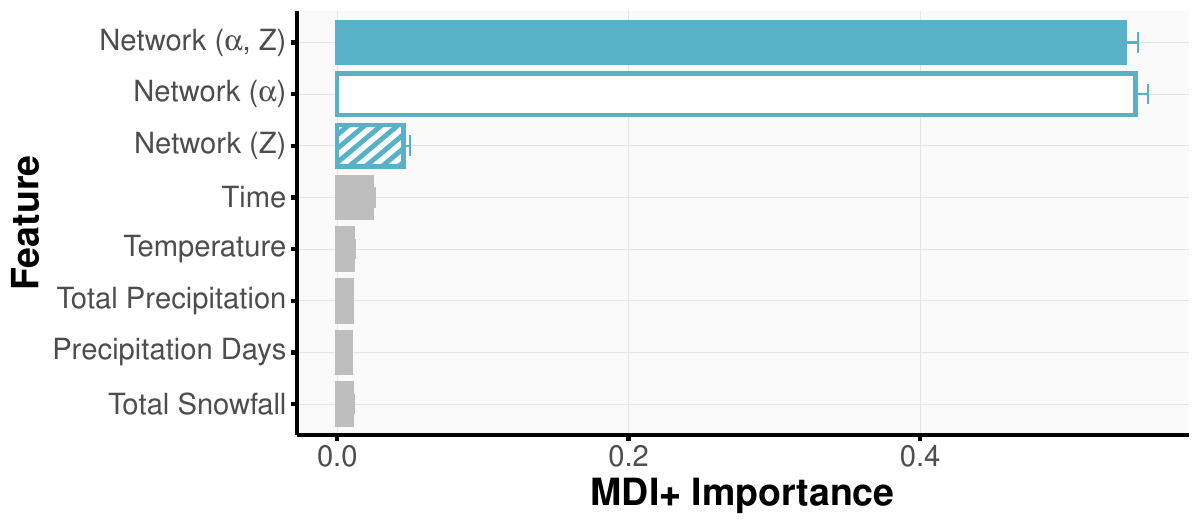}
    \caption{\mdiplus~feature importances for the \method~fit (using 1\% of the data for training and both the network cohesion and network embeddings) in the Philadelphia crime study.}
    \label{fig:philly_crime_mdiplus}
\end{figure}

In Figure~\ref{fig:philly_crime_mdiplus}, we show the \mdiplus~feature importances for the \method~fit in the Philadelphia crime case study. Similar to the permutation feature importances in Figure~\ref{fig:philly_crime}, the \mdiplus~feature importances reveal that most of the network's importance stems from the network cohesion component, as opposed to the network embeddings.

\section{Split-Conformal Prediction Intervals}\label{app:conformal}
As briefly mentioned in Section~\ref{subsec:predictions}, valid prediction intervals can be constructed by directly applying the split-conformal prediction framework to \method~(see Algorithm 1 in \citet{lunde2023conformal}). In this section, we empirically demonstrate that these split-conformal prediction intervals achieve the desired coverage in both simulations and real data.

\begin{figure}[tb]
    \spacingset{1}
    \centering
    \includegraphics[width=0.7\linewidth]{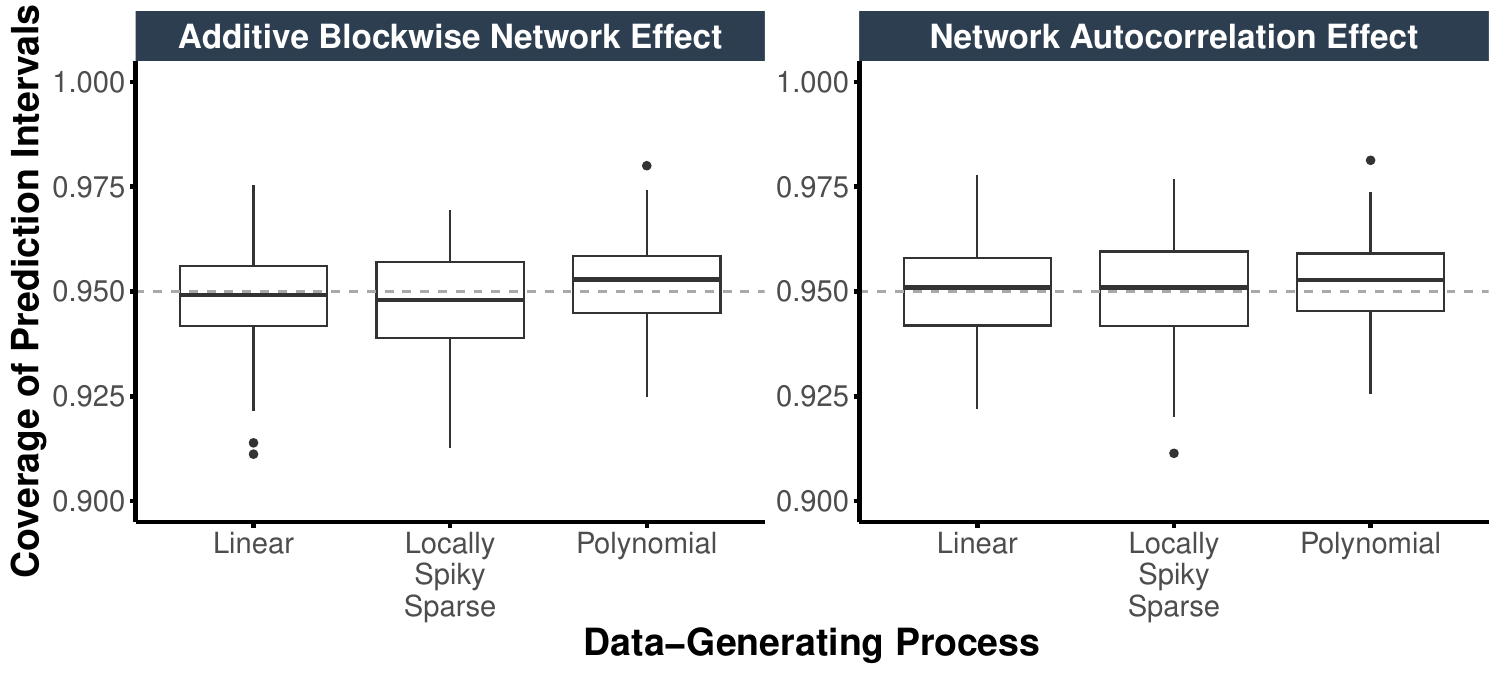}
    \caption{Coverage of split-conformal prediction intervals for \method~across various simulation settings. The dashed line indicates the desired nominal coverage level of 95\%. Each simulation replicate was evaluated using 10,000 test data points. Boxplots summarize results across 100 simulation replicates.}
    \label{fig:conformal_sims}
\end{figure}

\paragraph{Simulations.} 
Under each of the simulation settings described in Section~\ref{sec:sims}, we first split the simulated data into a training set (50\%) and a calibration set (50\%) according to the split-conformal procedure. 
We then evaluated the coverage of the resulting \method~prediction intervals (at level $\alpha = 0.05$) by generating 10,000 new test points from the same data generating process and computing the proportion of these test points that fall within the estimated prediction intervals.
This procedure was repeated 100 times for each simulation setting to obtain a distribution of coverage values, and we present the results in Figure~\ref{fig:conformal_sims}.
These simulations demonstrate that the coverage of the prediction intervals is close to the nominal level of 95\% across all simulation settings, demonstrating the validity of the split-conformal prediction intervals for \method. Moreover, in Table~\ref{tab:conformal_widths}, we summarize the average width of the prediction intervals across simulation settings and compare them to the average widths of 95\% posterior prediction intervals obtained from BART and Network BART. Across all simulation settings, the average width of the split-conformal prediction intervals for \method~is comparable to the average widths of the posterior prediction intervals from BART and Network BART.

\begin{table}[tb]
    \spacingset{1.1}
    \centering
    \small
    \begin{tabular}{lccc}
    \toprule
    \textbf{Data-Generating Process} & \textbf{\method} & \textbf{BART} & \textbf{Network BART}\\
    \midrule
    Linear Additive Blockwise Network Effect & \textbf{5.57} (0.03) & 6.64 (0.04) & 5.87 (0.03)\\
    Linear Network Autocorrelation Effect & \textbf{4.78} (0.03) & 5.07 (0.03) & 5.10 (0.03)\\
    LSS Additive Blockwise Network Effect & \textbf{4.30} (0.02) & 5.40 (0.03) & 4.32 (0.03)\\
    LSS Network Autocorrelation Effect & \textbf{3.09} (0.02) & 3.27 (0.02) & 3.12 (0.02)\\
    Polynomial Additive Blockwise Network Effect & \textbf{11.18} (0.06) & 11.77 (0.07) & \textbf{11.18} (0.07)\\
    Polynomial Network Autocorrelation Effect & \textbf{10.70} (0.06) & 11.01 (0.07) & 10.74 (0.07)\\
    \bottomrule
    \end{tabular}
    \caption{Average width (and standard error) of 95\% prediction intervals for \method~using split-conformal inference, compared to the average width of 95\% posterior prediction intervals from BART and Network BART across various simulation settings. Results are aggregated across 100 simulation replicates.}
    \label{tab:conformal_widths}
\end{table}

\paragraph{Case Studies.} 
In addition to the simulations, we also applied the split-conformal prediction framework to the real data case studies described in Section~\ref{sec:case_study}.
In these case studies, we evaluated the split-conformal prediction intervals (at level $\alpha = 0.05$) for \method~using a leave-one-out evaluation procedure.
Specifically, we held out one observation as a test point and used the remaining data to conduct the 50-50\% training-calibration split and compute the prediction interval for the held-out test point.
This procedure was repeated for each observation in the dataset, and thus, the prediction interval coverage can be evaluated by computing the proportion of test points that fell within the estimated intervals.
The results, summarized in Table~\ref{tab:conformal_case_studies}, show that the coverage of the prediction intervals is close to the nominal level of 95\%, further supporting the validity of the split-conformal prediction intervals for \method~in real-world applications.

\begin{table}[tb]
\spacingset{1.1}
\centering
\begin{minipage}{0.47\textwidth}
\raggedleft
\begin{tabular}[t]{cc}
\toprule
\textbf{Data} &  \makecell{\textbf{Prediction}\\\textbf{Interval}\\\textbf{Coverage}}\\
\midrule
Philly Crime & 0.941\\
\cellcolor{gray!10}{School 1} & \cellcolor{gray!10}{0.950}\\
School 3 & 0.944\\
\cellcolor{gray!10}{School 6} & \cellcolor{gray!10}{0.959}\\
School 9 & 0.959\\
\cellcolor{gray!10}{School 10} & \cellcolor{gray!10}{0.953}\\
School 13 & 0.960\\
\cellcolor{gray!10}{School 19} & \cellcolor{gray!10}{0.954}\\
School 20 & 0.953\\
\cellcolor{gray!10}{School 21} & \cellcolor{gray!10}{0.941}\\
School 22 & 0.954\\
\cellcolor{gray!10}{School 24} & \cellcolor{gray!10}{0.953}\\
School 26 & 0.977\\
\cellcolor{gray!10}{School 27} & \cellcolor{gray!10}{0.956}\\
School 29 & 0.954\\
\bottomrule
\end{tabular}
\end{minipage}
\hfill
\begin{minipage}{0.47\textwidth}
\raggedright
\begin{tabular}[t]{cc}
\toprule
\textbf{Data} &  \makecell{\textbf{Prediction}\\\textbf{Interval}\\\textbf{Coverage}}\\
\midrule
& \\
\cellcolor{gray!10}{School 31} & \cellcolor{gray!10}{0.955}\\
School 33 & 0.958\\
\cellcolor{gray!10}{School 34} & \cellcolor{gray!10}{0.940}\\
School 35 & 0.953\\
\cellcolor{gray!10}{School 40} & \cellcolor{gray!10}{0.949}\\
School 42 & 0.946\\
\cellcolor{gray!10}{School 44} & \cellcolor{gray!10}{0.967}\\
School 45 & 0.957\\
\cellcolor{gray!10}{School 48} & \cellcolor{gray!10}{0.946}\\
School 49 & 0.950\\
\cellcolor{gray!10}{School 51} & \cellcolor{gray!10}{0.960}\\
School 56 & 0.950\\
\cellcolor{gray!10}{School 58} & \cellcolor{gray!10}{0.948}\\
School 60 & 0.953\\
\bottomrule
\end{tabular}
\end{minipage}
\caption{Coverage of split-conformal 95\% prediction intervals for \method~applied to Philly Crime and School Conflict case studies, evaluated using a leave-one-out procedure.}
\label{tab:conformal_case_studies}
\end{table}

\end{document}